\documentclass{article}
\usepackage[table,xcdraw]{xcolor}
\usepackage{multirow} 
\usepackage{microtype}
\usepackage{graphicx}
\usepackage{subfigure}
\usepackage{booktabs} 
\usepackage{xspace}

\usepackage{hyperref}
\usepackage{wrapfig}

\usepackage{calc}
\usepackage[final]{neurips_2025}
\author{
  Dylan Zhang \\
  University of Illinois Urbana–Champaign \\
  \texttt{shizhuo2@illinois.edu}
  \And
  Qirun Dai \\
  University of Chicago \\
  \texttt{qirundai@uchicago.edu}
  \And
  Hao Peng \\
  University of Illinois Urbana–Champaign \\
  \texttt{haopeng@illinois.edu}
}

\usepackage{amsmath}
\usepackage{amssymb}
\usepackage{mathtools}
\usepackage{amsthm}

\usepackage[capitalize,noabbrev]{cleveref}

\theoremstyle{plain}

\theoremstyle{definition}

\theoremstyle{remark}

\usepackage[utf8]{inputenc}

\newcommand{\crmodify}[1]{{#1}}
\newcommand{\modify}[1]{{#1}}

\usepackage{graphicx}
\newcommand{\name}{GRAPE\xspace}
\usepackage{bbm}
\usepackage{graphicx}
\usepackage{arydshln} 
\title{The Best Instruction-Tuning Data are Those That Fit}
\begin{document}
\maketitle
\begin{abstract}

High-quality supervised finetuning (SFT) data are essential for unlocking pretrained LLMs’ capabilities.
Typically, instructions are paired with responses \modify{from various sources by humans annotators or other LMs},
which are often out of the distribution of the target model to be finetuned.
This, at scale, can lead to diminishing returns and even hurt the models’ performance and robustness. We hypothesize that SFT is most effective with data aligned to the model’s pretrained distribution and propose \name --- a novel SFT framework that tailors supervision to the target model.
For each instruction, it {\bf g}athers {\bf r}esponses from various \modify{sources}, and selects the one
that {\bf a}ligns most closely to the target model's {\bf p}r\textbf{e}trained distribution, as measured by the normalized probability. We then proceed with standard SFT with these selected responses.
We first evaluate \name with a controlled experiment, where we sample various solutions for each question in UltraInteract from multiple models and finetune on \name-selected data using LMs from different families including LLaMA.1-8B, Mistral-7B and Qwen2.5-7B. \name significantly outperforms strong baselines, including  distilling from the strongest model with absolute gain up to 13.8\% averaging across benchmarks, and a baseline trained on 3$\times$ more data with maximum 17.3\% performance improvements.
\name's strong performance  generalizes to off-the-shelf SFT data.
We use \name to subsample responses from the post-training data used for Tulu3 and Olmo-2. 
\name can outperform strong baselines with 4.5 times the data by 6.1\% and state-of-the-art data selection approaches by 3.9\% on average performance. Remarkably, using 1/3 data and half number of epochs, \name allows LLaMA.1-8B to surpass the performance of Tulu3-SFT by 3.5\%. 
Our findings highlight that aligning supervision with the pretrained distribution offers a simple yet powerful way to improve SFT efficiency and performance.


\end{abstract}

\section{Introduction}

High-quality, large-scale supervised data is crucial for supervised fine-tuing(SFT;~\citealp{Dolly,openassist,zhao2024wildchat,zheng2024lmsyschatm}). A common practice of collecting SFT data involves sampling responses from strong language models, predominantly focusing on expanding the size of the dataset and improving the overall quality of the responses~\citep{sun2024principle, alpaca, wang-etal-2023-self-instruct, xu2024magpie, chen2024genqa}. However, recent research suggests that there is more complex dynamics involved~\citep{xu2024strongermodelsstrongerteachers}. A plateau effect in synthetic data scaling, where performance either stagnates or even declines as the size of the synthetic data increases beyond a certain point, has been widely observed.
This phenomenon arises due to issues such as diminishing diversity~\citep{padmakumar2024writing, guo2023curious} and distortion in the data distribution~\citep{lebrun2021evaluating}, which ultimately undermine the base model's performance and robustness~\citep{alemohammad2024selfconsuming, gerstgrasser2024iscollapseinevitable, shumailov2023curse,dohmatob2024strongmodelcollapse,Hataya_2023_ICCV,martínez2023combining,martínez2023understanding,bohacek2023nepotistically,briesch2023large}.

Thus, effective SFT requires more than scaling up the data; 
it often needs  ``tailoring'' the data to the unique characteristics of the target model. Existing works focus on enhancing the model’s existing knowledge and capabilities~\citep{du2023mods} and optimizing the curriculum progression for instruction tuning \citep{zhao2024preliminarystudyintrinsicrelationship, lee2024instructiontuninghumancurriculum,feng2023citinglargelanguagemodels,setlur2024rlincorrectsyntheticdata}. They typically find questions the model should best learn, instead of what answers the model should imitate.

Meanwhile, tailoring the responses to the target model has been a crucial ingredient for the success of later phases of LLM development, particularly through on-policy preference learning~\citep{tajwar2024shoulduse,zhang2024selfexploringlanguagemodelsactive,zhang2024textbfplumimprovingcodelms,miao2024aligningcodellmsdirectpreference,gulcehre2023rest,azar2023ipo,tang2024understandingperformancegap,zhuang2023bpo}, and on-policy/online reinforcement learning~\citep{guo2024onlineaifeedback,liu2024provablymitigatingoveroptimizationrlhf,zhou2024wpoenhancingrlhfweighted}.

Inspired by these insights,
we hypothesize that SFT can similarly benefit from aligning data with the model,
the core idea behind \name. For each instruction, \name gathers and selects response(s) from various sources that are closest to the target model's  pretrained distribution. 
This is achieved by calculating the probability of each response using the target model and selects the one with the highest length-normalized probability(\S\ref{sec:methodology}).After obtaining these more ``in-distribution'' responses, \name proceeds with standard SFT without any modification to the training.

Unlike existing datasets that usually contains one-size-fits-all responses for each instruction without customization~\citep{yu2024metamath, yuan2024eurus,slimorca,OpenHermes}, \name curates model-dependent SFT datasets that better matches the base model's distribution, better mitigating the risks associated with distribution shift like spurious correlations~\citep{zhou2024explorespuriouscorrelationsconcept} and catastrophic forgetting~\citep{luo2025empiricalstudycatastrophicforgetting,kotha2024understanding}, while posing minimum overhead of single forward pass over the candidate set.  In return, \name allows better downstream performance with reduced training compute. 


To \name's advantage, many existing datasets share overlapping instructions but contain different high-quality responses, e.g.,
the instructions in Flan~\citep{longpre2023flan}, GSM-8K~\citep{cobbe2021trainingverifierssolvemath}, MATH~\citep{hendrycks2021measuringmathematicalproblemsolving},
and the post-training recipes that re-use SFT instructions for preference learning~\citep{lambert2024tulu3,olmo2025}.
Therefore, \name can directly select, for each model, the best fit(s) among the off-the-shelf responses without having to produce new responses, which proves effective in our experiments (\S~\ref{sec:tulu_olmo}).

We first validate our hypothesis through extensive controlled experiments on a reasoning dataset \crmodify{with chain-of-thoughts}, UltraInteract-SFT~\citep{yuan2024eurus}, and demonstrate the importance of supervising base models with in-distribution responses(\S\ref{sec:experiments_ultra}). We experimented on 4 popular pretrained LMs from Mistral~\citep{mistralai_codestral_2024}, Llama3.1~\citep{dubey2024llama3herdmodels} and Qwen2.5~\citep{hui2024qwen25} families. 
Notably, models fine-tuned with \name-selected responses outperform those trained on a 3$\times$ larger datasets up to $17.3\%$ absolute gain on average performances even with less compute, even surpass models trained on responses from the strongest teacher under consideration---\textsc{Llama3.1-405B-Instruct}---by significant margins.

We then experiment with a more realistic setting \crmodify{for general-domain instruction-tuning}, collecting responses from post-training data for Tulu3 and Olmo-v2. Again, \name demonstrates its effectiveness by outperforming state-of-the-art data selection approaches by avg. 4.6\% and a strong baseline trained on all these available data that is 4.5 times larger by up to 6.1\%.  Remarkably, \name allows finetuning a Llama3.1-8B base model to exceed the performance of Tulu-8B-SFT using 1/3 data and half of the epochs.

Our results reveal that distributional alignment is a crucial and previously underappreciated dimension of effective instruction tuning. GRAPE introduces this perspective with a simple, scalable algorithm that consistently outperforms strong baselines with far less data and compute for the actual training.


\section{Background and Motivation}
\label{sec:motivation}
\subsection{Data Engineering for Instruction Tuning}
Data is central to the success of effective instruction tuning, ~\citep{xu2023rethinkinginstructionqualitylift,xia2024less,chan2024balancingcosteffectivenesssynthetic}, featuring both automated data synthesis~\citep{xu2024wizardlm,zeng2024automatic,yu2024metamath,wei2023magicoder} and selection~\citep{xia2024less,chen2023alpagasus,parkar2024selectllm,li2024scar}. Some selection approaches focus on high-quality data by leveraging LLMs~\citep{chen2023alpagasus,parkar2024selectllm,li2024superfilteringweaktostrongdatafiltering} or employing principled metrics~\citep{kang2024getmoreforless,mekala2024smaller,xia2024less}, while others, such as ~\citet{yang2024s2l,das2023deft}, aim to identify diversity-optimized subsets for greater efficiency.
An emerging trend is the customization of training data based on the characteristics of the base models.~\citep{li2024quantitytoqulality,du2023mods,li2024selective}. These methods typically reweight or filter \textit{\textbf{instructions}}. However, a key overlooked aspect is \textit{\textbf{response}} selection—specifically, choosing responses that align with the model’s pretrained distribution, which may be critical for preserving useful behaviors and ensuring effective fine-tuning.
\begin{wrapfigure}{L}{0.34\textwidth}
  \centering
  \includegraphics[width=\linewidth]{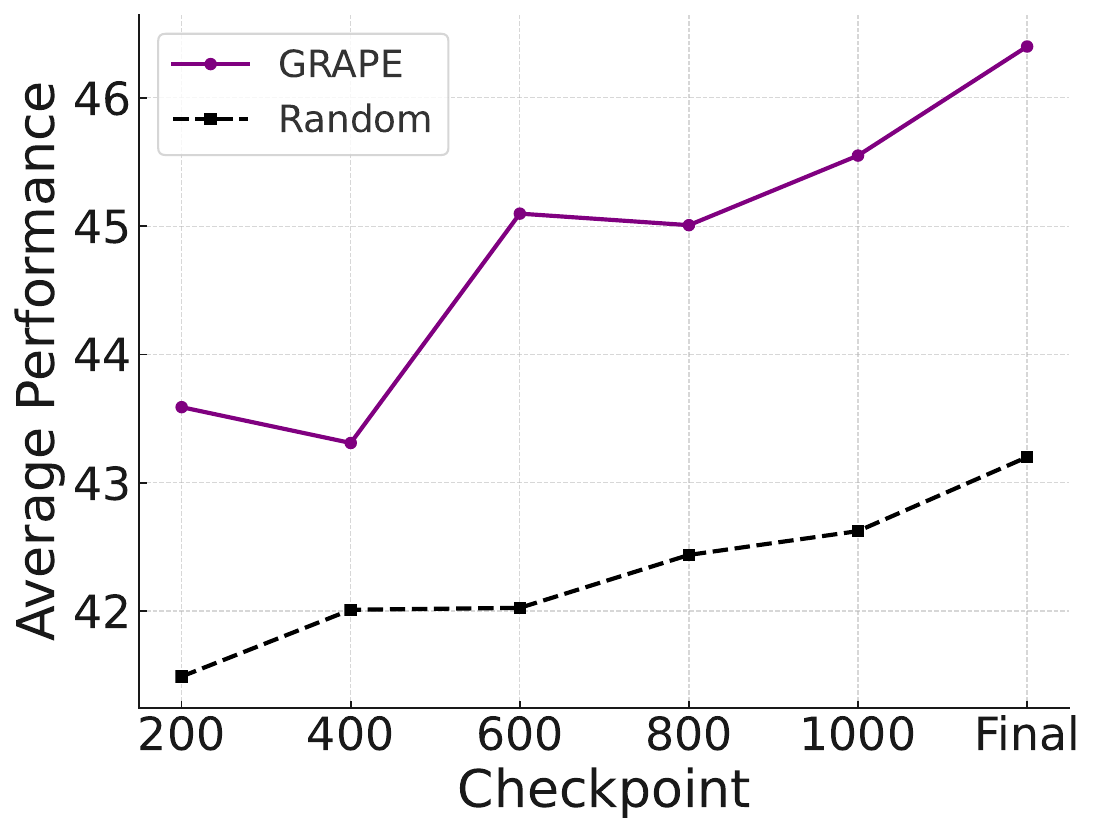}
  \caption{Average performance curve of \textsc{Llama3.1-8B} on \textsc{Tulu-3}-\textsc{Olmo-2} combo of \name against random selection}
  \label{fig:perf}
\end{wrapfigure}
\subsection{Toward Distribution-Aligned Supervised Fine-Tuning}

\paragraph{An Analogy from Reinforcement Learning and Preference Learning}

The investigation of this work into the distribution match between the pre-trained LM and supervised fine-tuning (SFT) data is inspired by recent findings on policy optimization for LM alignment with RL~\citep{ouyang2022training} and preference learning~\citep{rafailov2023dpo,ethayarajh2024kto}. 
While the importance of matching training data distribution with the policy has been well noted in both traditional RL~\citep{shi2023offlinereinforcementlearningonpolicy,fujimoto2018offpolicy,kumar2019offpolicy,peng2019offpolicy,wang2021criticregularizedregression,arora2023theory,jiang2016doublyrobustoffpolicyvalue,tang2010importancesampling} and LM settings~\citep{xiong2024iterative}, preference learning algorithms like DPO~\citep{rafailov2023dpo}, IPO~\citep{azar2023ipo} and KTO~\citep{ethayarajh2024kto} first emerged as off-policy algorithms. 
However, subsequent research has highlighted the performance gap between on-policy and off-policy training due to distribution shifts ~\citep{xu2024dposuperiorppo,tang2024understandingperformancegaponline} and proposed various mitigation strategies ~\citep{zhuang2023bpo,zhou2024wpoenhancingrlhfweighted,zhang2024textbfplumimprovingcodelms,xiong2024iterative,guo2024onlineaifeedback}, showing that training models on data more closely aligned with their policy distribution can significantly improve performance, while failing to do so can yield sub-optimal policies or those that are harder to generalize. \crmodify{Works like SPIN~\citep{yuan2024spin} echo the intuition by gradually improving policy through self-play to mitigate distribution shift.}

\begin{wrapfigure}{R}{0.4\textwidth}
  \centering
  \includegraphics[width=\linewidth]{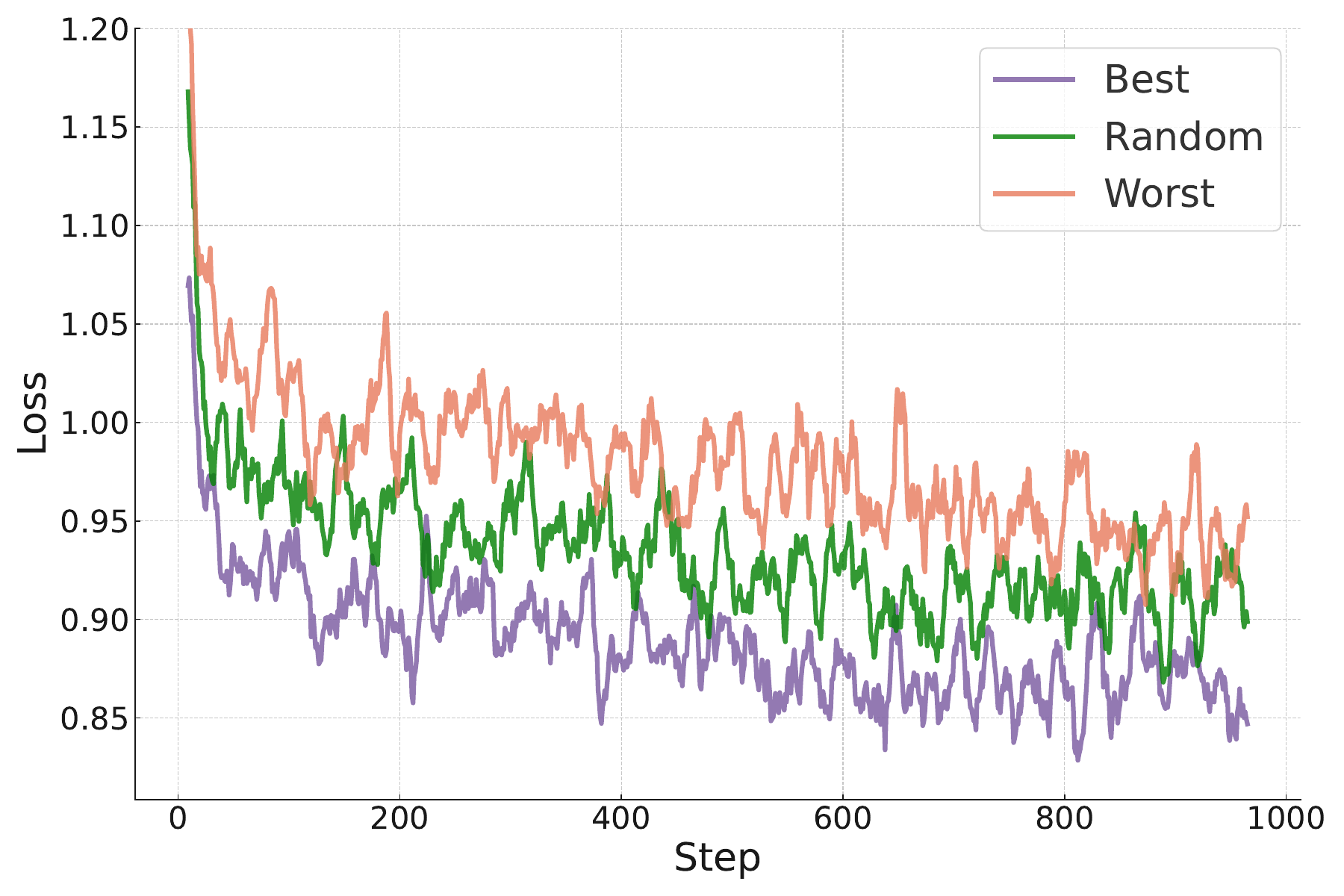}
  \caption{Training loss curve of Llama3.1-8B on \textsc{Tulu-3}-\textsc{Olmo-2}. \textbf{Best} (highest-probability, selected by \name), \textbf{Random}, and \textbf{Worst} (lowest-probability) responses show a clear loss ordering: \textbf{Best} $<$ \textbf{Random} $<$ \textbf{Worst}.}
  \label{fig:loss}
\end{wrapfigure}

\paragraph{Hypothesis: Supervised Fine-tuning Benefits From Data That Better Matches Base Distribution}
The base distribution of pre-trained language models---shaped by extensive training on vast and diverse datasets---is inherently robust and generalizable~\citep{brown2020languagemodelsfewshotlearners,saunshi2021mathematicalexplorationlanguagemodels}. Therefore, during supervised fine-tuning phase, the pre-trained distribution should be carefully preserved~\citep{kumar2022finetuningdistortpretrainedfeatures,cohenwang2024askdistributionshiftpretraining,he2023preservingpretrained,yang2024selfdistillationbridgesdistributiongap,ding2023peft}, to best retain the knowledge and capabilities that emerge during pre-training~\citep{zhou2023lima}. 
If the proximity between the pre-trained distribution and the fine-tuning data is not maintained, the limited number of training examples available during SFT, compared to the vast scale of pre-training data, can increase the risk of distribution distortion. This misalignment can lead to issues such as catastrophic forgetting~\citep{aghajanyan2020betterfinetuningreducingrepresentational,yang2024selfdistillationbridgesdistributiongap} and the emergence of spurious correlations~\citep{feldman2021doeslearningrequirememorization}.

The central premise of our work is that by using responses closely aligned to the pre-trained distribution, we can minimize distribution shift during SFT and therefore achieve better data efficiency and stronger performance. 
\paragraph{From On-Policy Alignment To Distribution-Aligned SFT}
\label{sec:on_policy}
We build on principles of on-policy alignment techniques with key distinctions tailored for SFT. 
Yet, given that SFT represents an earlier stage of post-training than RL, sampling responses from the base model itself alone can lead to model collapse, as noted in~\citep{collapse_when_trained_recursive}.
Prior studies have documented similar risks like instability, bias reinforcement, knowledge stagnation, and overfitting~\citep{herel2024collapseselftrainedlanguagemodel,mobahi2020selfdistillationamplifiesregularizationhilbert,allenzhu2023understandingensembleknowledgedistillation,ghosh2024closerlooklimitationsinstruction,dong2025selfboosting,zhang2024forcing}. To address this, we advocate for an approach that stays more in-distribution while delivering effective supervision to the base model. To this end, we propose to gather and select responses from various sources, and select one that is closest to the target model's pretrained distribution, which we name \name.

\section{Methodology}

\label{sec:methodology}
\begin{figure*}
    \centering
    \includegraphics[width=.8\linewidth]{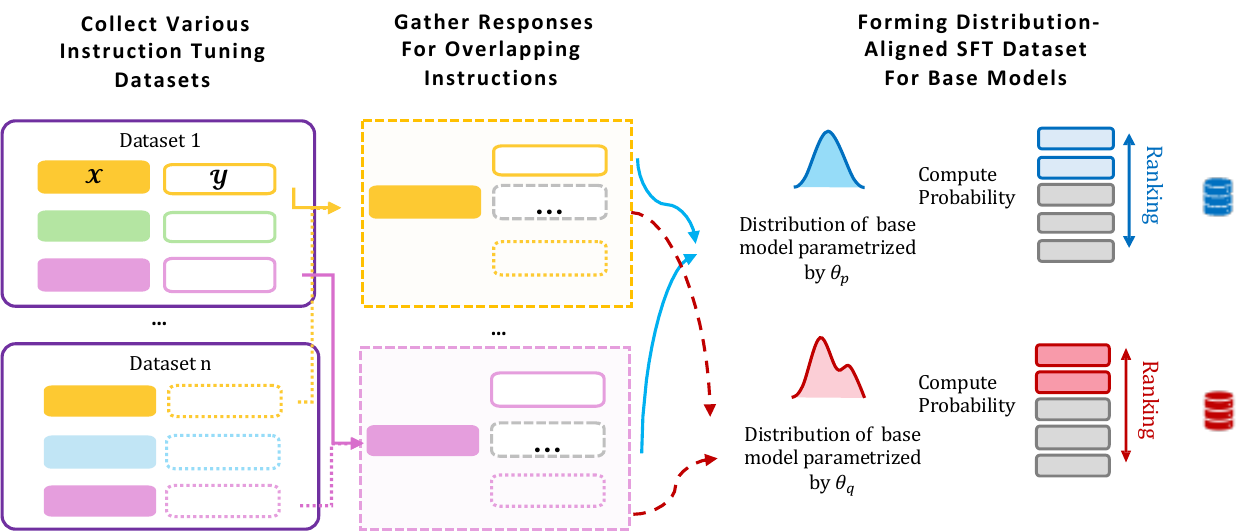}
    \caption{An overview of \name. \name takes multiple off-the-shelf existing datasets and optionally generates new responses, finds overlapping instructions with multiple different responses and selects responses that align with the base model's distribution. This  dataset is then used for standard supervised fine-tuning.}
    \label{fig:overview}
\end{figure*}

We introduce \name, a surprisingly simple yet effective methodology to enhance supervised fine-tuning (SFT) by customizing the training data for the base model. 
The key idea is to find a response, among a candidate pool, for each instruction \(x_i\) that aligns closely with the base model's pretrained distribution \(\pi_{\theta_0}\). 

As diagrammed in Figure~\ref{fig:overview}, \name consists of two main steps, followed by standard SFT:\\
\space\textbf{Response Collection (\S\ref{sec:method_response_collection})} Collect a pool of high-quality candidate responses from various sources. \\
\space\textbf{Customization (\S\ref{sec:method_customization}):} For the target model to be finetuned \(\pi_{\theta_0}\), find the response(s), for each instruction, that are closest to the pretrained distribution of \(\pi_{\theta_0}\).
\subsection{Collecting Responses from Existing Resources}
\label{sec:method_response_collection}

For instruction-tuning of language models, high-quality instructions are more difficult to collect than responses~\citep{xu2024magpie,liu2024best}. 
Therefore, it is a common practice to reuse existing instruction-tuning prompts while generating diverse responses using various methods tailored to specific requirements. For instance, instructions from Flan~\citep{longpre2023flan}, OpenOrca~\citep{OpenOrca}, ShareGPT~\citep{vicuna2023}, and the training splits of GSM-8K~\citep{cobbe2021trainingverifierssolvemath}, MATH~\citep{hendrycks2021measuringmathematicalproblemsolving}, and CodeContests~\citep{alphacode} are frequently reused in datasets like Olmo~\citep{olmo2025}, Tulu~\citep{lambert2024tulu3}, OpenHermes~\citep{OpenHermes}, OpenOrca~\citep{OpenOrca}, MetaMath~\citep{yu2024metamath}, MathInstruct~\citep{yue2023mammoth}, UltraFeedback~\citep{cui2024ultrafeedback}, and UltraInteract~\citep{yuan2024eurus}, whether for SFT or preference learning. The solutions are generated using different models or follow varying styles depending on the specific needs. 
This naturally leads to a situation where a single instruction with multiple responses becomes a readily available resource. \name therefore leverages these pre-existing response candidates to tailor training dataset that better aligns with the base model's distribution. 
When such resources are unavailable or insufficient, practitioners can generate new responses and apply \name.

For each instruction, we collect multiple responses from various datasets. These responses form a candidate set associated with the instruction. 
\subsection{Customize Dataset For Models}
\label{sec:method_customization}
We then compute the conditional probability of each response \(\pi_{\theta_0}(y{_i^j}\mid x_i)\). \modify{Practically, we format each example using a simple prompt template: \textbf{\textcolor{purple}{\texttt{Question:} \{instruction\} \textbackslash{}n \texttt{Answer:} \{response\}}}. For each instruction, we rank its candidate responses based on the conditional log-probability assigned by the base model, normalized by response length. This is equivalent to ranking them from lowest to highest perplexity where $\texttt{Perplexity} = \exp\left( -\frac{1}{N} \sum_{t=1}^{N} \log P(x_t \mid x_{<t}) \right)$.  We then select the responses with the \textbf{highest} normalized probability (i.e., lowest perplexity) for supervision.}
Figure~\ref{fig:breakdown} shows a clear difference in models' choices among the same candidate pool, indicating that \name-selected datasets are highly customized towards different models.

Since \name only involves forward-pass log-probability computation (no gradients or optimization), it is highly efficient and simple to integrate into any SFT pipeline with minimal overhead compared with model-based data selection approaches~\citep{xia2024less,yang2024s2l,liu2024deita,zhao2021datasetcondensationgradientmatching,TAGCOS,pan2024scalebioscalablebileveloptimization}.
Additionally, it is important to distinguish \name from the other perplexity-based data selection and curriculum planning methods~\citep{wu2024curriculumlearningqualitydrivendata,li2024superfilteringweaktostrongdatafiltering,liu2024letslearnstepstep}. Existing approaches focus on selecting \textbf{instructions} by using perplexity as a difficulty measure, which differs from \name that uses probability to select for each instruction in a fixed instruction set, responses that better matches with the base model's distribution. Our experiments in \S\ref{sec:tulu_olmo} demonstrate that low-probability responses with fixed set of instructions are detrimental to performance, further emphasizing the fundamental difference in the two processes. 

\section{\modify{Controlled Experiments Show the Benefits of Distributional Alignment}}
\label{sec:experiments_ultra}
We conduct controlled experiments using UltraInteract-SFT to test whether selecting responses aligned with a base model’s distribution improves fine-tuning outcomes more than relying on stronger generators or scaling data size. This setup—focused on verifiable tasks in coding, logic, and math—isolates the effect of distribution matching. Results show that \name-selected responses consistently outperform alternatives, validating distributional alignment as a key supervision signal. These findings motivate our broader evaluations in \S\ref{sec:tulu_olmo}.

\subsection{Experimental Setup}
\paragraph{Training Data Curation} 
In this controlled experiment, we focus on chain-of-thought  reasoning~\citep{wei2022chainofthought,wang2024mathshepherd,luo2024improvemathematicalreasoninglanguage,cobbe2021trainingverifierssolvemath,li2023making,lightman2023letsverifystepstep}. Different models may follow different reasoning paths to solve a problem, while their final solutions can be easily verified.

We use UltraInteract-SFT~\citep{yuan2024eurus}, which contains approximately $80,800$ unique instructions covering coding, math (chain-of-thought and program-aided) and logic reasoning domains , where each instruction is paired with varying numbers \modify{(avg. 3.5/instruction)} of different responses to contain a total $>280,000$ training examples. The responses in the dataset are strictly in step-wise format. 
\modify{For each instruction, we construct a response pool consisting of both (i) original UltraInteract-SFT responses and (ii) additional responses generated from a diverse set of LLMs. \name is then applied to select the most in-distribution response per instruction to this enlarged candidate pool and ensure the number of responses matches the original UltraInteract-SFT dataset for fair comparisons.}

We collect responses from a diverse set of models of various sizes across model families, including \textsc{Mixtral-7x7B-Instruct}~\citep{jiang2024mixtralexperts}, \textsc{Codestral-22B}~\citep{mistralai2024codestral22b}, \textsc{Mistral-Small}~\citep{mistralai2024mistralsmallinstruct}, \textsc{Llama-3.1-70B-Instruct} and \textsc{Llama-3.1-405B-Instruct}~\citep{dubey2024llama3herdmodels}, and \textsc{Qwen2.5-72B-Chat}~\citep{yang2024qwen2technicalreport}, resulting in approximately 10x additional responses per instruction. The responses are then filtered based on the answers to ensure their validity following~\citet{yuan2024eurus}.

\paragraph{Base Models} To demonstrate the generalizability of \name, we evaluate its performance across multiple LLMs, including \textsc{Llama-3.1-8B} and \textsc{Llama-3.2-3B} from \textsc{Llama-3}~\citep{grattafiori2024llama3herdmodels} family, \textsc{Mistral-7B}~\citep{jiang2023mistral7b} and \textsc{Qwen2.5-7B}~\citep{hui2024qwen25}.
\modify{We ensure that all training configurations (\name, baselines) use the same number of instructions and responses per instruction as original UltraInteract-SFT, unless otherwise stated.}
\paragraph{Evaluation}

We evaluate the model on coding and math reasoning benchmarks. For coding tasks, we consider HumanEval~\citep{chen2021humaneval}, MBPP~\citep{austin2021mbpp}, LeetCode~\citep{guo2024deepseekcoder}; for math datasets, we consider \textbf{MATH} dataset~\citep{hendrycks2021measuringmathematicalproblemsolving}, \textbf{GSM-Plus}~\citep{li2024gsmplus} and \textbf{TheoremQA}~\citep{chen2023theoremqatheoremdrivenquestionanswering} dataset. 
\textbf{HumanEval} and \textbf{MBPP} are natural-language-to-code benchmarks testing language models' ability to produce functionally correct programs. \textbf{LeetCode} contains interview-level programming problems that are more challenging. 
\textbf{MATH} contains high-school level math competition problems, whereas \textbf{GSM-Plus} is a more challenging variant of GSM-8k~\citep{cobbe2021trainingverifierssolvemath} 
and \textbf{Theorem-QA} contains complex math reasoning problems. 
\paragraph{Baselines}
We compare \name{} against several baselines to isolate the impact of response selection:
\textbf{Original Dataset} performs standard SFT on the unmodified UltraInteract-SFT dataset, which includes verified correct responses.
\textbf{Strongest-Model Responses} uses only responses from the most powerful generator available—\textsc{Llama3.1-405B-Instruct}. This represents a strong upper bound on data quality and helps determine whether \name{} offers additional gains beyond simply choosing a strong model.
\textbf{3$\times$Data:} \crmodify{ use 3 times the number of distinct, validated responses per instruction relative to UltraInteract, while keeping all training hyperparameters (learning rate, number of epochs, etc.) fixed. It directly tests whether \name{}’s gains stem from strategic data selection rather than merely scaling up data volume.} \footnote{For instance, if UltraInteract contains 3 validated responses for instruction $x$, this setting uses 9.}
We train all models for 1 epoch with a learning rate of $10^{-5}$.
 
\begin{table*}[h]
\centering
 \resizebox{\textwidth}{!}{
\small
\begin{tabular}{lcccccccccccc}
\hline
\multicolumn{1}{c}{} &
   &
   &
   &
   &
  \multicolumn{2}{c}{\textbf{MATH}} &
  \multicolumn{2}{c}{\textbf{GSMPlus}} &
  \multicolumn{2}{c}{\textbf{TheoremQA}} &
   &
   \\ \cline{6-11}
\multicolumn{1}{c}{\multirow{-2}{*}{\textbf{Model}}} &
  \multirow{-2}{*}{\textbf{Data}} &
  \multirow{-2}{*}{\textbf{HE}} &
  \multirow{-2}{*}{\textbf{LC}} &
  \multirow{-2}{*}{\textbf{MBPP}} &
  CoT &
  PoT &
  CoT &
  PoT &
  CoT &
  PoT &
  \multirow{-2}{*}{\textbf{Avg.}} &
  \multirow{-2}{*}{\textbf{\begin{tabular}[c]{@{}c@{}}Abs. \\ $\Delta$\end{tabular}}} \\ \hline
 &
  Original-UI &
  46.3 &
  15.6 &
  50.1 &
  21.6 &
  32.6 &
  45.9 &
  45.3 &
  16.8 &
  20.1 &
  32.7 &
  \textit{3.6} \\
 &
  Llama3.1-405B &
  44.5 &
  12.2 &
  46.9 &
  24.5 &
  33.8 &
  48.0 &
  50.0 &
  17.5 &
  16.8 &
  32.7 &
  \textit{3.6} \\
 &
  3x Data &
  48.1 &
  13.3 &
  52.8 &
  25.1 &
  26.1 &
  50.0 &
  49.4 &
  16.8 &
  12.4 &
  32.7 &
  \textit{3.6} \\
\multirow{-4}{*}{\textsc{Mistral-7B}} &
  \cellcolor[HTML]{F3F4FF}\textbf{\name} &
  \cellcolor[HTML]{F3F4FF}\textbf{52.4} &
  \cellcolor[HTML]{F3F4FF}\textbf{15.6} &
  \cellcolor[HTML]{F3F4FF}\textbf{53.4} &
  \cellcolor[HTML]{F3F4FF}\textbf{28.9} &
  \cellcolor[HTML]{F3F4FF}\textbf{34.6} &
  \cellcolor[HTML]{F3F4FF}\textbf{50.5} &
  \cellcolor[HTML]{F3F4FF}\textbf{52.8} &
  \cellcolor[HTML]{F3F4FF}\textbf{17.8} &
  \cellcolor[HTML]{F3F4FF}\textbf{20.6} &
  \cellcolor[HTML]{F3F4FF}\textbf{36.3} &
  \textit{-} \\ \hdashline
 &
  Original-UI &
  54.3 &
  11.1 &
  58.9 &
  29.7 &
  31.0 &
  53.7 &
  51.6 &
  20.0 &
  20.8 &
  36.8 &
  \textit{4.7} \\
 &
  Llama3.1-405B &
  56.7 &
  15.0 &
  60.0 &
  34.8 &
  38.1 &
  51.4 &
  55.4 &
  16.6 &
  21.0 &
  38.8 &
  \textit{2.7} \\
 &
  3x Data &
  48.8 &
  7.8 &
  57.9 &
  25.5 &
  11.2 &
  48.6 &
  45.1 &
  20.6 &
  19.6 &
  31.7 &
  \textit{9.8} \\
\multirow{-4}{*}{\textsc{Llama3.1-8B}} &
  \cellcolor[HTML]{F3F4FF}\textbf{\name} &
  \cellcolor[HTML]{F3F4FF}\textbf{57.3} &
  \cellcolor[HTML]{F3F4FF}\textbf{19.4} &
  \cellcolor[HTML]{F3F4FF}\textbf{63.8} &
  \cellcolor[HTML]{F3F4FF}\textbf{34.8} &
  \cellcolor[HTML]{F3F4FF}\textbf{39.2} &
  \cellcolor[HTML]{F3F4FF}\textbf{56.6} &
  \cellcolor[HTML]{F3F4FF}\textbf{56.1} &
  \cellcolor[HTML]{F3F4FF}\textbf{22.5} &
  \cellcolor[HTML]{F3F4FF}\textbf{23.9} &
  \cellcolor[HTML]{F3F4FF}\textbf{41.5} &
  \textit{-} \\ \hdashline
 &
  Original-UI &
  32.9 &
  3.9 &
  41.6 &
  12.8 &
  16.1 &
  30.8 &
  19.5 &
  14.6 &
  10.5 &
  20.3 &
  \textit{3.8} \\
 &
  Llama3.1-405B &
  31.7 &
  5.0 &
  43.3 &
  6.6 &
  6.6 &
  30.8 &
  20.6 &
  15.1 &
  10.8 &
  18.9 &
  \textit{5.1} \\
 &
  3x Data &
  42.6 &
  6.7 &
  42.9 &
  8.7 &
  5.1 &
  17.8 &
  19.5 &
  14.6 &
  \textbf{12.8} &
  19.0 &
  \textit{5.1} \\
\multirow{-4}{*}{\textsc{Llama3.2-3B}} &
  \cellcolor[HTML]{F3F4FF}\textbf{\name} &
  \cellcolor[HTML]{F3F4FF}\textbf{42.6} &
  \cellcolor[HTML]{F3F4FF}\textbf{13.3} &
  \cellcolor[HTML]{F3F4FF}\textbf{44.6} &
  \cellcolor[HTML]{F3F4FF}\textbf{16.4} &
  \cellcolor[HTML]{F3F4FF}\textbf{17.6} &
  \cellcolor[HTML]{F3F4FF}\textbf{34.9} &
  \cellcolor[HTML]{F3F4FF}\textbf{20.6} &
  \cellcolor[HTML]{F3F4FF}\textbf{15.1} &
  \cellcolor[HTML]{F3F4FF}{11.4}&
  \cellcolor[HTML]{F3F4FF}\textbf{24.1} &
  \textit{-} \\ \hdashline
 &
  Original-UI &
  67.0 &
  41.2 &
  60.0 &
  51.0 &
  38.3 &
  64.1 &
  59.5 &
  14.6 &
  10.5 &
  45.1 &
  \textit{11.6} \\
 &
  Llama3.1-405B &
  71.3 &
  45.0 &
  62.0 &
  31.5 &
  35.1 &
  40.1 &
  36.4 &
  33.3 &
  31.2 &
  42.9 &
  \textit{13.8} \\
 &
  3x Data &
  75.6 &
  48.3 &
  62.9 &
  32.8 &
  24.5 &
  47.1 &
  22.8 &
  21.6 &
  19.0 &
  39.4 &
  \textit{17.3} \\
\multirow{-4}{*}{\textsc{Qwen2.5-7B}} &
  \cellcolor[HTML]{F3F4FF}\textbf{\name} &
  \cellcolor[HTML]{F3F4FF}\textbf{77.4} &
  \cellcolor[HTML]{F3F4FF}\textbf{48.9} &
  \cellcolor[HTML]{F3F4FF}\textbf{70.7} &
  \cellcolor[HTML]{F3F4FF}\textbf{56.4} &
  \cellcolor[HTML]{F3F4FF}\textbf{45.3} &
  \cellcolor[HTML]{F3F4FF}\textbf{67.7} &
  \cellcolor[HTML]{F3F4FF}\textbf{66.3} &
  \cellcolor[HTML]{F3F4FF}\textbf{37.4} &
  \cellcolor[HTML]{F3F4FF}\textbf{40.1} &
  \cellcolor[HTML]{F3F4FF}\textbf{56.7} &
  \textit{-} \\ \hline

\end{tabular}}
\caption{Result of synthetic experiment on UltraInteract-SFT. The last column, \textbf{Abs. $\Delta$} is \name's absolute improvement on average performance over that row. }
\label{tab:main}
\end{table*}

\subsection{Results and Analysis}
Table~\ref{tab:main} summarizes the performance of \name{} across benchmarks.
Our approach consistently outperforms the various baselines across the board, including the original UltraInteract-SFT dataset. \name-selected solutions can outperform those directly sampled from the strongest model under consideration (\textsc{Llama3.1-405B-Instruct}) up to 13.8\% absolute improvement. 
This implies that customization for base models should be prioritized over identifying the presumably highest-quality responses. This verifies our central premise that being in-distribution with each base model is an important ingredient for the responses we supervise the base models on, to boosting downstream performance.
Furthermore, we demonstrate that merely adding more responses does not always lead to continuous improvement in model performance, which aligns with findings in prior studies~\citep{li2024quantitytoqulality,du2023mods}. By properly aligning with models' base distributions, \name outperforms those trained with 3x responses with at least 3.6\% and up to 17.3\% absolute improvement. These results reinforce the notion that scaling data without considering its alignment with the base model's initial distribution risks diminishing returns and, in some cases, even performance degradation.

\section{\name-Picking From Real-World SFT Datasets}
\label{sec:tulu_olmo}
In this section, we leverage the findings from the earlier experiments and demonstrate the effectiveness of \name to customize  training data for each base model by selecting from available datasets with overlapping instructions. 
Here, we do \emph{not} generate any new responses for the instructions; it only selects from existing ones.
We evaluate \name on the fully open dataset used in post-training phases of \textsc{Tulu-3}~\citep{lambert2024tulu3} and \textsc{Olmo-2}~\citep{olmo2025}. The details are presented below. We discuss additional results in Appendix~\ref{app:additional_exp}

\subsection{Data Mixture Details}

\textsc{Tulu-3}~\citep{lambert2024tulu3} is a fully open-source collection of post-training recipes, including supervised fine-tuning and preference alignment data. \textsc{Olmo-2}~\citep{olmo2025} is a fully open-source language model. Both \textsc{Tulu-3} and \textsc{Olmo-2} use the same data mixture during the supervised fine-tuning stage, but different data mixtures and source models for generating preference data for different sizes of their models: Tulu-3-8B/70B and Olmo-2-7B/13B. 
To demonstrate the effectiveness of \name, we collected the overlapping instructions from both models and gather their corresponding responses.
\begin{wrapfigure}{r}{0.55\linewidth}  
  \centering
  \includegraphics[width=\linewidth]{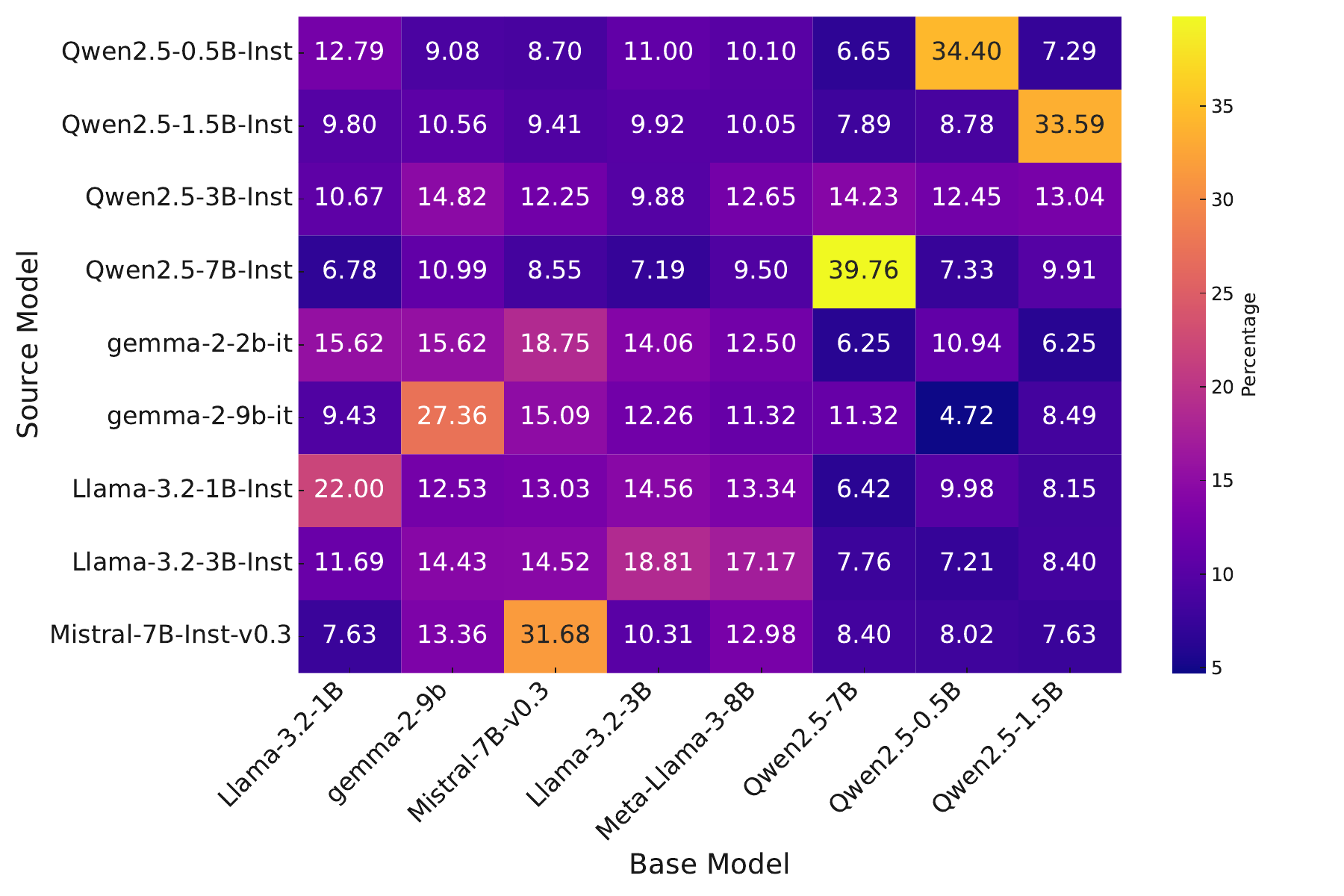}
  \vspace{-5mm}
  \caption{Breakdown of \name-selected responses for 1K Tulu instructions vary significantly across base models, reflecting its highly model-oriented nature over responses. Details in Appendix~\ref{app:heatmap}}
  \label{fig:breakdown}
\end{wrapfigure}
From the preference data, we retained only the winning responses. We formed our candidate pool with those instructions with at least two distinct responses, resulting in a dataset of 350.4K unique instructions and about 1.03 million total instruction-response pairs for evaluation with \name.
We do \emph{not} apply further processing of these data or any filtering on top of \name.
\subsection{Evaluation}

We evaluate on a set of commonly used benchmarks spanning over coding, math, knowledge and instruction-following. We evaluated on LeetCode~\citep{guo2024deepseekcoder}, MATH~\citep{hendrycks2021measuringmathematicalproblemsolving}, BigBenchHard(BBH)~\citep{suzgun2022bbh}, MMLU~\citep{hendrycks2021mmlu}, and AlpacaEval-V2~\citep{dubois2024alpacaevalv2}. LeetCode, MATH, BBH and MMLU are evaluated the in the same way as in~\citep{yuan2024eurus}, where we use zero-shot for MATH and MMLU, 3-shot example for BBH. We use the same AlpacaEval-v2 as in OpenInstruct.  
\subsection{Baselines}
\modify{We extensively compare against three baseline types: controlled baselines with fixed instructions, scaling baselines with increased data, and state-of-the-art selection methods. Additional details in Appendix~\ref{app:baseline_details} }
\paragraph{Controlled Baselines}
We include three baselines to isolate the effect of \name's response selection. 
\textbf{SFT-only} replaces \name-selected responses with those from the original SFT dataset, pairing each instruction with a standard reference response to measure improvement over presumably good SFT responses. 
\textbf{Random} selects candidate responses uniformly at random from the pool for the same set of instructions, establishing a noise-tolerant baseline.
\textbf{Reverse-\name} instead selects responses with the highest perplexity. We test if responses diverging the most from the base model’s distribution degrade performance, providing a contrast that sharpens the effectiveness of \name.

\paragraph{Scaling Baselines}
To assess how \name compares under larger-scale training, we consider three scaling-oriented baselines.
\textbf{Tulu3-SFT} uses all 939K SFT instances from the Tulu3 training mixture to test whether \name can still outperform despite using only a subset of the instruction pool.
\textbf{All Responses} trains over the entire 1.04M-instance candidate pool to demonstrate the effect of selection versus brute-force inclusion.
\textbf{All Available Data} uses all 1.58M instruction-response pairs under consideration, roughly 4.5$\times$ the data used by \name, to test whether data volume alone suffices.

\paragraph{State-Of-The-Art SFT Data Selection Approaches}
We compare \name against recent state-of-the-art data selection methods.
\textbf{LESS}~\citep{xia2024less} selects data based on influence scores on validation tasks. We follow the implementation setup from Dai et al.~\citep{dai2025bids}. 
\textbf{Emb-NV} selects training data close to validation in embedding space using NV-Embed-V2~\citep{lee2025nvemb}, following~\citep{ivison2025largescaledataselectioninstruction}.
\textbf{S2L}~\citep{yang2024s2l} clusters data via loss trajectories from small reference models (\textsc{Llama-3.2-1B}, \textsc{Qwen2.5-0.5B}, \textsc{Mistral-v0.3-7B}) and samples uniformly across clusters to match \name's data budget.

\begin{table*}[h]
\small
\centering
 \resizebox{\textwidth}{!}{
\begin{tabular}{ccccccccccc}
\hline
 &                             &        & \multicolumn{2}{c}{\textbf{AlpacaEval2}} &               &      &               &               &      &     \\ \cline{4-5}
\multirow{-2}{*}{\textbf{Model}} &
  \multirow{-2}{*}{\textbf{Data}} &
  \multirow{-2}{*}{\textbf{Num. Instances}} &
  \textbf{LC} &
  \textbf{WR} &
  \multirow{-2}{*}{\textbf{BBH}} &
  \multirow{-2}{*}{\textbf{MMLU}} &
  \multirow{-2}{*}{\textbf{MATH}} &
  \multirow{-2}{*}{\textbf{LeetCode}} &
  \multirow{-2}{*}{\textbf{Avg.}} &
  \multirow{-2}{*}{\textbf{\begin{tabular}[c]{@{}c@{}}Abs.\\ $\Delta$\end{tabular}}} \\ \hline
 & \textbf{Highest}            & 350.4k & 10.1                & 6.2                & 68.9          & 63.2 & 22.9          & 13.3          & 30.8 & 5.2 \\
 & \textbf{Random}             & 350.4k & 12.8                & 10.8               & 68.6          & 63.1 & 27.9          & 13.3          & 32.8 & 3.2 \\
 & \textbf{SFT-Only}           & 350.4k & 7.1                 & 5.5                & 68.9          & 64.1 & 20.2          & 17.2          & 30.5 & 5.4 \\
 &
  \cellcolor[HTML]{FFFFE8}\textbf{Tulu3-SFT} &
  \cellcolor[HTML]{FFFFE8}939k &
  \cellcolor[HTML]{FFFFE8}12.4 &
  \cellcolor[HTML]{FFFFE8}8.0 &
  \cellcolor[HTML]{FFFFE8}67.9 &
  \cellcolor[HTML]{FFFFE8}\textbf{65.9} &
  \cellcolor[HTML]{FFFFE8}31.5* &
  \cellcolor[HTML]{FFFFE8}7.8 &
  \cellcolor[HTML]{FFFFE8}32.4 &
  \cellcolor[HTML]{FFFFE8}3.5 \\
 & \textbf{All Responses}      & 1.03M  & 12.9                & 11.4               & 68.7          & 62.8 & 32.1          & 17.2          & 34.2 & 1.8 \\
 & \textbf{S2L}                & 350.4k & 8.5                 & 7.6                & 68.6          & 63.1 & 26.5          & 16.1          & 31.7 & 4.2 \\
   & \textbf{Emb-NV}             & 350.4k & 8.4                & 7.0              & 67.9          & 63.7 & 32.1          & 17.2          & 32.7 & 3.2 \\
 & \textbf{LESS}               & 350.4k & 7.3                & 6.0               & 68.2          & 63.3 & 25.1          & 16.1          & 31.0 & 4.0 \\
 & \textbf{All Available Data} & 1.58M  & 8.8                 & 10.1               & \textbf{69.8} & 62.1 & \textbf{32.5} & 16.1          & 33.2 & 2.7 \\

\multirow{-8}{*}{\textsc{\begin{tabular}[c]{@{}c@{}}Llama\\ 3.1-8B\end{tabular}}} &
  \cellcolor[HTML]{F3F4FF}\textbf{\name} &
  \cellcolor[HTML]{F3F4FF}350.4k &
  \cellcolor[HTML]{F3F4FF}\textbf{14.8} &
  \cellcolor[HTML]{F3F4FF}\textbf{15.2} &
  \cellcolor[HTML]{F3F4FF}69.6 &
  \cellcolor[HTML]{F3F4FF}64.5 &
  \cellcolor[HTML]{F3F4FF}32.1 &
  \cellcolor[HTML]{F3F4FF}\textbf{19.4} &
  \cellcolor[HTML]{F3F4FF}\textbf{35.9} &
  \cellcolor[HTML]{F3F4FF}- \\ \hdashline
 & \textbf{Highest}            & 350.4k & 7.0                 & 5.5                & 58.8          & 56.1 & 15.1          & 10.6          & 25.5 & 6.4 \\
 & \textbf{Random}             & 350.4k & 10.6                & 9.2                & 60.0          & 57.8 & 19.6          & 11.1          & 28.1 & 3.9 \\
 & \textbf{SFT-Only}           & 350.4k & 7.1                 & 5.3                & 53.9          & 57.0 & 14.4          & 12.0          & 24.9 & 7.0 \\
 & \textbf{Full-SFT-Data}      & 939k   & 11.5                & 10.5               & 59.0          & 55.2 & \textbf{25.8} & 15.6          & 29.9 & 2.0 \\
 & \textbf{All Responses}      & 1.03M  & 10.5                & 11.5               & 61.0          & 57.9 & 24.2          & 14.4          & 29.9 & 2.0 \\
 & \textbf{S2L}                & 350.4k & 10.5                & 11.9               & 61.9          & 57.1 & 22.4          & 13.9          & 29.6 & 2.3 \\
   & \textbf{Emb-NV}             & 350.4k & 6.0                & 5.2                & 60.7          & 56.5 & 23.9          & 11.7          & 27.3 & 4.6 \\
 & \textbf{LESS}               & 350.4k & 6.4                 & 4.8                & 59.2          & 55.4 & 16.0          & 8.3           & 25.0 & 6.9 \\
 & \textbf{All Available Data} & 1.58M  & 8.0                 & 7.0                & 55.3          & 53.7 & 25.4          & 12.3          & 26.9 & 5.0 \\\multirow{-8}{*}{\textsc{\begin{tabular}[c]{@{}c@{}}Mistral\\ -7B\end{tabular}}} &
  \cellcolor[HTML]{F3F4FF}\textbf{\name} &
  \cellcolor[HTML]{F3F4FF}350.4k &
  \cellcolor[HTML]{F3F4FF}\textbf{13.6} &
  \cellcolor[HTML]{F3F4FF}\textbf{13.9} &
  \cellcolor[HTML]{F3F4FF}\textbf{62.3} &
  \cellcolor[HTML]{F3F4FF}\textbf{59.2} &
  \cellcolor[HTML]{F3F4FF}24.2 &
  \cellcolor[HTML]{F3F4FF}\textbf{18.3} &
  \cellcolor[HTML]{F3F4FF}\textbf{31.9} &
  \cellcolor[HTML]{F3F4FF}- \\ \hdashline
 & \textbf{Highest}            & 350.4k & 8.0                 & 10.7               & 72.2          & 73.2 & 49.4          & 42.2          & 42.6 & 5.9 \\
 & \textbf{Random}             & 350.4k & 16.1                & 14.9               & \textbf{73.3} & 73.1 & 56.0          & 43.3          & 46.1 & 2.4 \\
 & \textbf{SFT-Only}           & 350.4k & 9.9                 & 7.8                & 71.2          & \textbf{74.1} & 51.1          & 46.6          & 43.4 & 5.1 \\
 & \textbf{Full-SFT-Data}      & 939k   & 9.5                 & 7.1                & 71.4          & 73.1 & 47.0          & \textbf{48.3} & 42.7 & 5.8 \\
 & \textbf{All Responses}      & 1.03M  & 16.0                & 14.5               & 71.4          & 72.1 & 51.7          & 43.3          & 44.8 & 3.7 \\
 & \textbf{S2L}                & 350.4k & 13.4                & 14.9               & 72.7          & 73.1 & 53.4          & 40.6          & 44.7 & 3.9 \\
  & \textbf{Emb-NV}             & 350.4k & 11.9                & 10.6             & 72.1          & 72.5 & 53.3          & 43.3          & 44.0 & 4.6 \\
 & \textbf{LESS}               & 350.4k & 7.8                & 5.9               & 71.3          & 72.9 & 48.3          & 41.1          & 41.2 & 7.4 \\
 
 & \textbf{All Available Data} & 1.58M  & 13.3                & 12.3               & 70.3          & 71.8 & 44.0          & 42.8          & 42.4 & 6.1 \\
 
\multirow{-8}{*}{\textsc{\begin{tabular}[c]{@{}c@{}}Qwen2.5\\ -7B\end{tabular}}} &
  \cellcolor[HTML]{F3F4FF}\textbf{\name} &
  \cellcolor[HTML]{F3F4FF}350.4k &
  \cellcolor[HTML]{F3F4FF}\textbf{20.0} &
  \cellcolor[HTML]{F3F4FF}\textbf{20.4} &
  \cellcolor[HTML]{F3F4FF}73.2 &
  \cellcolor[HTML]{F3F4FF}73.3 &
  \cellcolor[HTML]{F3F4FF}\textbf{60.0} &
  \cellcolor[HTML]{F3F4FF}44.4 &
  \cellcolor[HTML]{F3F4FF}\textbf{48.6} &
  \cellcolor[HTML]{F3F4FF}- \\ \hline
\end{tabular}
}
\caption{\name on the Tulu-Olmo collection. For Llama3.1-8B base model, we included Tulu3-SFT model's results. The ``*"-marked number for MATH is obtained using 4-shot prompting. We train all the models (except that we took Tulu3-SFT numbers directly) for 1 epoch with a learning rate of $10^{-5}$. Abs.$\Delta$ is \name's absolute improvement on average performance over that row. }
\label{tab:olmo-tulu}
\end{table*}
\subsection{Results}

\begin{wraptable}{r}{0.5\textwidth}   
  \vspace{-1.5\baselineskip}           
  \small
  \centering
   \resizebox{\linewidth}{!}{
  \begin{tabular}{c
                  >{\columncolor[HTML]{F3F4FF}}c
                  ccc}
    \hline
     &
     \textsc{\name} &
     \textsc{\begin{tabular}[c]{@{}c@{}}Qwen2.5\\ -72B\end{tabular}} &
     \textsc{\begin{tabular}[c]{@{}c@{}}Llama3.1\\ -405B\end{tabular}} &
     \textsc{\begin{tabular}[c]{@{}c@{}}Gemma\\ -it-9B\end{tabular}} \\
    \hline
    \textbf{LC} &
      \textbf{28.1} &
      25.8 &
      16.3 &
      26.9 \\
    \hline
    \textbf{WR} &
      \textbf{33.1} &
      24.3 &
      17.0 &
      20.2 \\
    \hline
  \end{tabular}}
  \caption{Alpaca-Eval2 On Magpie-Zoo~\cite{xu2024strongermodelsstrongerteachers}.}
  \label{tab:magpie_zoo}
\end{wraptable}
As shown in Table~\ref{tab:olmo-tulu}, models fine-tuned on responses selected by \name outperforms the strong baselines we constructed, especially the one that trains over all available data by significant margins across the 3 models. 

Remarkably, using roughly 1/6 training computation (Tulu3-8B-SFT was trained for 2 epochs on 3 times of data), our performance exceeds that of \textsc{Tulu3-8B-SFT}. 

Also, \name outperforms state-of-the-art data-selection approaches like S2L, despite its simplicity and efficiency, further highlighting its effectiveness in diverse real-world scenarios and making it a practical option for real-world SFT setups with minimal data engineering effort.
Without the need to synthesize any new data, one can easily leverage established datasets sourced from the web to customize a dataset for each base model that yields better fine-tuning outcome. 

These results highlight \name as an effective and efficient selection strategy for real-world SFT.

\subsection{Ablations}

\paragraph{Remains Effective Even with a Single Response Source}

In Section~\ref{sec:experiments_ultra}, we demonstrated how \name enables practitioners to refine model-generated responses for improved training outcomes. Beyond that, \name can optimize responses from a single generator. 
\modify{The Magpie-Zoo~\citep{xu2024strongermodelsstrongerteachers} dataset contains a fixed set of instructions and multiple versions of response sets each generated by different language models.}
Using the Magpie-Zoo instruction set, we sample 10 responses per instruction from Qwen2.5-72B-Instruct, select in-distribution responses with \name, to train a Mistral-v0.3-7B model. \modify{We compare with replicas that achieve top-3 performance from Magpie-Zoo.}


\begin{wraptable}{r}{0.4\textwidth}   
  \vspace{-1.5\baselineskip}          
  \small
  \centering\resizebox{\linewidth}{!}{
  \begin{tabular}{ccc}
    \hline
                           &                           &                       \\
    \multirow{-2}{*}{\textbf{Model}} 
                           & \multirow{-2}{*}{\textbf{Data}} 
                                                         & \multirow{-2}{*}{\textbf{Avg.}} \\
    \hline
                           & \cellcolor[HTML]{F8E8E7}Self-Distilled & \cellcolor[HTML]{F8E8E7}28.4\textcolor{red}{\textbf{(-)}} \\
    \multirow{-2}{*}{\textsc{Mistral-7B}} 
                           & Original-UI                & 32.7                  \\ 
    \hline
                           & \cellcolor[HTML]{F8E8E7}Self-Distilled & \cellcolor[HTML]{F8E8E7}29.4\textcolor{red}{\textbf{(-)}} \\
    \multirow{-2}{*}{\textsc{Llama3.1-8B}} 
                           & Original-UI                & 36.8                  \\ 
    \hline
                           & \cellcolor[HTML]{F8E8E7}Self-Distilled & \cellcolor[HTML]{F8E8E7}15.1\textcolor{red}{\textbf{(-)}} \\
    \multirow{-2}{*}{\textsc{Llama3.2-3B}} 
                           & Original-UI                & 20.3                  \\ 
    \hline
  \end{tabular}}
  \caption{Performance Degradation From Self-Distillation On UI.}
  \label{tab:self_distill_ultra}
  \vspace{-1.5\baselineskip}  
\end{wraptable}
As shown in Table~\ref{tab:magpie_zoo}, \name-selected responses further boost the performance. \modify{
Given that batch sampling from a strong model introduces minimal latency~\citep{zhong2024distservedisaggregatingprefilldecoding,zhou2024survey}, this result positions \name as a practical and efficient data curation strategy—achieving strong results with just a single generator and no additional engineering overhead.
}
\paragraph{Why \name Outperforms Self-Generated Responses}
\label{sec:self_distillation}


\begin{wraptable}{l}{0.6\textwidth} 
  \centering
  \small \resizebox{\linewidth}{!}{
  \begin{tabular}{ccccc}
    \hline
    \textbf{Dataset} & \textbf{Model} & \textbf{Response Generator} & \textbf{N} & \textbf{Acc.} \\ 
    \hline
    MATH & \textsc{Mistral} & Llama3.1-70B      & 10 & 18.2     \\
    \rowcolor[HTML]{F8E8E7} 
    MATH & \textsc{Mistral} & FT-Mistral        & 10 & 15.9 \textbf{\textcolor{red}{(-)}} \\ 
    \hline
    MATH & \textsc{Llemma}  & Llama3.1-70B      & 10 & 26.2     \\
    \rowcolor[HTML]{F8E8E7} 
    MATH & \textsc{Llemma}  & FT-Llemma         & 10 & 23.6 \textbf{\textcolor{red}{(-)}} \\ 
    \hline
    MATH & \textsc{Mistral} & MM-AnsAug         & –  & 22.3     \\
    \rowcolor[HTML]{F8E8E7} 
    MATH & \textsc{Mistral} & FT-Mistral        & 10 & 20.6 \textbf{\textcolor{red}{(-)}} \\ 
    \hline
    MATH & \textsc{Llemma}  & MM-AnsAug         & –  & 28.1     \\
    \rowcolor[HTML]{F8E8E7} 
    MATH & \textsc{Llemma}  & FT-Llemma         & 10 & 21.4 \textbf{\textcolor{red}{(-)}}\\ 
    \hline
  \end{tabular}}
  \caption{Self-generation on MATH dataset. \textsc{FT-Mistral} refers to the model right above that row finetuned from either MM-AnsAug or Llama3.1-70B-Instruct produced solutions. N stands for the number of responses sampled.}
  \label{tab:self_gen_math}
\end{wraptable}
To probe the effectiveness of \name, we ablate it against a degenerate alternative: self-generation, where the model is fine-tuned on its own outputs.

We fine-tune a base model using responses generated by its previously fine-tuned variant. This setup consistently degrades performance (Table~\ref{tab:self_distill_ultra}). We further confirm this on the MATH dataset~\citep{hendrycks2021measuringmathematicalproblemsolving}: responses from strong models like \textsc{Llama3.1-70B-Instruct} or MetaMathQA~\citep{yu2024metamath} are used to fine-tune a model that then generates new solutions for another round of fine-tuning. Again, performance drops (Table~\ref{tab:self_gen_math}).

This failure arises from distributional collapse (see \S~\ref{sec:on_policy}): self-generated responses become increasingly narrow and repetitive, reinforcing biases and reducing exposure to diverse reasoning. Correctness alone is insufficient—external diversity is essential for generalization.

\name avoids this collapse by selecting external responses that are both diverse and distribution-aligned, preserving semantic breadth and stylistic variability while staying true to the model’s pretraining. This enables more stable and generalizable fine-tuning.
\section{Conclusion}
We present \name, a simple yet effective method for improving supervised fine-tuning by selecting responses aligned with the base model’s pretrained distribution. GRAPE requires only a forward pass over candidate responses, making it highly efficient and easy to integrate.
Despite its simplicity, \name consistently outperforms stronger baselines using significantly larger datasets and surpasses more complex, costly data selection methods.
Our study affirms that carefully aligning SFT data with a model’s pretrained distribution yields substantial performance and efficiency gains.

\section{Discussion and Limitations}
\paragraph{Response versus Instance Level Selection}
\name selects responses: it begins with a fixed set of instructions and evaluates multiple candidate responses for each. This differs from instance-level selection approaches (e.g., \cite{kung2023ait, li2024quantitytoqulality, wang2024diversitymeasurementsubsetselection}), which focus on choosing which instructions to include based on factors such as coverage, skill-balancing or difficulty.
In contrast, \name focuses on the quality of supervision—that is, selecting responses to provide the most effective learning signal. Importantly, this response-centric perspective is complementary, not contradictory, to instance-level or instruction-based selection methods; both might be combined to enhance overall data quality and training efficiency.

\paragraph{Limitations} Like many other data selection algorithms~\citep{du2023mods,li2024quantitytoqulality,xia2024less,das2023deft,kang2024getmoreforless,mekala2024smaller,yang2024smalltolarge,TAGCOS,pan2024scalebioscalablebileveloptimization,dai2025bids,ivison2025largescaledataselectioninstruction,FacilityLocations,yin2024computeconstraineddataselection,liu2024deita}, \name assumes a quality-controlled candidate pool from which to select samples. Furthermore, because \name relies on the base model itself, its selection effectiveness may be influenced by the model’s inherent capabilities.
\section{Acknowledgment}
This project is partly supported by NSF under award No. 2019897. This research used the DeltaAI advanced computing and data resource, which is supported by the National Science Foundation (award OAC 2320345) and the State of Illinois. DeltaAI is a joint effort of the University of Illinois Urbana-Champaign and its National Center for Supercomputing Applications. 
\bibliography{custom,beyond-model-collapse,plum}

\begin{thebibliography}{160}
\providecommand{\natexlab}[1]{#1}
\providecommand{\url}[1]{\texttt{#1}}
\expandafter\ifx\csname urlstyle\endcsname\relax
  \providecommand{\doi}[1]{doi: #1}\else
  \providecommand{\doi}{doi: \begingroup \urlstyle{rm}\Url}\fi

\bibitem[Aghajanyan et~al.(2020)Aghajanyan, Shrivastava, Gupta, Goyal, Zettlemoyer, and Gupta]{aghajanyan2020betterfinetuningreducingrepresentational}
Aghajanyan, A., Shrivastava, A., Gupta, A., Goyal, N., Zettlemoyer, L., and Gupta, S.
\newblock Better fine-tuning by reducing representational collapse, 2020.
\newblock URL \url{https://arxiv.org/abs/2008.03156}.

\bibitem[Alemohammad et~al.(2024)Alemohammad, Casco-Rodriguez, Luzi, Humayun, Babaei, LeJeune, Siahkoohi, and Baraniuk]{alemohammad2024selfconsuming}
Alemohammad, S., Casco-Rodriguez, J., Luzi, L., Humayun, A.~I., Babaei, H., LeJeune, D., Siahkoohi, A., and Baraniuk, R.
\newblock Self-consuming generative models go {MAD}.
\newblock In \emph{The Twelfth International Conference on Learning Representations}, 2024.
\newblock URL \url{https://openreview.net/forum?id=ShjMHfmPs0}.

\bibitem[Allen-Zhu \& Li(2023)Allen-Zhu and Li]{allenzhu2023understandingensembleknowledgedistillation}
Allen-Zhu, Z. and Li, Y.
\newblock Towards understanding ensemble, knowledge distillation and self-distillation in deep learning, 2023.
\newblock URL \url{https://arxiv.org/abs/2012.09816}.

\bibitem[Ankner et~al.(2024)Ankner, Blakeney, Sreenivasan, Marion, Leavitt, and Paul]{ankner2024perplexed}
Ankner, Z., Blakeney, C., Sreenivasan, K., Marion, M., Leavitt, M.~L., and Paul, M.
\newblock Perplexed by perplexity: Perplexity-based data pruning with small reference models, 2024.
\newblock URL \url{https://arxiv.org/abs/2405.20541}.

\bibitem[Arora \& Goyal(2023)Arora and Goyal]{arora2023theory}
Arora, S. and Goyal, A.
\newblock A theory for emergence of complex skills in language models.
\newblock \emph{arXiv preprint ar{X}iv:2307.15936}, 2023.

\bibitem[Austin et~al.(2021)Austin, Odena, Nye, Bosma, Michalewski, Dohan, Jiang, Cai, Terry, Le, and Sutton]{austin2021mbpp}
Austin, J., Odena, A., Nye, M., Bosma, M., Michalewski, H., Dohan, D., Jiang, E., Cai, C., Terry, M., Le, Q., and Sutton, C.
\newblock Program synthesis with large language models, 2021.

\bibitem[Azar et~al.(2023)Azar, Rowland, Piot, Guo, Calandriello, Valko, and Munos]{azar2023ipo}
Azar, M.~G., Rowland, M., Piot, B., Guo, D., Calandriello, D., Valko, M., and Munos, R.
\newblock A general theoretical paradigm to understand learning from human preferences, 2023.

\bibitem[Bhatt et~al.(2024)Bhatt, Chen, Das, Zhang, Truong, Mussmann, Zhu, Bilmes, Du, Jamieson, Ash, and Nowak]{FacilityLocations}
Bhatt, G., Chen, Y., Das, A.~M., Zhang, J., Truong, S.~T., Mussmann, S., Zhu, Y., Bilmes, J., Du, S.~S., Jamieson, K., Ash, J.~T., and Nowak, R.~D.
\newblock An experimental design framework for label-efficient supervised finetuning of large language models, 2024.
\newblock URL \url{https://arxiv.org/abs/2401.06692}.

\bibitem[Bohacek \& Farid(2023)Bohacek and Farid]{bohacek2023nepotistically}
Bohacek, M. and Farid, H.
\newblock Nepotistically trained generative-ai models collapse, 2023.

\bibitem[Briesch et~al.(2023)Briesch, Sobania, and Rothlauf]{briesch2023large}
Briesch, M., Sobania, D., and Rothlauf, F.
\newblock Large language models suffer from their own output: An analysis of the self-consuming training loop, 2023.

\bibitem[Brown et~al.(2020)Brown, Mann, Ryder, Subbiah, Kaplan, Dhariwal, Neelakantan, Shyam, Sastry, Askell, Agarwal, Herbert-Voss, Krueger, Henighan, Child, Ramesh, Ziegler, Wu, Winter, Hesse, Chen, Sigler, Litwin, Gray, Chess, Clark, Berner, McCandlish, Radford, Sutskever, and Amodei]{brown2020languagemodelsfewshotlearners}
Brown, T.~B., Mann, B., Ryder, N., Subbiah, M., Kaplan, J., Dhariwal, P., Neelakantan, A., Shyam, P., Sastry, G., Askell, A., Agarwal, S., Herbert-Voss, A., Krueger, G., Henighan, T., Child, R., Ramesh, A., Ziegler, D.~M., Wu, J., Winter, C., Hesse, C., Chen, M., Sigler, E., Litwin, M., Gray, S., Chess, B., Clark, J., Berner, C., McCandlish, S., Radford, A., Sutskever, I., and Amodei, D.
\newblock Language models are few-shot learners, 2020.
\newblock URL \url{https://arxiv.org/abs/2005.14165}.

\bibitem[Chan et~al.(2024)Chan, Pu, Shanker, Suresh, Jenks, Heyer, and Denton]{chan2024balancingcosteffectivenesssynthetic}
Chan, Y.-C., Pu, G., Shanker, A., Suresh, P., Jenks, P., Heyer, J., and Denton, S.
\newblock Balancing cost and effectiveness of synthetic data generation strategies for llms, 2024.
\newblock URL \url{https://arxiv.org/abs/2409.19759}.

\bibitem[Chen et~al.(2024)Chen, Qadri, Wen, Jain, Kirchenbauer, Zhou, and Goldstein]{chen2024genqa}
Chen, J., Qadri, R., Wen, Y., Jain, N., Kirchenbauer, J., Zhou, T., and Goldstein, T.
\newblock Genqa: Generating millions of instructions from a handful of prompts.
\newblock \emph{arXiv preprint arXiv:2406.10323}, 2024.

\bibitem[Chen et~al.(2023{\natexlab{a}})Chen, Li, Yan, Wang, Gunaratna, Yadav, Tang, Srinivasan, Zhou, Huang, et~al.]{chen2023alpagasus}
Chen, L., Li, S., Yan, J., Wang, H., Gunaratna, K., Yadav, V., Tang, Z., Srinivasan, V., Zhou, T., Huang, H., et~al.
\newblock Alpagasus: Training a better alpaca with fewer data.
\newblock \emph{arXiv preprint arXiv:2307.08701}, 2023{\natexlab{a}}.

\bibitem[Chen et~al.(2021)Chen, Tworek, Jun, Yuan, de~Oliveira~Pinto, Kaplan, Edwards, Burda, Joseph, Brockman, Ray, Puri, Krueger, Petrov, Khlaaf, Sastry, Mishkin, Chan, Gray, Ryder, Pavlov, Power, Kaiser, Bavarian, Winter, Tillet, Such, Cummings, Plappert, Chantzis, Barnes, Herbert-Voss, Guss, Nichol, Paino, Tezak, Tang, Babuschkin, Balaji, Jain, Saunders, Hesse, Carr, Leike, Achiam, Misra, Morikawa, Radford, Knight, Brundage, Murati, Mayer, Welinder, McGrew, Amodei, McCandlish, Sutskever, and Zaremba]{chen2021humaneval}
Chen, M., Tworek, J., Jun, H., Yuan, Q., de~Oliveira~Pinto, H.~P., Kaplan, J., Edwards, H., Burda, Y., Joseph, N., Brockman, G., Ray, A., Puri, R., Krueger, G., Petrov, M., Khlaaf, H., Sastry, G., Mishkin, P., Chan, B., Gray, S., Ryder, N., Pavlov, M., Power, A., Kaiser, L., Bavarian, M., Winter, C., Tillet, P., Such, F.~P., Cummings, D., Plappert, M., Chantzis, F., Barnes, E., Herbert-Voss, A., Guss, W.~H., Nichol, A., Paino, A., Tezak, N., Tang, J., Babuschkin, I., Balaji, S., Jain, S., Saunders, W., Hesse, C., Carr, A.~N., Leike, J., Achiam, J., Misra, V., Morikawa, E., Radford, A., Knight, M., Brundage, M., Murati, M., Mayer, K., Welinder, P., McGrew, B., Amodei, D., McCandlish, S., Sutskever, I., and Zaremba, W.
\newblock Evaluating large language models trained on code, 2021.

\bibitem[Chen et~al.(2023{\natexlab{b}})Chen, Yin, Ku, Lu, Wan, Ma, Xu, Wang, and Xia]{chen2023theoremqatheoremdrivenquestionanswering}
Chen, W., Yin, M., Ku, M., Lu, P., Wan, Y., Ma, X., Xu, J., Wang, X., and Xia, T.
\newblock Theoremqa: A theorem-driven question answering dataset, 2023{\natexlab{b}}.
\newblock URL \url{https://arxiv.org/abs/2305.12524}.

\bibitem[Cobbe et~al.(2021)Cobbe, Kosaraju, Bavarian, Chen, Jun, Kaiser, Plappert, Tworek, Hilton, Nakano, Hesse, and Schulman]{cobbe2021trainingverifierssolvemath}
Cobbe, K., Kosaraju, V., Bavarian, M., Chen, M., Jun, H., Kaiser, L., Plappert, M., Tworek, J., Hilton, J., Nakano, R., Hesse, C., and Schulman, J.
\newblock Training verifiers to solve math word problems, 2021.
\newblock URL \url{https://arxiv.org/abs/2110.14168}.

\bibitem[Cohen-Wang et~al.(2024)Cohen-Wang, Vendrow, and Madry]{cohenwang2024askdistributionshiftpretraining}
Cohen-Wang, B., Vendrow, J., and Madry, A.
\newblock Ask your distribution shift if pre-training is right for you, 2024.
\newblock URL \url{https://arxiv.org/abs/2403.00194}.

\bibitem[Cover \& Thomas(2006)Cover and Thomas]{cover_thomas_information_theory}
Cover, T.~M. and Thomas, J.~A.
\newblock \emph{Elements of Information Theory}.
\newblock Wiley-Interscience, Hoboken, NJ, USA, 2nd edition, 2006.
\newblock ISBN 978-0-471-24195-9.
\newblock URL \url{https://onlinelibrary.wiley.com/doi/book/10.1002/047174882X}.

\bibitem[Cui et~al.(2024)Cui, Yuan, Ding, Yao, He, Zhu, Ni, Xie, Xie, Lin, Liu, and Sun]{cui2024ultrafeedback}
Cui, G., Yuan, L., Ding, N., Yao, G., He, B., Zhu, W., Ni, Y., Xie, G., Xie, R., Lin, Y., Liu, Z., and Sun, M.
\newblock Ultrafeedback: Boosting language models with scaled ai feedback, 2024.
\newblock URL \url{https://arxiv.org/abs/2310.01377}.

\bibitem[Dai et~al.(2025)Dai, Zhang, Ma, and Peng]{dai2025bids}
Dai, Q., Zhang, D., Ma, J.~W., and Peng, H.
\newblock Improving influence-based instruction tuning data selection for balanced learning of diverse capabilities, 2025.
\newblock URL \url{https://arxiv.org/abs/2501.12147}.

\bibitem[Das \& Khetan(2023)Das and Khetan]{das2023deft}
Das, D. and Khetan, V.
\newblock Deft: Data efficient fine-tuning for large language models via unsupervised core-set selection.
\newblock \emph{arXiv preprint arXiv:2310.16776}, 2023.

\bibitem[Databricks(2023)]{Dolly}
Databricks.
\newblock Databricks dolly-15k, 2023.
\newblock URL \url{https://huggingface.co/datasets/databricks/databricks-dolly-15k}.

\bibitem[DeepSeek-AI et~al.(2025)DeepSeek-AI, Guo, Yang, Zhang, Song, Zhang, Xu, Zhu, Ma, Wang, Bi, Zhang, Yu, Wu, Wu, Gou, Shao, Li, Gao, Liu, Xue, Wang, Wu, Feng, Lu, Zhao, Deng, Zhang, Ruan, Dai, Chen, Ji, Li, Lin, Dai, Luo, Hao, Chen, Li, Zhang, Bao, Xu, Wang, Ding, Xin, Gao, Qu, Li, Guo, Li, Wang, Chen, Yuan, Qiu, Li, Cai, Ni, Liang, Chen, Dong, Hu, Gao, Guan, Huang, Yu, Wang, Zhang, Zhao, Wang, Zhang, Xu, Xia, Zhang, Zhang, Tang, Li, Wang, Li, Tian, Huang, Zhang, Wang, Chen, Du, Ge, Zhang, Pan, Wang, Chen, Jin, Chen, Lu, Zhou, Chen, Ye, Wang, Yu, Zhou, Pan, Li, Zhou, Wu, Ye, Yun, Pei, Sun, Wang, Zeng, Zhao, Liu, Liang, Gao, Yu, Zhang, Xiao, An, Liu, Wang, Chen, Nie, Cheng, Liu, Xie, Liu, Yang, Li, Su, Lin, Li, Jin, Shen, Chen, Sun, Wang, Song, Zhou, Wang, Shan, Li, Wang, Wei, Zhang, Xu, Li, Zhao, Sun, Wang, Yu, Zhang, Shi, Xiong, He, Piao, Wang, Tan, Ma, Liu, Guo, Ou, Wang, Gong, Zou, He, Xiong, Luo, You, Liu, Zhou, Zhu, Xu, Huang, Li, Zheng, Zhu, Ma, Tang, Zha, Yan, Ren, Ren, Sha, Fu, Xu, Xie, Zhang,
  Hao, Ma, Yan, Wu, Gu, Zhu, Liu, Li, Xie, Song, Pan, Huang, Xu, Zhang, and Zhang]{deepseekai2025deepseekr1incentivizingreasoningcapability}
DeepSeek-AI, Guo, D., Yang, D., Zhang, H., Song, J., Zhang, R., Xu, R., Zhu, Q., Ma, S., Wang, P., Bi, X., Zhang, X., Yu, X., Wu, Y., Wu, Z.~F., Gou, Z., Shao, Z., Li, Z., Gao, Z., Liu, A., Xue, B., Wang, B., Wu, B., Feng, B., Lu, C., Zhao, C., Deng, C., Zhang, C., Ruan, C., Dai, D., Chen, D., Ji, D., Li, E., Lin, F., Dai, F., Luo, F., Hao, G., Chen, G., Li, G., Zhang, H., Bao, H., Xu, H., Wang, H., Ding, H., Xin, H., Gao, H., Qu, H., Li, H., Guo, J., Li, J., Wang, J., Chen, J., Yuan, J., Qiu, J., Li, J., Cai, J.~L., Ni, J., Liang, J., Chen, J., Dong, K., Hu, K., Gao, K., Guan, K., Huang, K., Yu, K., Wang, L., Zhang, L., Zhao, L., Wang, L., Zhang, L., Xu, L., Xia, L., Zhang, M., Zhang, M., Tang, M., Li, M., Wang, M., Li, M., Tian, N., Huang, P., Zhang, P., Wang, Q., Chen, Q., Du, Q., Ge, R., Zhang, R., Pan, R., Wang, R., Chen, R.~J., Jin, R.~L., Chen, R., Lu, S., Zhou, S., Chen, S., Ye, S., Wang, S., Yu, S., Zhou, S., Pan, S., Li, S.~S., Zhou, S., Wu, S., Ye, S., Yun, T., Pei, T., Sun, T., Wang, T., Zeng, W.,
  Zhao, W., Liu, W., Liang, W., Gao, W., Yu, W., Zhang, W., Xiao, W.~L., An, W., Liu, X., Wang, X., Chen, X., Nie, X., Cheng, X., Liu, X., Xie, X., Liu, X., Yang, X., Li, X., Su, X., Lin, X., Li, X.~Q., Jin, X., Shen, X., Chen, X., Sun, X., Wang, X., Song, X., Zhou, X., Wang, X., Shan, X., Li, Y.~K., Wang, Y.~Q., Wei, Y.~X., Zhang, Y., Xu, Y., Li, Y., Zhao, Y., Sun, Y., Wang, Y., Yu, Y., Zhang, Y., Shi, Y., Xiong, Y., He, Y., Piao, Y., Wang, Y., Tan, Y., Ma, Y., Liu, Y., Guo, Y., Ou, Y., Wang, Y., Gong, Y., Zou, Y., He, Y., Xiong, Y., Luo, Y., You, Y., Liu, Y., Zhou, Y., Zhu, Y.~X., Xu, Y., Huang, Y., Li, Y., Zheng, Y., Zhu, Y., Ma, Y., Tang, Y., Zha, Y., Yan, Y., Ren, Z.~Z., Ren, Z., Sha, Z., Fu, Z., Xu, Z., Xie, Z., Zhang, Z., Hao, Z., Ma, Z., Yan, Z., Wu, Z., Gu, Z., Zhu, Z., Liu, Z., Li, Z., Xie, Z., Song, Z., Pan, Z., Huang, Z., Xu, Z., Zhang, Z., and Zhang, Z.
\newblock Deepseek-r1: Incentivizing reasoning capability in llms via reinforcement learning, 2025.
\newblock URL \url{https://arxiv.org/abs/2501.12948}.

\bibitem[Ding et~al.(2023)Ding, Qin, Yang, Wei, Yang, Su, Hu, Chen, Chan, Chen, et~al.]{ding2023peft}
Ding, N., Qin, Y., Yang, G., Wei, F., Yang, Z., Su, Y., Hu, S., Chen, Y., Chan, C.-M., Chen, W., et~al.
\newblock Parameter-efficient fine-tuning of large-scale pre-trained language models.
\newblock \emph{Nature Machine Intelligence}, 5\penalty0 (3):\penalty0 220--235, 2023.

\bibitem[Dohmatob et~al.(2024)Dohmatob, Feng, Subramonian, and Kempe]{dohmatob2024strongmodelcollapse}
Dohmatob, E., Feng, Y., Subramonian, A., and Kempe, J.
\newblock Strong model collapse, 2024.
\newblock URL \url{https://arxiv.org/abs/2410.04840}.

\bibitem[Dong et~al.(2025)Dong, Dong, Zhang, Sui, and Wei]{dong2025selfboosting}
Dong, Q., Dong, L., Zhang, X., Sui, Z., and Wei, F.
\newblock Self-boosting large language models with synthetic preference data.
\newblock In \emph{The Thirteenth International Conference on Learning Representations}, 2025.
\newblock URL \url{https://openreview.net/forum?id=7visV100Ms}.

\bibitem[Du et~al.(2023)Du, Zong, and Zhang]{du2023mods}
Du, Q., Zong, C., and Zhang, J.
\newblock Mods: Model-oriented data selection for instruction tuning.
\newblock \emph{arXiv preprint arXiv:2311.15653}, 2023.

\bibitem[Dubey et~al.(2024)Dubey, Jauhri, Pandey, Kadian, Al-Dahle, Letman, Mathur, Schelten, Yang, Fan, Goyal, Hartshorn, Yang, Mitra, Sravankumar, Korenev, Hinsvark, Rao, Zhang, Rodriguez, Gregerson, Spataru, Roziere, Biron, Tang, Chern, Caucheteux, Nayak, Bi, Marra, McConnell, Keller, Touret, Wu, Wong, Ferrer, Nikolaidis, Allonsius, Song, Pintz, Livshits, Esiobu, Choudhary, Mahajan, Garcia-Olano, Perino, Hupkes, Lakomkin, AlBadawy, Lobanova, Dinan, Smith, Radenovic, Zhang, Synnaeve, Lee, Anderson, Nail, Mialon, Pang, Cucurell, Nguyen, Korevaar, Xu, Touvron, Zarov, Ibarra, Kloumann, Misra, Evtimov, Copet, Lee, Geffert, Vranes, Park, Mahadeokar, Shah, van~der Linde, Billock, Hong, Lee, Fu, Chi, Huang, Liu, Wang, Yu, Bitton, Spisak, Park, Rocca, Johnstun, Saxe, Jia, Alwala, Upasani, Plawiak, Li, Heafield, Stone, El-Arini, Iyer, Malik, Chiu, Bhalla, Rantala-Yeary, van~der Maaten, Chen, Tan, Jenkins, Martin, Madaan, Malo, Blecher, Landzaat, de~Oliveira, Muzzi, Pasupuleti, Singh, Paluri, Kardas, Oldham, Rita,
  Pavlova, Kambadur, Lewis, Si, Singh, Hassan, Goyal, Torabi, Bashlykov, Bogoychev, Chatterji, Duchenne, Çelebi, Alrassy, Zhang, Li, Vasic, Weng, Bhargava, Dubal, Krishnan, Koura, Xu, He, Dong, Srinivasan, Ganapathy, Calderer, Cabral, Stojnic, Raileanu, Girdhar, Patel, Sauvestre, Polidoro, Sumbaly, Taylor, Silva, Hou, Wang, Hosseini, Chennabasappa, Singh, Bell, Kim, Edunov, Nie, Narang, Raparthy, Shen, Wan, Bhosale, Zhang, Vandenhende, Batra, Whitman, Sootla, Collot, Gururangan, Borodinsky, Herman, Fowler, Sheasha, Georgiou, Scialom, Speckbacher, Mihaylov, Xiao, Karn, Goswami, Gupta, Ramanathan, Kerkez, Gonguet, Do, Vogeti, Petrovic, Chu, Xiong, Fu, Meers, Martinet, Wang, Tan, Xie, Jia, Wang, Goldschlag, Gaur, Babaei, Wen, Song, Zhang, Li, Mao, Coudert, Yan, Chen, Papakipos, Singh, Grattafiori, Jain, Kelsey, Shajnfeld, Gangidi, Victoria, Goldstand, Menon, Sharma, Boesenberg, Vaughan, Baevski, Feinstein, Kallet, Sangani, Yunus, Lupu, Alvarado, Caples, Gu, Ho, Poulton, Ryan, Ramchandani, Franco, Saraf,
  Chowdhury, Gabriel, Bharambe, Eisenman, Yazdan, James, Maurer, Leonhardi, Huang, Loyd, Paola, Paranjape, Liu, Wu, Ni, Hancock, Wasti, Spence, Stojkovic, Gamido, Montalvo, Parker, Burton, Mejia, Wang, Kim, Zhou, Hu, Chu, Cai, Tindal, Feichtenhofer, Civin, Beaty, Kreymer, Li, Wyatt, Adkins, Xu, Testuggine, David, Parikh, Liskovich, Foss, Wang, Le, Holland, Dowling, Jamil, Montgomery, Presani, Hahn, Wood, Brinkman, Arcaute, Dunbar, Smothers, Sun, Kreuk, Tian, Ozgenel, Caggioni, Guzmán, Kanayet, Seide, Florez, Schwarz, Badeer, Swee, Halpern, Thattai, Herman, Sizov, Guangyi, Zhang, Lakshminarayanan, Shojanazeri, Zou, Wang, Zha, Habeeb, Rudolph, Suk, Aspegren, Goldman, Damlaj, Molybog, Tufanov, Veliche, Gat, Weissman, Geboski, Kohli, Asher, Gaya, Marcus, Tang, Chan, Zhen, Reizenstein, Teboul, Zhong, Jin, Yang, Cummings, Carvill, Shepard, McPhie, Torres, Ginsburg, Wang, Wu, U, Saxena, Prasad, Khandelwal, Zand, Matosich, Veeraraghavan, Michelena, Li, Huang, Chawla, Lakhotia, Huang, Chen, Garg, A, Silva, Bell,
  Zhang, Guo, Yu, Moshkovich, Wehrstedt, Khabsa, Avalani, Bhatt, Tsimpoukelli, Mankus, Hasson, Lennie, Reso, Groshev, Naumov, Lathi, Keneally, Seltzer, Valko, Restrepo, Patel, Vyatskov, Samvelyan, Clark, Macey, Wang, Hermoso, Metanat, Rastegari, Bansal, Santhanam, Parks, White, Bawa, Singhal, Egebo, Usunier, Laptev, Dong, Zhang, Cheng, Chernoguz, Hart, Salpekar, Kalinli, Kent, Parekh, Saab, Balaji, Rittner, Bontrager, Roux, Dollar, Zvyagina, Ratanchandani, Yuvraj, Liang, Alao, Rodriguez, Ayub, Murthy, Nayani, Mitra, Li, Hogan, Battey, Wang, Maheswari, Howes, Rinott, Bondu, Datta, Chugh, Hunt, Dhillon, Sidorov, Pan, Verma, Yamamoto, Ramaswamy, Lindsay, Lindsay, Feng, Lin, Zha, Shankar, Zhang, Zhang, Wang, Agarwal, Sajuyigbe, Chintala, Max, Chen, Kehoe, Satterfield, Govindaprasad, Gupta, Cho, Virk, Subramanian, Choudhury, Goldman, Remez, Glaser, Best, Kohler, Robinson, Li, Zhang, Matthews, Chou, Shaked, Vontimitta, Ajayi, Montanez, Mohan, Kumar, Mangla, Albiero, Ionescu, Poenaru, Mihailescu, Ivanov, Li, Wang,
  Jiang, Bouaziz, Constable, Tang, Wang, Wu, Wang, Xia, Wu, Gao, Chen, Hu, Jia, Qi, Li, Zhang, Zhang, Adi, Nam, Yu, Wang, Hao, Qian, He, Rait, DeVito, Rosnbrick, Wen, Yang, and Zhao]{dubey2024llama3herdmodels}
Dubey, A., Jauhri, A., Pandey, A., Kadian, A., Al-Dahle, A., Letman, A., Mathur, A., Schelten, A., Yang, A., Fan, A., Goyal, A., Hartshorn, A., Yang, A., Mitra, A., Sravankumar, A., Korenev, A., Hinsvark, A., Rao, A., Zhang, A., Rodriguez, A., Gregerson, A., Spataru, A., Roziere, B., Biron, B., Tang, B., Chern, B., Caucheteux, C., Nayak, C., Bi, C., Marra, C., McConnell, C., Keller, C., Touret, C., Wu, C., Wong, C., Ferrer, C.~C., Nikolaidis, C., Allonsius, D., Song, D., Pintz, D., Livshits, D., Esiobu, D., Choudhary, D., Mahajan, D., Garcia-Olano, D., Perino, D., Hupkes, D., Lakomkin, E., AlBadawy, E., Lobanova, E., Dinan, E., Smith, E.~M., Radenovic, F., Zhang, F., Synnaeve, G., Lee, G., Anderson, G.~L., Nail, G., Mialon, G., Pang, G., Cucurell, G., Nguyen, H., Korevaar, H., Xu, H., Touvron, H., Zarov, I., Ibarra, I.~A., Kloumann, I., Misra, I., Evtimov, I., Copet, J., Lee, J., Geffert, J., Vranes, J., Park, J., Mahadeokar, J., Shah, J., van~der Linde, J., Billock, J., Hong, J., Lee, J., Fu, J., Chi, J.,
  Huang, J., Liu, J., Wang, J., Yu, J., Bitton, J., Spisak, J., Park, J., Rocca, J., Johnstun, J., Saxe, J., Jia, J., Alwala, K.~V., Upasani, K., Plawiak, K., Li, K., Heafield, K., Stone, K., El-Arini, K., Iyer, K., Malik, K., Chiu, K., Bhalla, K., Rantala-Yeary, L., van~der Maaten, L., Chen, L., Tan, L., Jenkins, L., Martin, L., Madaan, L., Malo, L., Blecher, L., Landzaat, L., de~Oliveira, L., Muzzi, M., Pasupuleti, M., Singh, M., Paluri, M., Kardas, M., Oldham, M., Rita, M., Pavlova, M., Kambadur, M., Lewis, M., Si, M., Singh, M.~K., Hassan, M., Goyal, N., Torabi, N., Bashlykov, N., Bogoychev, N., Chatterji, N., Duchenne, O., Çelebi, O., Alrassy, P., Zhang, P., Li, P., Vasic, P., Weng, P., Bhargava, P., Dubal, P., Krishnan, P., Koura, P.~S., Xu, P., He, Q., Dong, Q., Srinivasan, R., Ganapathy, R., Calderer, R., Cabral, R.~S., Stojnic, R., Raileanu, R., Girdhar, R., Patel, R., Sauvestre, R., Polidoro, R., Sumbaly, R., Taylor, R., Silva, R., Hou, R., Wang, R., Hosseini, S., Chennabasappa, S., Singh, S.,
  Bell, S., Kim, S.~S., Edunov, S., Nie, S., Narang, S., Raparthy, S., Shen, S., Wan, S., Bhosale, S., Zhang, S., Vandenhende, S., Batra, S., Whitman, S., Sootla, S., Collot, S., Gururangan, S., Borodinsky, S., Herman, T., Fowler, T., Sheasha, T., Georgiou, T., Scialom, T., Speckbacher, T., Mihaylov, T., Xiao, T., Karn, U., Goswami, V., Gupta, V., Ramanathan, V., Kerkez, V., Gonguet, V., Do, V., Vogeti, V., Petrovic, V., Chu, W., Xiong, W., Fu, W., Meers, W., Martinet, X., Wang, X., Tan, X.~E., Xie, X., Jia, X., Wang, X., Goldschlag, Y., Gaur, Y., Babaei, Y., Wen, Y., Song, Y., Zhang, Y., Li, Y., Mao, Y., Coudert, Z.~D., Yan, Z., Chen, Z., Papakipos, Z., Singh, A., Grattafiori, A., Jain, A., Kelsey, A., Shajnfeld, A., Gangidi, A., Victoria, A., Goldstand, A., Menon, A., Sharma, A., Boesenberg, A., Vaughan, A., Baevski, A., Feinstein, A., Kallet, A., Sangani, A., Yunus, A., Lupu, A., Alvarado, A., Caples, A., Gu, A., Ho, A., Poulton, A., Ryan, A., Ramchandani, A., Franco, A., Saraf, A., Chowdhury, A., Gabriel,
  A., Bharambe, A., Eisenman, A., Yazdan, A., James, B., Maurer, B., Leonhardi, B., Huang, B., Loyd, B., Paola, B.~D., Paranjape, B., Liu, B., Wu, B., Ni, B., Hancock, B., Wasti, B., Spence, B., Stojkovic, B., Gamido, B., Montalvo, B., Parker, C., Burton, C., Mejia, C., Wang, C., Kim, C., Zhou, C., Hu, C., Chu, C.-H., Cai, C., Tindal, C., Feichtenhofer, C., Civin, D., Beaty, D., Kreymer, D., Li, D., Wyatt, D., Adkins, D., Xu, D., Testuggine, D., David, D., Parikh, D., Liskovich, D., Foss, D., Wang, D., Le, D., Holland, D., Dowling, E., Jamil, E., Montgomery, E., Presani, E., Hahn, E., Wood, E., Brinkman, E., Arcaute, E., Dunbar, E., Smothers, E., Sun, F., Kreuk, F., Tian, F., Ozgenel, F., Caggioni, F., Guzmán, F., Kanayet, F., Seide, F., Florez, G.~M., Schwarz, G., Badeer, G., Swee, G., Halpern, G., Thattai, G., Herman, G., Sizov, G., Guangyi, Zhang, Lakshminarayanan, G., Shojanazeri, H., Zou, H., Wang, H., Zha, H., Habeeb, H., Rudolph, H., Suk, H., Aspegren, H., Goldman, H., Damlaj, I., Molybog, I.,
  Tufanov, I., Veliche, I.-E., Gat, I., Weissman, J., Geboski, J., Kohli, J., Asher, J., Gaya, J.-B., Marcus, J., Tang, J., Chan, J., Zhen, J., Reizenstein, J., Teboul, J., Zhong, J., Jin, J., Yang, J., Cummings, J., Carvill, J., Shepard, J., McPhie, J., Torres, J., Ginsburg, J., Wang, J., Wu, K., U, K.~H., Saxena, K., Prasad, K., Khandelwal, K., Zand, K., Matosich, K., Veeraraghavan, K., Michelena, K., Li, K., Huang, K., Chawla, K., Lakhotia, K., Huang, K., Chen, L., Garg, L., A, L., Silva, L., Bell, L., Zhang, L., Guo, L., Yu, L., Moshkovich, L., Wehrstedt, L., Khabsa, M., Avalani, M., Bhatt, M., Tsimpoukelli, M., Mankus, M., Hasson, M., Lennie, M., Reso, M., Groshev, M., Naumov, M., Lathi, M., Keneally, M., Seltzer, M.~L., Valko, M., Restrepo, M., Patel, M., Vyatskov, M., Samvelyan, M., Clark, M., Macey, M., Wang, M., Hermoso, M.~J., Metanat, M., Rastegari, M., Bansal, M., Santhanam, N., Parks, N., White, N., Bawa, N., Singhal, N., Egebo, N., Usunier, N., Laptev, N.~P., Dong, N., Zhang, N., Cheng, N.,
  Chernoguz, O., Hart, O., Salpekar, O., Kalinli, O., Kent, P., Parekh, P., Saab, P., Balaji, P., Rittner, P., Bontrager, P., Roux, P., Dollar, P., Zvyagina, P., Ratanchandani, P., Yuvraj, P., Liang, Q., Alao, R., Rodriguez, R., Ayub, R., Murthy, R., Nayani, R., Mitra, R., Li, R., Hogan, R., Battey, R., Wang, R., Maheswari, R., Howes, R., Rinott, R., Bondu, S.~J., Datta, S., Chugh, S., Hunt, S., Dhillon, S., Sidorov, S., Pan, S., Verma, S., Yamamoto, S., Ramaswamy, S., Lindsay, S., Lindsay, S., Feng, S., Lin, S., Zha, S.~C., Shankar, S., Zhang, S., Zhang, S., Wang, S., Agarwal, S., Sajuyigbe, S., Chintala, S., Max, S., Chen, S., Kehoe, S., Satterfield, S., Govindaprasad, S., Gupta, S., Cho, S., Virk, S., Subramanian, S., Choudhury, S., Goldman, S., Remez, T., Glaser, T., Best, T., Kohler, T., Robinson, T., Li, T., Zhang, T., Matthews, T., Chou, T., Shaked, T., Vontimitta, V., Ajayi, V., Montanez, V., Mohan, V., Kumar, V.~S., Mangla, V., Albiero, V., Ionescu, V., Poenaru, V., Mihailescu, V.~T., Ivanov, V., Li,
  W., Wang, W., Jiang, W., Bouaziz, W., Constable, W., Tang, X., Wang, X., Wu, X., Wang, X., Xia, X., Wu, X., Gao, X., Chen, Y., Hu, Y., Jia, Y., Qi, Y., Li, Y., Zhang, Y., Zhang, Y., Adi, Y., Nam, Y., Yu, Wang, Hao, Y., Qian, Y., He, Y., Rait, Z., DeVito, Z., Rosnbrick, Z., Wen, Z., Yang, Z., and Zhao, Z.
\newblock The llama 3 herd of models, 2024.
\newblock URL \url{https://arxiv.org/abs/2407.21783}.

\bibitem[Dubois et~al.(2024)Dubois, Galambosi, Liang, and Hashimoto]{dubois2024alpacaevalv2}
Dubois, Y., Galambosi, B., Liang, P., and Hashimoto, T.~B.
\newblock Length-controlled alpacaeval: A simple way to debias automatic evaluators, 2024.
\newblock URL \url{https://arxiv.org/abs/2404.04475}.

\bibitem[Ethayarajh et~al.(2024)Ethayarajh, Xu, Muennighoff, Jurafsky, and Kiela]{ethayarajh2024kto}
Ethayarajh, K., Xu, W., Muennighoff, N., Jurafsky, D., and Kiela, D.
\newblock Kto: Model alignment as prospect theoretic optimization, 2024.

\bibitem[Face(2025)]{openr1}
Face, H.
\newblock Open r1: A fully open reproduction of deepseek-r1, January 2025.
\newblock URL \url{https://github.com/huggingface/open-r1}.

\bibitem[Feldman(2021)]{feldman2021doeslearningrequirememorization}
Feldman, V.
\newblock Does learning require memorization? a short tale about a long tail, 2021.
\newblock URL \url{https://arxiv.org/abs/1906.05271}.

\bibitem[Feng et~al.(2023)Feng, Wang, and Sun]{feng2023citinglargelanguagemodels}
Feng, T., Wang, Z., and Sun, J.
\newblock Citing: Large language models create curriculum for instruction tuning, 2023.
\newblock URL \url{https://arxiv.org/abs/2310.02527}.

\bibitem[Fujimoto et~al.(2018)Fujimoto, Meger, and Precup]{fujimoto2018offpolicy}
Fujimoto, S., Meger, D., and Precup, D.
\newblock Off-policy deep reinforcement learning without exploration.
\newblock In \emph{International Conference on Machine Learning}, 2018.
\newblock URL \url{https://api.semanticscholar.org/CorpusID:54457299}.

\bibitem[Gerstgrasser et~al.(2024)Gerstgrasser, Schaeffer, Dey, Rafailov, Korbak, Sleight, Agrawal, Hughes, Pai, Gromov, Roberts, Yang, Donoho, and Koyejo]{gerstgrasser2024iscollapseinevitable}
Gerstgrasser, M., Schaeffer, R., Dey, A., Rafailov, R., Korbak, T., Sleight, H., Agrawal, R., Hughes, J., Pai, D.~B., Gromov, A., Roberts, D., Yang, D., Donoho, D.~L., and Koyejo, S.
\newblock Is model collapse inevitable? breaking the curse of recursion by accumulating real and synthetic data.
\newblock In \emph{First Conference on Language Modeling}, 2024.
\newblock URL \url{https://openreview.net/forum?id=5B2K4LRgmz}.

\bibitem[Ghosh et~al.(2024)Ghosh, Evuru, Kumar, S, Aneja, Jin, Duraiswami, and Manocha]{ghosh2024closerlooklimitationsinstruction}
Ghosh, S., Evuru, C. K.~R., Kumar, S., S, R., Aneja, D., Jin, Z., Duraiswami, R., and Manocha, D.
\newblock A closer look at the limitations of instruction tuning, 2024.
\newblock URL \url{https://arxiv.org/abs/2402.05119}.

\bibitem[Grattafiori et~al.(2024)Grattafiori, Dubey, Jauhri, Pandey, Kadian, Al-Dahle, Letman, Mathur, Schelten, Vaughan, Yang, Fan, Goyal, Hartshorn, Yang, Mitra, Sravankumar, Korenev, Hinsvark, Rao, Zhang, Rodriguez, Gregerson, Spataru, Roziere, Biron, Tang, Chern, Caucheteux, Nayak, Bi, Marra, McConnell, Keller, Touret, Wu, Wong, Ferrer, Nikolaidis, Allonsius, Song, Pintz, Livshits, Wyatt, Esiobu, Choudhary, Mahajan, Garcia-Olano, Perino, Hupkes, Lakomkin, AlBadawy, Lobanova, Dinan, Smith, Radenovic, Guzmán, Zhang, Synnaeve, Lee, Anderson, Thattai, Nail, Mialon, Pang, Cucurell, Nguyen, Korevaar, Xu, Touvron, Zarov, Ibarra, Kloumann, Misra, Evtimov, Zhang, Copet, Lee, Geffert, Vranes, Park, Mahadeokar, Shah, van~der Linde, Billock, Hong, Lee, Fu, Chi, Huang, Liu, Wang, Yu, Bitton, Spisak, Park, Rocca, Johnstun, Saxe, Jia, Alwala, Prasad, Upasani, Plawiak, Li, Heafield, Stone, El-Arini, Iyer, Malik, Chiu, Bhalla, Lakhotia, Rantala-Yeary, van~der Maaten, Chen, Tan, Jenkins, Martin, Madaan, Malo, Blecher,
  Landzaat, de~Oliveira, Muzzi, Pasupuleti, Singh, Paluri, Kardas, Tsimpoukelli, Oldham, Rita, Pavlova, Kambadur, Lewis, Si, Singh, Hassan, Goyal, Torabi, Bashlykov, Bogoychev, Chatterji, Zhang, Duchenne, Çelebi, Alrassy, Zhang, Li, Vasic, Weng, Bhargava, Dubal, Krishnan, Koura, Xu, He, Dong, Srinivasan, Ganapathy, Calderer, Cabral, Stojnic, Raileanu, Maheswari, Girdhar, Patel, Sauvestre, Polidoro, Sumbaly, Taylor, Silva, Hou, Wang, Hosseini, Chennabasappa, Singh, Bell, Kim, Edunov, Nie, Narang, Raparthy, Shen, Wan, Bhosale, Zhang, Vandenhende, Batra, Whitman, Sootla, Collot, Gururangan, Borodinsky, Herman, Fowler, Sheasha, Georgiou, Scialom, Speckbacher, Mihaylov, Xiao, Karn, Goswami, Gupta, Ramanathan, Kerkez, Gonguet, Do, Vogeti, Albiero, Petrovic, Chu, Xiong, Fu, Meers, Martinet, Wang, Wang, Tan, Xia, Xie, Jia, Wang, Goldschlag, Gaur, Babaei, Wen, Song, Zhang, Li, Mao, Coudert, Yan, Chen, Papakipos, Singh, Srivastava, Jain, Kelsey, Shajnfeld, Gangidi, Victoria, Goldstand, Menon, Sharma, Boesenberg,
  Baevski, Feinstein, Kallet, Sangani, Teo, Yunus, Lupu, Alvarado, Caples, Gu, Ho, Poulton, Ryan, Ramchandani, Dong, Franco, Goyal, Saraf, Chowdhury, Gabriel, Bharambe, Eisenman, Yazdan, James, Maurer, Leonhardi, Huang, Loyd, Paola, Paranjape, Liu, Wu, Ni, Hancock, Wasti, Spence, Stojkovic, Gamido, Montalvo, Parker, Burton, Mejia, Liu, Wang, Kim, Zhou, Hu, Chu, Cai, Tindal, Feichtenhofer, Gao, Civin, Beaty, Kreymer, Li, Adkins, Xu, Testuggine, David, Parikh, Liskovich, Foss, Wang, Le, Holland, Dowling, Jamil, Montgomery, Presani, Hahn, Wood, Le, Brinkman, Arcaute, Dunbar, Smothers, Sun, Kreuk, Tian, Kokkinos, Ozgenel, Caggioni, Kanayet, Seide, Florez, Schwarz, Badeer, Swee, Halpern, Herman, Sizov, Guangyi, Zhang, Lakshminarayanan, Inan, Shojanazeri, Zou, Wang, Zha, Habeeb, Rudolph, Suk, Aspegren, Goldman, Zhan, Damlaj, Molybog, Tufanov, Leontiadis, Veliche, Gat, Weissman, Geboski, Kohli, Lam, Asher, Gaya, Marcus, Tang, Chan, Zhen, Reizenstein, Teboul, Zhong, Jin, Yang, Cummings, Carvill, Shepard, McPhie,
  Torres, Ginsburg, Wang, Wu, U, Saxena, Khandelwal, Zand, Matosich, Veeraraghavan, Michelena, Li, Jagadeesh, Huang, Chawla, Huang, Chen, Garg, A, Silva, Bell, Zhang, Guo, Yu, Moshkovich, Wehrstedt, Khabsa, Avalani, Bhatt, Mankus, Hasson, Lennie, Reso, Groshev, Naumov, Lathi, Keneally, Liu, Seltzer, Valko, Restrepo, Patel, Vyatskov, Samvelyan, Clark, Macey, Wang, Hermoso, Metanat, Rastegari, Bansal, Santhanam, Parks, White, Bawa, Singhal, Egebo, Usunier, Mehta, Laptev, Dong, Cheng, Chernoguz, Hart, Salpekar, Kalinli, Kent, Parekh, Saab, Balaji, Rittner, Bontrager, Roux, Dollar, Zvyagina, Ratanchandani, Yuvraj, Liang, Alao, Rodriguez, Ayub, Murthy, Nayani, Mitra, Parthasarathy, Li, Hogan, Battey, Wang, Howes, Rinott, Mehta, Siby, Bondu, Datta, Chugh, Hunt, Dhillon, Sidorov, Pan, Mahajan, Verma, Yamamoto, Ramaswamy, Lindsay, Lindsay, Feng, Lin, Zha, Patil, Shankar, Zhang, Zhang, Wang, Agarwal, Sajuyigbe, Chintala, Max, Chen, Kehoe, Satterfield, Govindaprasad, Gupta, Deng, Cho, Virk, Subramanian, Choudhury,
  Goldman, Remez, Glaser, Best, Koehler, Robinson, Li, Zhang, Matthews, Chou, Shaked, Vontimitta, Ajayi, Montanez, Mohan, Kumar, Mangla, Ionescu, Poenaru, Mihailescu, Ivanov, Li, Wang, Jiang, Bouaziz, Constable, Tang, Wu, Wang, Wu, Gao, Kleinman, Chen, Hu, Jia, Qi, Li, Zhang, Zhang, Adi, Nam, Yu, Wang, Zhao, Hao, Qian, Li, He, Rait, DeVito, Rosnbrick, Wen, Yang, Zhao, and Ma]{grattafiori2024llama3herdmodels}
Grattafiori, A., Dubey, A., Jauhri, A., Pandey, A., Kadian, A., Al-Dahle, A., Letman, A., Mathur, A., Schelten, A., Vaughan, A., Yang, A., Fan, A., Goyal, A., Hartshorn, A., Yang, A., Mitra, A., Sravankumar, A., Korenev, A., Hinsvark, A., Rao, A., Zhang, A., Rodriguez, A., Gregerson, A., Spataru, A., Roziere, B., Biron, B., Tang, B., Chern, B., Caucheteux, C., Nayak, C., Bi, C., Marra, C., McConnell, C., Keller, C., Touret, C., Wu, C., Wong, C., Ferrer, C.~C., Nikolaidis, C., Allonsius, D., Song, D., Pintz, D., Livshits, D., Wyatt, D., Esiobu, D., Choudhary, D., Mahajan, D., Garcia-Olano, D., Perino, D., Hupkes, D., Lakomkin, E., AlBadawy, E., Lobanova, E., Dinan, E., Smith, E.~M., Radenovic, F., Guzmán, F., Zhang, F., Synnaeve, G., Lee, G., Anderson, G.~L., Thattai, G., Nail, G., Mialon, G., Pang, G., Cucurell, G., Nguyen, H., Korevaar, H., Xu, H., Touvron, H., Zarov, I., Ibarra, I.~A., Kloumann, I., Misra, I., Evtimov, I., Zhang, J., Copet, J., Lee, J., Geffert, J., Vranes, J., Park, J., Mahadeokar, J.,
  Shah, J., van~der Linde, J., Billock, J., Hong, J., Lee, J., Fu, J., Chi, J., Huang, J., Liu, J., Wang, J., Yu, J., Bitton, J., Spisak, J., Park, J., Rocca, J., Johnstun, J., Saxe, J., Jia, J., Alwala, K.~V., Prasad, K., Upasani, K., Plawiak, K., Li, K., Heafield, K., Stone, K., El-Arini, K., Iyer, K., Malik, K., Chiu, K., Bhalla, K., Lakhotia, K., Rantala-Yeary, L., van~der Maaten, L., Chen, L., Tan, L., Jenkins, L., Martin, L., Madaan, L., Malo, L., Blecher, L., Landzaat, L., de~Oliveira, L., Muzzi, M., Pasupuleti, M., Singh, M., Paluri, M., Kardas, M., Tsimpoukelli, M., Oldham, M., Rita, M., Pavlova, M., Kambadur, M., Lewis, M., Si, M., Singh, M.~K., Hassan, M., Goyal, N., Torabi, N., Bashlykov, N., Bogoychev, N., Chatterji, N., Zhang, N., Duchenne, O., Çelebi, O., Alrassy, P., Zhang, P., Li, P., Vasic, P., Weng, P., Bhargava, P., Dubal, P., Krishnan, P., Koura, P.~S., Xu, P., He, Q., Dong, Q., Srinivasan, R., Ganapathy, R., Calderer, R., Cabral, R.~S., Stojnic, R., Raileanu, R., Maheswari, R., Girdhar,
  R., Patel, R., Sauvestre, R., Polidoro, R., Sumbaly, R., Taylor, R., Silva, R., Hou, R., Wang, R., Hosseini, S., Chennabasappa, S., Singh, S., Bell, S., Kim, S.~S., Edunov, S., Nie, S., Narang, S., Raparthy, S., Shen, S., Wan, S., Bhosale, S., Zhang, S., Vandenhende, S., Batra, S., Whitman, S., Sootla, S., Collot, S., Gururangan, S., Borodinsky, S., Herman, T., Fowler, T., Sheasha, T., Georgiou, T., Scialom, T., Speckbacher, T., Mihaylov, T., Xiao, T., Karn, U., Goswami, V., Gupta, V., Ramanathan, V., Kerkez, V., Gonguet, V., Do, V., Vogeti, V., Albiero, V., Petrovic, V., Chu, W., Xiong, W., Fu, W., Meers, W., Martinet, X., Wang, X., Wang, X., Tan, X.~E., Xia, X., Xie, X., Jia, X., Wang, X., Goldschlag, Y., Gaur, Y., Babaei, Y., Wen, Y., Song, Y., Zhang, Y., Li, Y., Mao, Y., Coudert, Z.~D., Yan, Z., Chen, Z., Papakipos, Z., Singh, A., Srivastava, A., Jain, A., Kelsey, A., Shajnfeld, A., Gangidi, A., Victoria, A., Goldstand, A., Menon, A., Sharma, A., Boesenberg, A., Baevski, A., Feinstein, A., Kallet, A.,
  Sangani, A., Teo, A., Yunus, A., Lupu, A., Alvarado, A., Caples, A., Gu, A., Ho, A., Poulton, A., Ryan, A., Ramchandani, A., Dong, A., Franco, A., Goyal, A., Saraf, A., Chowdhury, A., Gabriel, A., Bharambe, A., Eisenman, A., Yazdan, A., James, B., Maurer, B., Leonhardi, B., Huang, B., Loyd, B., Paola, B.~D., Paranjape, B., Liu, B., Wu, B., Ni, B., Hancock, B., Wasti, B., Spence, B., Stojkovic, B., Gamido, B., Montalvo, B., Parker, C., Burton, C., Mejia, C., Liu, C., Wang, C., Kim, C., Zhou, C., Hu, C., Chu, C.-H., Cai, C., Tindal, C., Feichtenhofer, C., Gao, C., Civin, D., Beaty, D., Kreymer, D., Li, D., Adkins, D., Xu, D., Testuggine, D., David, D., Parikh, D., Liskovich, D., Foss, D., Wang, D., Le, D., Holland, D., Dowling, E., Jamil, E., Montgomery, E., Presani, E., Hahn, E., Wood, E., Le, E.-T., Brinkman, E., Arcaute, E., Dunbar, E., Smothers, E., Sun, F., Kreuk, F., Tian, F., Kokkinos, F., Ozgenel, F., Caggioni, F., Kanayet, F., Seide, F., Florez, G.~M., Schwarz, G., Badeer, G., Swee, G., Halpern, G.,
  Herman, G., Sizov, G., Guangyi, Zhang, Lakshminarayanan, G., Inan, H., Shojanazeri, H., Zou, H., Wang, H., Zha, H., Habeeb, H., Rudolph, H., Suk, H., Aspegren, H., Goldman, H., Zhan, H., Damlaj, I., Molybog, I., Tufanov, I., Leontiadis, I., Veliche, I.-E., Gat, I., Weissman, J., Geboski, J., Kohli, J., Lam, J., Asher, J., Gaya, J.-B., Marcus, J., Tang, J., Chan, J., Zhen, J., Reizenstein, J., Teboul, J., Zhong, J., Jin, J., Yang, J., Cummings, J., Carvill, J., Shepard, J., McPhie, J., Torres, J., Ginsburg, J., Wang, J., Wu, K., U, K.~H., Saxena, K., Khandelwal, K., Zand, K., Matosich, K., Veeraraghavan, K., Michelena, K., Li, K., Jagadeesh, K., Huang, K., Chawla, K., Huang, K., Chen, L., Garg, L., A, L., Silva, L., Bell, L., Zhang, L., Guo, L., Yu, L., Moshkovich, L., Wehrstedt, L., Khabsa, M., Avalani, M., Bhatt, M., Mankus, M., Hasson, M., Lennie, M., Reso, M., Groshev, M., Naumov, M., Lathi, M., Keneally, M., Liu, M., Seltzer, M.~L., Valko, M., Restrepo, M., Patel, M., Vyatskov, M., Samvelyan, M., Clark,
  M., Macey, M., Wang, M., Hermoso, M.~J., Metanat, M., Rastegari, M., Bansal, M., Santhanam, N., Parks, N., White, N., Bawa, N., Singhal, N., Egebo, N., Usunier, N., Mehta, N., Laptev, N.~P., Dong, N., Cheng, N., Chernoguz, O., Hart, O., Salpekar, O., Kalinli, O., Kent, P., Parekh, P., Saab, P., Balaji, P., Rittner, P., Bontrager, P., Roux, P., Dollar, P., Zvyagina, P., Ratanchandani, P., Yuvraj, P., Liang, Q., Alao, R., Rodriguez, R., Ayub, R., Murthy, R., Nayani, R., Mitra, R., Parthasarathy, R., Li, R., Hogan, R., Battey, R., Wang, R., Howes, R., Rinott, R., Mehta, S., Siby, S., Bondu, S.~J., Datta, S., Chugh, S., Hunt, S., Dhillon, S., Sidorov, S., Pan, S., Mahajan, S., Verma, S., Yamamoto, S., Ramaswamy, S., Lindsay, S., Lindsay, S., Feng, S., Lin, S., Zha, S.~C., Patil, S., Shankar, S., Zhang, S., Zhang, S., Wang, S., Agarwal, S., Sajuyigbe, S., Chintala, S., Max, S., Chen, S., Kehoe, S., Satterfield, S., Govindaprasad, S., Gupta, S., Deng, S., Cho, S., Virk, S., Subramanian, S., Choudhury, S.,
  Goldman, S., Remez, T., Glaser, T., Best, T., Koehler, T., Robinson, T., Li, T., Zhang, T., Matthews, T., Chou, T., Shaked, T., Vontimitta, V., Ajayi, V., Montanez, V., Mohan, V., Kumar, V.~S., Mangla, V., Ionescu, V., Poenaru, V., Mihailescu, V.~T., Ivanov, V., Li, W., Wang, W., Jiang, W., Bouaziz, W., Constable, W., Tang, X., Wu, X., Wang, X., Wu, X., Gao, X., Kleinman, Y., Chen, Y., Hu, Y., Jia, Y., Qi, Y., Li, Y., Zhang, Y., Zhang, Y., Adi, Y., Nam, Y., Yu, Wang, Zhao, Y., Hao, Y., Qian, Y., Li, Y., He, Y., Rait, Z., DeVito, Z., Rosnbrick, Z., Wen, Z., Yang, Z., Zhao, Z., and Ma, Z.
\newblock The llama 3 herd of models, 2024.
\newblock URL \url{https://arxiv.org/abs/2407.21783}.

\bibitem[Gulcehre et~al.(2023)Gulcehre, Paine, Srinivasan, Konyushkova, Weerts, Sharma, Siddhant, Ahern, Wang, Gu, Macherey, Doucet, Firat, and de~Freitas]{gulcehre2023rest}
Gulcehre, C., Paine, T.~L., Srinivasan, S., Konyushkova, K., Weerts, L., Sharma, A., Siddhant, A., Ahern, A., Wang, M., Gu, C., Macherey, W., Doucet, A., Firat, O., and de~Freitas, N.
\newblock Reinforced self-training (rest) for language modeling, 2023.
\newblock URL \url{https://arxiv.org/abs/2308.08998}.

\bibitem[Guo et~al.(2024{\natexlab{a}})Guo, Zhu, Yang, Xie, Dong, Zhang, Chen, Bi, Wu, Li, Luo, Xiong, and Liang]{guo2024deepseekcoder}
Guo, D., Zhu, Q., Yang, D., Xie, Z., Dong, K., Zhang, W., Chen, G., Bi, X., Wu, Y., Li, Y.~K., Luo, F., Xiong, Y., and Liang, W.
\newblock Deepseek-coder: When the large language model meets programming -- the rise of code intelligence, 2024{\natexlab{a}}.

\bibitem[Guo et~al.(2024{\natexlab{b}})Guo, Zhang, Liu, Liu, Khalman, Llinares, Rame, Mesnard, Zhao, Piot, Ferret, and Blondel]{guo2024onlineaifeedback}
Guo, S., Zhang, B., Liu, T., Liu, T., Khalman, M., Llinares, F., Rame, A., Mesnard, T., Zhao, Y., Piot, B., Ferret, J., and Blondel, M.
\newblock Direct language model alignment from online ai feedback, 2024{\natexlab{b}}.
\newblock URL \url{https://arxiv.org/abs/2402.04792}.

\bibitem[Guo et~al.(2023)Guo, Shang, Vazirgiannis, and Clavel]{guo2023curious}
Guo, Y., Shang, G., Vazirgiannis, M., and Clavel, C.
\newblock The curious decline of linguistic diversity: Training language models on synthetic text, 2023.

\bibitem[Hanawa et~al.(2021)Hanawa, Yokoi, Hara, and Inui]{RDS2}
Hanawa, K., Yokoi, S., Hara, S., and Inui, K.
\newblock Evaluation of similarity-based explanations, 2021.
\newblock URL \url{https://arxiv.org/abs/2006.04528}.

\bibitem[Hataya et~al.(2023)Hataya, Bao, and Arai]{Hataya_2023_ICCV}
Hataya, R., Bao, H., and Arai, H.
\newblock Will large-scale generative models corrupt future datasets?
\newblock In \emph{Proceedings of the IEEE/CVF International Conference on Computer Vision (ICCV)}, pp.\  20555--20565, October 2023.

\bibitem[He et~al.(2023)He, Chen, and Zhu]{he2023preservingpretrained}
He, G., Chen, J., and Zhu, J.
\newblock Preserving pre-trained features helps calibrate fine-tuned language models.
\newblock In \emph{The Eleventh International Conference on Learning Representations}, 2023.
\newblock URL \url{https://openreview.net/forum?id=NI7StoWHJPT}.

\bibitem[Hendrycks et~al.(2021{\natexlab{a}})Hendrycks, Burns, Basart, Zou, Mazeika, Song, and Steinhardt]{hendrycks2021mmlu}
Hendrycks, D., Burns, C., Basart, S., Zou, A., Mazeika, M., Song, D., and Steinhardt, J.
\newblock Measuring massive multitask language understanding, 2021{\natexlab{a}}.
\newblock URL \url{https://arxiv.org/abs/2009.03300}.

\bibitem[Hendrycks et~al.(2021{\natexlab{b}})Hendrycks, Burns, Kadavath, Arora, Basart, Tang, Song, and Steinhardt]{hendrycks2021measuringmathematicalproblemsolving}
Hendrycks, D., Burns, C., Kadavath, S., Arora, A., Basart, S., Tang, E., Song, D., and Steinhardt, J.
\newblock Measuring mathematical problem solving with the math dataset, 2021{\natexlab{b}}.
\newblock URL \url{https://arxiv.org/abs/2103.03874}.

\bibitem[Herel \& Mikolov(2024)Herel and Mikolov]{herel2024collapseselftrainedlanguagemodel}
Herel, D. and Mikolov, T.
\newblock Collapse of self-trained language models, 2024.
\newblock URL \url{https://arxiv.org/1abs/2404.02305}.

\bibitem[Huang et~al.(2024)Huang, Piqueres, Rasul, Schmid, Vila, and Tunstall]{open_hermes_preferences}
Huang, S.~C., Piqueres, A., Rasul, K., Schmid, P., Vila, D., and Tunstall, L.
\newblock Open hermes preferences.
\newblock \url{https://huggingface.co/datasets/argilla/OpenHermesPreferences}, 2024.

\bibitem[HuggingFace-H4(2024)]{huggingface2024openhermes}
HuggingFace-H4.
\newblock Openhermes-2.5-preferences-v0-deduped, 2024.
\newblock URL \url{https://huggingface.co/datasets/HuggingFaceH4/OpenHermes-2.5-preferences-v0-deduped}.

\bibitem[Hui et~al.(2024)Hui, Yang, Cui, Yang, Liu, Zhang, Liu, Zhang, Yu, Lu, Dang, Fan, Zhang, Yang, Men, Huang, Zheng, Miao, Quan, Feng, Ren, Ren, Zhou, and Lin]{hui2024qwen25}
Hui, B., Yang, J., Cui, Z., Yang, J., Liu, D., Zhang, L., Liu, T., Zhang, J., Yu, B., Lu, K., Dang, K., Fan, Y., Zhang, Y., Yang, A., Men, R., Huang, F., Zheng, B., Miao, Y., Quan, S., Feng, Y., Ren, X., Ren, X., Zhou, J., and Lin, J.
\newblock Qwen2.5-coder technical report, 2024.
\newblock URL \url{https://arxiv.org/abs/2409.12186}.

\bibitem[ichi Amari(2016)]{information_geometry_applications}
ichi Amari, S.
\newblock \emph{Information Geometry and Its Applications}, volume 194 of \emph{Applied Mathematical Sciences}.
\newblock Springer, Tokyo, Japan, 1st edition, 2016.
\newblock ISBN 978-4-431-55977-3.
\newblock \doi{10.1007/978-4-431-55978-0}.
\newblock URL \url{https://doi.org/10.1007/978-4-431-55978-0}.

\bibitem[Ivison et~al.(2025)Ivison, Zhang, Brahman, Koh, and Dasigi]{ivison2025largescaledataselectioninstruction}
Ivison, H., Zhang, M., Brahman, F., Koh, P.~W., and Dasigi, P.
\newblock Large-scale data selection for instruction tuning, 2025.
\newblock URL \url{https://arxiv.org/abs/2503.01807}.

\bibitem[Jiang et~al.(2023)Jiang, Sablayrolles, Mensch, Bamford, Chaplot, de~las Casas, Bressand, Lengyel, Lample, Saulnier, Lavaud, Lachaux, Stock, Scao, Lavril, Wang, Lacroix, and Sayed]{jiang2023mistral7b}
Jiang, A.~Q., Sablayrolles, A., Mensch, A., Bamford, C., Chaplot, D.~S., de~las Casas, D., Bressand, F., Lengyel, G., Lample, G., Saulnier, L., Lavaud, L.~R., Lachaux, M.-A., Stock, P., Scao, T.~L., Lavril, T., Wang, T., Lacroix, T., and Sayed, W.~E.
\newblock Mistral 7b, 2023.
\newblock URL \url{https://arxiv.org/abs/2310.06825}.

\bibitem[Jiang et~al.(2024)Jiang, Sablayrolles, Roux, Mensch, Savary, Bamford, Chaplot, de~las Casas, Hanna, Bressand, Lengyel, Bour, Lample, Lavaud, Saulnier, Lachaux, Stock, Subramanian, Yang, Antoniak, Scao, Gervet, Lavril, Wang, Lacroix, and Sayed]{jiang2024mixtralexperts}
Jiang, A.~Q., Sablayrolles, A., Roux, A., Mensch, A., Savary, B., Bamford, C., Chaplot, D.~S., de~las Casas, D., Hanna, E.~B., Bressand, F., Lengyel, G., Bour, G., Lample, G., Lavaud, L.~R., Saulnier, L., Lachaux, M.-A., Stock, P., Subramanian, S., Yang, S., Antoniak, S., Scao, T.~L., Gervet, T., Lavril, T., Wang, T., Lacroix, T., and Sayed, W.~E.
\newblock Mixtral of experts, 2024.
\newblock URL \url{https://arxiv.org/abs/2401.04088}.

\bibitem[Jiang \& Li(2016)Jiang and Li]{jiang2016doublyrobustoffpolicyvalue}
Jiang, N. and Li, L.
\newblock Doubly robust off-policy value evaluation for reinforcement learning, 2016.
\newblock URL \url{https://arxiv.org/abs/1511.03722}.

\bibitem[Kang et~al.(2024)Kang, Just, Sun, Jahagirdar, Zhang, Du, Sahu, and Jia]{kang2024getmoreforless}
Kang, F., Just, H.~A., Sun, Y., Jahagirdar, H., Zhang, Y., Du, R., Sahu, A.~K., and Jia, R.
\newblock Get more for less: Principled data selection for warming up fine-tuning in llms.
\newblock \emph{arXiv preprint arXiv:2405.02774}, 2024.

\bibitem[K\"{o}pf et~al.(2023)K\"{o}pf, Kilcher, von R\"{u}tte, Anagnostidis, Tam, Stevens, Barhoum, Nguyen, Stanley, Nagyfi, ES, Suri, Glushkov, Dantuluri, Maguire, Schuhmann, Nguyen, and Mattick]{openassist}
K\"{o}pf, A., Kilcher, Y., von R\"{u}tte, D., Anagnostidis, S., Tam, Z.~R., Stevens, K., Barhoum, A., Nguyen, D., Stanley, O., Nagyfi, R., ES, S., Suri, S., Glushkov, D., Dantuluri, A., Maguire, A., Schuhmann, C., Nguyen, H., and Mattick, A.
\newblock Openassistant conversations - democratizing large language model alignment.
\newblock In Oh, A., Naumann, T., Globerson, A., Saenko, K., Hardt, M., and Levine, S. (eds.), \emph{Advances in Neural Information Processing Systems}, volume~36, pp.\  47669--47681. Curran Associates, Inc., 2023.
\newblock URL \url{https://proceedings.neurips.cc/paper_files/paper/2023/file/949f0f8f32267d297c2d4e3ee10a2e7e-Paper-Datasets_and_Benchmarks.pdf}.

\bibitem[Kotha et~al.(2024)Kotha, Springer, and Raghunathan]{kotha2024understanding}
Kotha, S., Springer, J.~M., and Raghunathan, A.
\newblock Understanding catastrophic forgetting in language models via implicit inference.
\newblock In \emph{The Twelfth International Conference on Learning Representations}, 2024.
\newblock URL \url{https://openreview.net/forum?id=VrHiF2hsrm}.

\bibitem[Kumar et~al.(2019)Kumar, Fu, Soh, Tucker, and Levine]{kumar2019offpolicy}
Kumar, A., Fu, J., Soh, M., Tucker, G., and Levine, S.
\newblock Stabilizing off-policy q-learning via bootstrapping error reduction.
\newblock In Wallach, H., Larochelle, H., Beygelzimer, A., d\textquotesingle Alch\'{e}-Buc, F., Fox, E., and Garnett, R. (eds.), \emph{Advances in Neural Information Processing Systems}, volume~32. Curran Associates, Inc., 2019.
\newblock URL \url{https://proceedings.neurips.cc/paper_files/paper/2019/file/c2073ffa77b5357a498057413bb09d3a-Paper.pdf}.

\bibitem[Kumar et~al.(2022)Kumar, Raghunathan, Jones, Ma, and Liang]{kumar2022finetuningdistortpretrainedfeatures}
Kumar, A., Raghunathan, A., Jones, R., Ma, T., and Liang, P.
\newblock Fine-tuning can distort pretrained features and underperform out-of-distribution, 2022.
\newblock URL \url{https://arxiv.org/abs/2202.10054}.

\bibitem[Kung et~al.(2023)Kung, Yin, Wu, Chang, and Peng]{kung2023ait}
Kung, P.-N., Yin, F., Wu, D., Chang, K.-W., and Peng, N.
\newblock Active instruction tuning: Improving cross-task generalization by training on prompt sensitive tasks, 2023.
\newblock URL \url{https://arxiv.org/abs/2311.00288}.

\bibitem[Lambert et~al.(2024)Lambert, Morrison, Pyatkin, Huang, Ivison, Brahman, Miranda, Liu, Dziri, Lyu, Gu, Malik, Graf, Hwang, Yang, Bras, Tafjord, Wilhelm, Soldaini, Smith, Wang, Dasigi, and Hajishirzi]{lambert2024tulu3}
Lambert, N., Morrison, J., Pyatkin, V., Huang, S., Ivison, H., Brahman, F., Miranda, L. J.~V., Liu, A., Dziri, N., Lyu, S., Gu, Y., Malik, S., Graf, V., Hwang, J.~D., Yang, J., Bras, R.~L., Tafjord, O., Wilhelm, C., Soldaini, L., Smith, N.~A., Wang, Y., Dasigi, P., and Hajishirzi, H.
\newblock Tulu 3: Pushing frontiers in open language model post-training, 2024.
\newblock URL \url{https://arxiv.org/abs/2411.15124}.

\bibitem[LeBrun et~al.(2021)LeBrun, Sordoni, and O'Donnell]{lebrun2021evaluating}
LeBrun, B., Sordoni, A., and O'Donnell, T.~J.
\newblock Evaluating distributional distortion in neural language modeling.
\newblock In \emph{International Conference on Learning Representations}, 2021.

\bibitem[Lee et~al.(2024)Lee, Cho, and Yoo]{lee2024instructiontuninghumancurriculum}
Lee, B.~W., Cho, H., and Yoo, K.~M.
\newblock Instruction tuning with human curriculum, 2024.
\newblock URL \url{https://arxiv.org/abs/2310.09518}.

\bibitem[Lee et~al.(2025)Lee, Roy, Xu, Raiman, Shoeybi, Catanzaro, and Ping]{lee2025nvemb}
Lee, C., Roy, R., Xu, M., Raiman, J., Shoeybi, M., Catanzaro, B., and Ping, W.
\newblock Nv-embed: Improved techniques for training llms as generalist embedding models, 2025.
\newblock URL \url{https://arxiv.org/abs/2405.17428}.

\bibitem[Li et~al.(2024{\natexlab{a}})Li, Chen, Chen, He, Gu, and Zhou]{li2024selective}
Li, M., Chen, L., Chen, J., He, S., Gu, J., and Zhou, T.
\newblock Selective reflection-tuning: Student-selected data recycling for {LLM} instruction-tuning.
\newblock In Ku, L.-W., Martins, A., and Srikumar, V. (eds.), \emph{Findings of the Association for Computational Linguistics: ACL 2024}, pp.\  16189--16211, Bangkok, Thailand, August 2024{\natexlab{a}}. Association for Computational Linguistics.
\newblock \doi{10.18653/v1/2024.findings-acl.958}.
\newblock URL \url{https://aclanthology.org/2024.findings-acl.958}.

\bibitem[Li et~al.(2024{\natexlab{b}})Li, Zhang, He, Li, Zhao, Wang, Cheng, and Zhou]{li2024superfilteringweaktostrongdatafiltering}
Li, M., Zhang, Y., He, S., Li, Z., Zhao, H., Wang, J., Cheng, N., and Zhou, T.
\newblock Superfiltering: Weak-to-strong data filtering for fast instruction-tuning, 2024{\natexlab{b}}.
\newblock URL \url{https://arxiv.org/abs/2402.00530}.

\bibitem[Li et~al.(2024{\natexlab{c}})Li, Zhang, Li, Chen, Chen, Cheng, Wang, Zhou, and Xiao]{li2024quantitytoqulality}
Li, M., Zhang, Y., Li, Z., Chen, J., Chen, L., Cheng, N., Wang, J., Zhou, T., and Xiao, J.
\newblock From quantity to quality: Boosting {LLM} performance with self-guided data selection for instruction tuning.
\newblock In Duh, K., Gomez, H., and Bethard, S. (eds.), \emph{Proceedings of the 2024 Conference of the North American Chapter of the Association for Computational Linguistics: Human Language Technologies (Volume 1: Long Papers)}, pp.\  7602--7635, Mexico City, Mexico, June 2024{\natexlab{c}}. Association for Computational Linguistics.
\newblock \doi{10.18653/v1/2024.naacl-long.421}.
\newblock URL \url{https://aclanthology.org/2024.naacl-long.421}.

\bibitem[Li et~al.(2024{\natexlab{d}})Li, Cui, Zhao, Kong, and Bi]{li2024gsmplus}
Li, Q., Cui, L., Zhao, X., Kong, L., and Bi, W.
\newblock Gsm-plus: A comprehensive benchmark for evaluating the robustness of llms as mathematical problem solvers, 2024{\natexlab{d}}.
\newblock URL \url{https://arxiv.org/abs/2402.19255}.

\bibitem[Li et~al.(2022)Li, Choi, Chung, Kushman, Schrittwieser, Leblond, Eccles, Keeling, Gimeno, Dal~Lago, Hubert, Choy, de~Masson~d’Autume, Babuschkin, Chen, Huang, Welbl, Gowal, Cherepanov, Molloy, Mankowitz, Sutherland~Robson, Kohli, de~Freitas, Kavukcuoglu, and Vinyals]{alphacode}
Li, Y., Choi, D., Chung, J., Kushman, N., Schrittwieser, J., Leblond, R., Eccles, T., Keeling, J., Gimeno, F., Dal~Lago, A., Hubert, T., Choy, P., de~Masson~d’Autume, C., Babuschkin, I., Chen, X., Huang, P.-S., Welbl, J., Gowal, S., Cherepanov, A., Molloy, J., Mankowitz, D.~J., Sutherland~Robson, E., Kohli, P., de~Freitas, N., Kavukcuoglu, K., and Vinyals, O.
\newblock Competition-level code generation with alphacode.
\newblock \emph{Science}, 378\penalty0 (6624):\penalty0 1092–1097, December 2022.
\newblock ISSN 1095-9203.
\newblock \doi{10.1126/science.abq1158}.
\newblock URL \url{http://dx.doi.org/10.1126/science.abq1158}.

\bibitem[Li et~al.(2023)Li, Lin, Zhang, Fu, Chen, Lou, and Chen]{li2023making}
Li, Y., Lin, Z., Zhang, S., Fu, Q., Chen, B., Lou, J.-G., and Chen, W.
\newblock Making language models better reasoners with step-aware verifier.
\newblock In Rogers, A., Boyd-Graber, J., and Okazaki, N. (eds.), \emph{Proceedings of the 61st Annual Meeting of the Association for Computational Linguistics (Volume 1: Long Papers)}, pp.\  5315--5333, Toronto, Canada, July 2023. Association for Computational Linguistics.
\newblock \doi{10.18653/v1/2023.acl-long.291}.
\newblock URL \url{https://aclanthology.org/2023.acl-long.291}.

\bibitem[Li et~al.(2024{\natexlab{e}})Li, Hua, Vu, Zhan, Qu, and Haffari]{li2024scar}
Li, Z., Hua, Y., Vu, T.-T., Zhan, H., Qu, L., and Haffari, G.
\newblock Scar: Efficient instruction-tuning for large language models via style consistency-aware response ranking, 2024{\natexlab{e}}.
\newblock URL \url{https://arxiv.org/abs/2406.10882}.

\bibitem[Lian et~al.(2023{\natexlab{a}})Lian, Goodson, Pentland, Cook, Vong, and "Teknium"]{OpenOrca}
Lian, W., Goodson, B., Pentland, E., Cook, A., Vong, C., and "Teknium".
\newblock Openorca: An open dataset of gpt augmented flan reasoning traces.
\newblock \url{https://https://huggingface.co/Open-Orca/OpenOrca}, 2023{\natexlab{a}}.

\bibitem[Lian et~al.(2023{\natexlab{b}})Lian, Wang, Goodson, Pentland, Cook, Vong, and "Teknium"]{slimorca}
Lian, W., Wang, G., Goodson, B., Pentland, E., Cook, A., Vong, C., and "Teknium".
\newblock Slimorca: An open dataset of gpt-4 augmented flan reasoning traces, with verification, 2023{\natexlab{b}}.
\newblock URL \url{https://https://huggingface.co/Open-Orca/SlimOrca}.

\bibitem[Lightman et~al.(2023)Lightman, Kosaraju, Burda, Edwards, Baker, Lee, Leike, Schulman, Sutskever, and Cobbe]{lightman2023letsverifystepstep}
Lightman, H., Kosaraju, V., Burda, Y., Edwards, H., Baker, B., Lee, T., Leike, J., Schulman, J., Sutskever, I., and Cobbe, K.
\newblock Let's verify step by step, 2023.
\newblock URL \url{https://arxiv.org/abs/2305.20050}.

\bibitem[Liu et~al.(2024{\natexlab{a}})Liu, Wei, Liu, Si, Zhang, Rao, Zheng, Peng, Yang, Zhou, et~al.]{liu2024best}
Liu, R., Wei, J., Liu, F., Si, C., Zhang, Y., Rao, J., Zheng, S., Peng, D., Yang, D., Zhou, D., et~al.
\newblock Best practices and lessons learned on synthetic data for language models.
\newblock \emph{arXiv preprint arXiv:2404.07503}, 2024{\natexlab{a}}.

\bibitem[Liu et~al.(2024{\natexlab{b}})Liu, Zeng, He, Jiang, and He]{liu2024deita}
Liu, W., Zeng, W., He, K., Jiang, Y., and He, J.
\newblock What makes good data for alignment? a comprehensive study of automatic data selection in instruction tuning, 2024{\natexlab{b}}.
\newblock URL \url{https://arxiv.org/abs/2312.15685}.

\bibitem[Liu et~al.(2024{\natexlab{c}})Liu, Liu, Shi, Cheng, Huang, and Lu]{liu2024letslearnstepstep}
Liu, Y., Liu, J., Shi, X., Cheng, Q., Huang, Y., and Lu, W.
\newblock Let's learn step by step: Enhancing in-context learning ability with curriculum learning, 2024{\natexlab{c}}.
\newblock URL \url{https://arxiv.org/abs/2402.10738}.

\bibitem[Liu et~al.(2024{\natexlab{d}})Liu, Lu, Zhang, Liu, Guo, Yang, Blanchet, and Wang]{liu2024provablymitigatingoveroptimizationrlhf}
Liu, Z., Lu, M., Zhang, S., Liu, B., Guo, H., Yang, Y., Blanchet, J., and Wang, Z.
\newblock Provably mitigating overoptimization in rlhf: Your sft loss is implicitly an adversarial regularizer, 2024{\natexlab{d}}.
\newblock URL \url{https://arxiv.org/abs/2405.16436}.

\bibitem[Longpre et~al.(2023)Longpre, Hou, Vu, Webson, Chung, Tay, Zhou, Le, Zoph, Wei, and Roberts]{longpre2023flan}
Longpre, S., Hou, L., Vu, T., Webson, A., Chung, H.~W., Tay, Y., Zhou, D., Le, Q.~V., Zoph, B., Wei, J., and Roberts, A.
\newblock The flan collection: Designing data and methods for effective instruction tuning, 2023.
\newblock URL \url{https://arxiv.org/abs/2301.13688}.

\bibitem[Luo et~al.(2024)Luo, Liu, Liu, Phatale, Guo, Lara, Li, Shu, Zhu, Meng, Sun, and Rastogi]{luo2024improvemathematicalreasoninglanguage}
Luo, L., Liu, Y., Liu, R., Phatale, S., Guo, M., Lara, H., Li, Y., Shu, L., Zhu, Y., Meng, L., Sun, J., and Rastogi, A.
\newblock Improve mathematical reasoning in language models by automated process supervision, 2024.
\newblock URL \url{https://arxiv.org/abs/2406.06592}.

\bibitem[Luo et~al.(2025)Luo, Yang, Meng, Li, Zhou, and Zhang]{luo2025empiricalstudycatastrophicforgetting}
Luo, Y., Yang, Z., Meng, F., Li, Y., Zhou, J., and Zhang, Y.
\newblock An empirical study of catastrophic forgetting in large language models during continual fine-tuning, 2025.
\newblock URL \url{https://arxiv.org/abs/2308.08747}.

\bibitem[Marion et~al.(2023{\natexlab{a}})Marion, Üstün, Pozzobon, Wang, Fadaee, and Hooker]{PPL1}
Marion, M., Üstün, A., Pozzobon, L., Wang, A., Fadaee, M., and Hooker, S.
\newblock When less is more: Investigating data pruning for pretraining llms at scale, 2023{\natexlab{a}}.
\newblock URL \url{https://arxiv.org/abs/2309.04564}.

\bibitem[Marion et~al.(2023{\natexlab{b}})Marion, Üstün, Pozzobon, Wang, Fadaee, and Hooker]{Uncertainty1}
Marion, M., Üstün, A., Pozzobon, L., Wang, A., Fadaee, M., and Hooker, S.
\newblock When less is more: Investigating data pruning for pretraining llms at scale, 2023{\natexlab{b}}.
\newblock URL \url{https://arxiv.org/abs/2309.04564}.

\bibitem[Martínez et~al.(2023{\natexlab{a}})Martínez, Watson, Reviriego, Hernández, Juarez, and Sarkar]{martínez2023combining}
Martínez, G., Watson, L., Reviriego, P., Hernández, J.~A., Juarez, M., and Sarkar, R.
\newblock Combining generative artificial intelligence (ai) and the internet: Heading towards evolution or degradation?
\newblock \emph{arXiv preprint arxiv: 2303.01255}, 2023{\natexlab{a}}.

\bibitem[Martínez et~al.(2023{\natexlab{b}})Martínez, Watson, Reviriego, Hernández, Juarez, and Sarkar]{martínez2023understanding}
Martínez, G., Watson, L., Reviriego, P., Hernández, J.~A., Juarez, M., and Sarkar, R.
\newblock Towards understanding the interplay of generative artificial intelligence and the internet.
\newblock \emph{arXiv preprint arxiv: 2306.06130}, 2023{\natexlab{b}}.

\bibitem[Mekala et~al.(2024)Mekala, Nguyen, and Shang]{mekala2024smaller}
Mekala, D., Nguyen, A., and Shang, J.
\newblock Smaller language models are capable of selecting instruction-tuning training data for larger language models.
\newblock \emph{arXiv preprint arXiv:2402.10430}, 2024.

\bibitem[Miao et~al.(2024)Miao, Gao, Quan, Lin, Zan, Liu, Yang, Liu, and Deng]{miao2024aligningcodellmsdirectpreference}
Miao, Y., Gao, B., Quan, S., Lin, J., Zan, D., Liu, J., Yang, J., Liu, T., and Deng, Z.
\newblock Aligning codellms with direct preference optimization, 2024.
\newblock URL \url{https://arxiv.org/abs/2410.18585}.

\bibitem[Mindermann et~al.(2022)Mindermann, Brauner, Razzak, Sharma, Kirsch, Xu, Höltgen, Gomez, Morisot, Farquhar, and Gal]{Learnability1}
Mindermann, S., Brauner, J., Razzak, M., Sharma, M., Kirsch, A., Xu, W., Höltgen, B., Gomez, A.~N., Morisot, A., Farquhar, S., and Gal, Y.
\newblock Prioritized training on points that are learnable, worth learning, and not yet learnt, 2022.
\newblock URL \url{https://arxiv.org/abs/2206.07137}.

\bibitem[MistralAI(2024{\natexlab{a}})]{mistralai2024codestral22b}
MistralAI.
\newblock Codestral-22b-v0.1.
\newblock \url{https://huggingface.co/mistralai/Codestral-22B-v0.1}, 2024{\natexlab{a}}.
\newblock Accessed: 2024-12-13.

\bibitem[MistralAI(2024{\natexlab{b}})]{mistralai2024mistralsmallinstruct}
MistralAI.
\newblock Mistral-small-instruct-2409.
\newblock \url{https://huggingface.co/mistralai/Mistral-Small-Instruct-2409}, 2024{\natexlab{b}}.
\newblock Accessed: 2024-12-13.

\bibitem[MistralAI(2024{\natexlab{c}})]{mistralai_codestral_2024}
MistralAI.
\newblock Codestral-22b-v0.1, 2024{\natexlab{c}}.
\newblock URL \url{https://huggingface.co/mistralai/Codestral-22B-v0.1}.
\newblock Accessed: 2024-09-28.

\bibitem[Mobahi et~al.(2020)Mobahi, Farajtabar, and Bartlett]{mobahi2020selfdistillationamplifiesregularizationhilbert}
Mobahi, H., Farajtabar, M., and Bartlett, P.~L.
\newblock Self-distillation amplifies regularization in hilbert space, 2020.
\newblock URL \url{https://arxiv.org/abs/2002.05715}.

\bibitem[OLMo et~al.(2025)OLMo, Walsh, Soldaini, Groeneveld, Lo, Arora, Bhagia, Gu, Huang, Jordan, Lambert, Schwenk, Tafjord, Anderson, Atkinson, Brahman, Clark, Dasigi, Dziri, Guerquin, Ivison, Koh, Liu, Malik, Merrill, Miranda, Morrison, Murray, Nam, Pyatkin, Rangapur, Schmitz, Skjonsberg, Wadden, Wilhelm, Wilson, Zettlemoyer, Farhadi, Smith, and Hajishirzi]{olmo2025}
OLMo, T., Walsh, P., Soldaini, L., Groeneveld, D., Lo, K., Arora, S., Bhagia, A., Gu, Y., Huang, S., Jordan, M., Lambert, N., Schwenk, D., Tafjord, O., Anderson, T., Atkinson, D., Brahman, F., Clark, C., Dasigi, P., Dziri, N., Guerquin, M., Ivison, H., Koh, P.~W., Liu, J., Malik, S., Merrill, W., Miranda, L. J.~V., Morrison, J., Murray, T., Nam, C., Pyatkin, V., Rangapur, A., Schmitz, M., Skjonsberg, S., Wadden, D., Wilhelm, C., Wilson, M., Zettlemoyer, L., Farhadi, A., Smith, N.~A., and Hajishirzi, H.
\newblock 2 olmo 2 furious, 2025.
\newblock URL \url{https://arxiv.org/abs/2501.00656}.

\bibitem[Ouyang et~al.(2022)Ouyang, Wu, Jiang, Almeida, Wainwright, Mishkin, Zhang, Agarwal, Slama, Ray, et~al.]{ouyang2022training}
Ouyang, L., Wu, J., Jiang, X., Almeida, D., Wainwright, C., Mishkin, P., Zhang, C., Agarwal, S., Slama, K., Ray, A., et~al.
\newblock Training language models to follow instructions with human feedback.
\newblock \emph{Advances in Neural Information Processing Systems}, 35:\penalty0 27730--27744, 2022.

\bibitem[Padmakumar \& He(2024)Padmakumar and He]{padmakumar2024writing}
Padmakumar, V. and He, H.
\newblock Does writing with language models reduce content diversity?
\newblock In \emph{International Conference on Learning Representations (ICLR)}, 2024.

\bibitem[Pan et~al.(2024)Pan, Zhang, Pan, Pi, Wang, and Zhang]{pan2024scalebioscalablebileveloptimization}
Pan, R., Zhang, J., Pan, X., Pi, R., Wang, X., and Zhang, T.
\newblock Scalebio: Scalable bilevel optimization for llm data reweighting, 2024.
\newblock URL \url{https://arxiv.org/abs/2406.19976}.

\bibitem[Parkar et~al.(2024)Parkar, Kim, Park, and Kang]{parkar2024selectllm}
Parkar, R.~S., Kim, J., Park, J.~I., and Kang, D.
\newblock Selectllm: Can llms select important instructions to annotate?
\newblock \emph{arXiv preprint arXiv:2401.16553}, 2024.

\bibitem[Paul et~al.(2023)Paul, Ganguli, and Dziugaite]{GraNd}
Paul, M., Ganguli, S., and Dziugaite, G.~K.
\newblock Deep learning on a data diet: Finding important examples early in training, 2023.
\newblock URL \url{https://arxiv.org/abs/2107.07075}.

\bibitem[Peng et~al.(2019)Peng, Kumar, Zhang, and Levine]{peng2019offpolicy}
Peng, X.~B., Kumar, A., Zhang, G., and Levine, S.
\newblock Advantage-weighted regression: Simple and scalable off-policy reinforcement learning, 2019.
\newblock URL \url{https://arxiv.org/abs/1910.00177}.

\bibitem[Qin et~al.(2024)Qin, Yang, Guo, Li, Shao, Shi, Xu, Gu, Li, and Sun]{Coreset}
Qin, Y., Yang, Y., Guo, P., Li, G., Shao, H., Shi, Y., Xu, Z., Gu, Y., Li, K., and Sun, X.
\newblock Unleashing the power of data tsunami: A comprehensive survey on data assessment and selection for instruction tuning of language models, 2024.
\newblock URL \url{https://arxiv.org/abs/2408.02085}.

\bibitem[Rafailov et~al.(2023)Rafailov, Sharma, Mitchell, Ermon, Manning, and Finn]{rafailov2023dpo}
Rafailov, R., Sharma, A., Mitchell, E., Ermon, S., Manning, C.~D., and Finn, C.
\newblock Direct preference optimization: Your language model is secretly a reward model, 2023.

\bibitem[Rubin et~al.(2022)Rubin, Herzig, and Berant]{RDS1}
Rubin, O., Herzig, J., and Berant, J.
\newblock Learning to retrieve prompts for in-context learning, 2022.
\newblock URL \url{https://arxiv.org/abs/2112.08633}.

\bibitem[Saunshi et~al.(2021)Saunshi, Malladi, and Arora]{saunshi2021mathematicalexplorationlanguagemodels}
Saunshi, N., Malladi, S., and Arora, S.
\newblock A mathematical exploration of why language models help solve downstream tasks, 2021.
\newblock URL \url{https://arxiv.org/abs/2010.03648}.

\bibitem[Setlur et~al.(2024)Setlur, Garg, Geng, Garg, Smith, and Kumar]{setlur2024rlincorrectsyntheticdata}
Setlur, A., Garg, S., Geng, X., Garg, N., Smith, V., and Kumar, A.
\newblock Rl on incorrect synthetic data scales the efficiency of llm math reasoning by eight-fold, 2024.
\newblock URL \url{https://arxiv.org/abs/2406.14532}.

\bibitem[Shi et~al.(2023)Shi, Dadashi, Chi, Castro, and Geist]{shi2023offlinereinforcementlearningonpolicy}
Shi, L., Dadashi, R., Chi, Y., Castro, P.~S., and Geist, M.
\newblock Offline reinforcement learning with on-policy q-function regularization, 2023.
\newblock URL \url{https://arxiv.org/abs/2307.13824}.

\bibitem[Shumailov et~al.(2023)Shumailov, Shumaylov, Zhao, Gal, Papernot, and Anderson]{shumailov2023curse}
Shumailov, I., Shumaylov, Z., Zhao, Y., Gal, Y., Papernot, N., and Anderson, R.
\newblock The curse of recursion: Training on generated data makes models forget.
\newblock \emph{arXiv preprint arxiv:2305.17493}, 2023.

\bibitem[Shumailov et~al.(2024)Shumailov, Shumaylov, Zhao, Papernot, Anderson, and Gal]{collapse_when_trained_recursive}
Shumailov, I., Shumaylov, Z., Zhao, Y., Papernot, N., Anderson, R.~J., and Gal, Y.
\newblock Ai models collapse when trained on recursively generated data.
\newblock \emph{Nat.}, 631\penalty0 (8022):\penalty0 755--759, July 2024.
\newblock URL \url{https://doi.org/10.1038/s41586-024-07566-y}.

\bibitem[Sun et~al.(2023)Sun, Shen, Zhou, Zhang, Chen, Cox, Yang, and Gan]{sun2024principle}
Sun, Z., Shen, Y., Zhou, Q., Zhang, H., Chen, Z., Cox, D., Yang, Y., and Gan, C.
\newblock Principle-driven self-alignment of language models from scratch with minimal human supervision.
\newblock \emph{Advances in Neural Information Processing Systems}, 36, 2023.

\bibitem[Suzgun et~al.(2022)Suzgun, Scales, Schärli, Gehrmann, Tay, Chung, Chowdhery, Le, Chi, Zhou, and Wei]{suzgun2022bbh}
Suzgun, M., Scales, N., Schärli, N., Gehrmann, S., Tay, Y., Chung, H.~W., Chowdhery, A., Le, Q.~V., Chi, E.~H., Zhou, D., and Wei, J.
\newblock Challenging big-bench tasks and whether chain-of-thought can solve them, 2022.
\newblock URL \url{https://arxiv.org/abs/2210.09261}.

\bibitem[Tajwar et~al.(2024)Tajwar, Singh, Sharma, Rafailov, Schneider, Xie, Ermon, Finn, and Kumar]{tajwar2024shoulduse}
Tajwar, F., Singh, A., Sharma, A., Rafailov, R., Schneider, J., Xie, T., Ermon, S., Finn, C., and Kumar, A.
\newblock Preference fine-tuning of llms should leverage suboptimal, on-policy data.
\newblock \emph{arXiv preprint arXiv:2404.14367}, 2024.

\bibitem[Tang \& Abbeel(2010)Tang and Abbeel]{tang2010importancesampling}
Tang, J. and Abbeel, P.
\newblock On a connection between importance sampling and the likelihood ratio policy gradient.
\newblock pp.\  1000--1008, 01 2010.

\bibitem[Tang et~al.(2024{\natexlab{a}})Tang, Guo, Zheng, Calandriello, Cao, Tarassov, Munos, Pires, Valko, Cheng, et~al.]{tang2024understandingperformancegap}
Tang, Y., Guo, D.~Z., Zheng, Z., Calandriello, D., Cao, Y., Tarassov, E., Munos, R., Pires, B.~{\'A}., Valko, M., Cheng, Y., et~al.
\newblock Understanding the performance gap between online and offline alignment algorithms.
\newblock \emph{arXiv preprint arXiv:2405.08448}, 2024{\natexlab{a}}.

\bibitem[Tang et~al.(2024{\natexlab{b}})Tang, Guo, Zheng, Calandriello, Cao, Tarassov, Munos, Ávila Pires, Valko, Cheng, and Dabney]{tang2024understandingperformancegaponline}
Tang, Y., Guo, D.~Z., Zheng, Z., Calandriello, D., Cao, Y., Tarassov, E., Munos, R., Ávila Pires, B., Valko, M., Cheng, Y., and Dabney, W.
\newblock Understanding the performance gap between online and offline alignment algorithms, 2024{\natexlab{b}}.
\newblock URL \url{https://arxiv.org/abs/2405.08448}.

\bibitem[Taori et~al.(2023)Taori, Gulrajani, Zhang, Dubois, Li, Guestrin, Liang, and Hashimoto]{alpaca}
Taori, R., Gulrajani, I., Zhang, T., Dubois, Y., Li, X., Guestrin, C., Liang, P., and Hashimoto, T.~B.
\newblock Stanford alpaca: An instruction-following llama model.
\newblock \url{https://github.com/tatsu-lab/stanford_alpaca}, 2023.

\bibitem[Team(2025)]{openthoughts}
Team, O.~T.
\newblock {Open Thoughts}, January 2025.

\bibitem[Team(2023)]{vicuna2023}
Team, V.~D.
\newblock Vicuna llm: An open-source chatbot developed by fine-tuning the llama model on user-shared conversations, achieving performance comparable to other advanced chatbots.
\newblock \url{https://lmsys.org/blog/2023-03-30-vicuna/}, 2023.
\newblock Accessed: 2025-01-27.

\bibitem[Teknium(2023)]{OpenHermes}
Teknium.
\newblock Openhermes 2.5: An open dataset of synthetic data for generalist llm assistants, 2023.
\newblock URL \url{https://huggingface.co/datasets/teknium/OpenHermes-2.5}.

\bibitem[Wang et~al.(2024{\natexlab{a}})Wang, Li, Shao, Xu, Dai, Li, Chen, Wu, and Sui]{wang2024mathshepherd}
Wang, P., Li, L., Shao, Z., Xu, R.~X., Dai, D., Li, Y., Chen, D., Wu, Y., and Sui, Z.
\newblock Math-shepherd: Verify and reinforce llms step-by-step without human annotations, 2024{\natexlab{a}}.

\bibitem[Wang et~al.(2024{\natexlab{b}})Wang, Shen, Guo, Stallone, Kim, Golland, and Panda]{wang2024diversitymeasurementsubsetselection}
Wang, P., Shen, Y., Guo, Z., Stallone, M., Kim, Y., Golland, P., and Panda, R.
\newblock Diversity measurement and subset selection for instruction tuning datasets, 2024{\natexlab{b}}.
\newblock URL \url{https://arxiv.org/abs/2402.02318}.

\bibitem[Wang et~al.(2023)Wang, Kordi, Mishra, Liu, Smith, Khashabi, and Hajishirzi]{wang-etal-2023-self-instruct}
Wang, Y., Kordi, Y., Mishra, S., Liu, A., Smith, N.~A., Khashabi, D., and Hajishirzi, H.
\newblock Self-instruct: Aligning language models with self-generated instructions.
\newblock In \emph{Proceedings of the 61st Annual Meeting of the Association for Computational Linguistics (Volume 1: Long Papers)}, pp.\  13484--13508, Toronto, Canada, 2023. Association for Computational Linguistics.

\bibitem[Wang et~al.(2021)Wang, Novikov, Zolna, Springenberg, Reed, Shahriari, Siegel, Merel, Gulcehre, Heess, and de~Freitas]{wang2021criticregularizedregression}
Wang, Z., Novikov, A., Zolna, K., Springenberg, J.~T., Reed, S., Shahriari, B., Siegel, N., Merel, J., Gulcehre, C., Heess, N., and de~Freitas, N.
\newblock Critic regularized regression, 2021.
\newblock URL \url{https://arxiv.org/abs/2006.15134}.

\bibitem[Wei et~al.(2022)Wei, Wang, Schuurmans, Bosma, brian ichter, Xia, Chi, Le, and Zhou]{wei2022chainofthought}
Wei, J., Wang, X., Schuurmans, D., Bosma, M., brian ichter, Xia, F., Chi, E.~H., Le, Q.~V., and Zhou, D.
\newblock Chain of thought prompting elicits reasoning in large language models.
\newblock In Oh, A.~H., Agarwal, A., Belgrave, D., and Cho, K. (eds.), \emph{Advances in Neural Information Processing Systems}, 2022.
\newblock URL \url{https://openreview.net/forum?id=_VjQlMeSB_J}.

\bibitem[Wei et~al.(2023)Wei, Wang, Liu, Ding, and Zhang]{wei2023magicoder}
Wei, Y., Wang, Z., Liu, J., Ding, Y., and Zhang, L.
\newblock Magicoder: Source code is all you need, 2023.

\bibitem[Wu et~al.(2024)Wu, Meng, and Chen]{wu2024curriculumlearningqualitydrivendata}
Wu, B., Meng, F., and Chen, L.
\newblock Curriculum learning with quality-driven data selection, 2024.
\newblock URL \url{https://arxiv.org/abs/2407.00102}.

\bibitem[Xia et~al.(2023)Xia, Artetxe, Zhou, Lin, Pasunuru, Chen, Zettlemoyer, and Stoyanov]{S2LRef}
Xia, M., Artetxe, M., Zhou, C., Lin, X.~V., Pasunuru, R., Chen, D., Zettlemoyer, L., and Stoyanov, V.
\newblock Training trajectories of language models across scales, 2023.
\newblock URL \url{https://arxiv.org/abs/2212.09803}.

\bibitem[Xia et~al.(2024)Xia, Malladi, Gururangan, Arora, and Chen]{xia2024less}
Xia, M., Malladi, S., Gururangan, S., Arora, S., and Chen, D.
\newblock {LESS}: Selecting influential data for targeted instruction tuning.
\newblock In \emph{International Conference on Machine Learning (ICML)}, 2024.

\bibitem[Xiong et~al.(2024)Xiong, Dong, Ye, Wang, Zhong, Ji, Jiang, and Zhang]{xiong2024iterative}
Xiong, W., Dong, H., Ye, C., Wang, Z., Zhong, H., Ji, H., Jiang, N., and Zhang, T.
\newblock Iterative preference learning from human feedback: Bridging theory and practice for {RLHF} under {KL}-constraint.
\newblock In \emph{Forty-first International Conference on Machine Learning}, 2024.
\newblock URL \url{https://openreview.net/forum?id=c1AKcA6ry1}.

\bibitem[Xu et~al.(2024{\natexlab{a}})Xu, Sun, Zheng, Geng, Zhao, Feng, Tao, Lin, and Jiang]{xu2024wizardlm}
Xu, C., Sun, Q., Zheng, K., Geng, X., Zhao, P., Feng, J., Tao, C., Lin, Q., and Jiang, D.
\newblock Wizard{LM}: Empowering large pre-trained language models to follow complex instructions.
\newblock In \emph{The Twelfth International Conference on Learning Representations}, 2024{\natexlab{a}}.
\newblock URL \url{https://openreview.net/forum?id=CfXh93NDgH}.

\bibitem[Xu et~al.(2023)Xu, Yao, Huang, Qi, Wang, Gu, and Sundaresan]{xu2023rethinkinginstructionqualitylift}
Xu, Y., Yao, Y., Huang, Y., Qi, M., Wang, M., Gu, B., and Sundaresan, N.
\newblock Rethinking the instruction quality: Lift is what you need, 2023.
\newblock URL \url{https://arxiv.org/abs/2312.11508}.

\bibitem[Xu et~al.(2024{\natexlab{b}})Xu, Jiang, Niu, Deng, Poovendran, Choi, and Lin]{xu2024magpie}
Xu, Z., Jiang, F., Niu, L., Deng, Y., Poovendran, R., Choi, Y., and Lin, B.~Y.
\newblock Magpie: Alignment data synthesis from scratch by prompting aligned llms with nothing.
\newblock \emph{arXiv preprint arXiv:2406.08464}, 2024{\natexlab{b}}.

\bibitem[Xu et~al.(2024{\natexlab{c}})Xu, Jiang, Niu, Lin, and Poovendran]{xu2024strongermodelsstrongerteachers}
Xu, Z., Jiang, F., Niu, L., Lin, B.~Y., and Poovendran, R.
\newblock Stronger models are not stronger teachers for instruction tuning, 2024{\natexlab{c}}.
\newblock URL \url{https://arxiv.org/abs/2411.07133}.

\bibitem[Yan et~al.(2025)Yan, Li, Hu, Wang, Cui, Qu, Cheng, and Zhang]{luffy}
Yan, J., Li, Y., Hu, Z., Wang, Z., Cui, G., Qu, X., Cheng, Y., and Zhang, Y.
\newblock Learning to reason under off-policy guidance, 2025.
\newblock URL \url{https://arxiv.org/abs/2504.14945}.

\bibitem[Yang et~al.(2024{\natexlab{a}})Yang, Yang, Hui, Zheng, Yu, Zhou, Li, Li, Liu, Huang, Dong, Wei, Lin, Tang, Wang, Yang, Tu, Zhang, Ma, Yang, Xu, Zhou, Bai, He, Lin, Dang, Lu, Chen, Yang, Li, Xue, Ni, Zhang, Wang, Peng, Men, Gao, Lin, Wang, Bai, Tan, Zhu, Li, Liu, Ge, Deng, Zhou, Ren, Zhang, Wei, Ren, Liu, Fan, Yao, Zhang, Wan, Chu, Liu, Cui, Zhang, Guo, and Fan]{yang2024qwen2technicalreport}
Yang, A., Yang, B., Hui, B., Zheng, B., Yu, B., Zhou, C., Li, C., Li, C., Liu, D., Huang, F., Dong, G., Wei, H., Lin, H., Tang, J., Wang, J., Yang, J., Tu, J., Zhang, J., Ma, J., Yang, J., Xu, J., Zhou, J., Bai, J., He, J., Lin, J., Dang, K., Lu, K., Chen, K., Yang, K., Li, M., Xue, M., Ni, N., Zhang, P., Wang, P., Peng, R., Men, R., Gao, R., Lin, R., Wang, S., Bai, S., Tan, S., Zhu, T., Li, T., Liu, T., Ge, W., Deng, X., Zhou, X., Ren, X., Zhang, X., Wei, X., Ren, X., Liu, X., Fan, Y., Yao, Y., Zhang, Y., Wan, Y., Chu, Y., Liu, Y., Cui, Z., Zhang, Z., Guo, Z., and Fan, Z.
\newblock Qwen2 technical report, 2024{\natexlab{a}}.
\newblock URL \url{https://arxiv.org/abs/2407.10671}.

\bibitem[Yang et~al.(2024{\natexlab{b}})Yang, Mishra, Chiang, and Mirzasoleiman]{yang2024s2l}
Yang, Y., Mishra, S., Chiang, J.~N., and Mirzasoleiman, B.
\newblock Smalltolarge (s2l): Scalable data selection for fine-tuning large language models by summarizing training trajectories of small models, 2024{\natexlab{b}}.
\newblock URL \url{https://arxiv.org/abs/2403.07384}.

\bibitem[Yang et~al.(2024{\natexlab{c}})Yang, Mishra, Chiang, and Mirzasoleiman]{yang2024smalltolarge}
Yang, Y., Mishra, S., Chiang, J.~N., and Mirzasoleiman, B.
\newblock Smalltolarge (s2l): Scalable data selection for fine-tuning large language models by summarizing training trajectories of small models.
\newblock \emph{arXiv preprint arXiv:2403.07384}, 2024{\natexlab{c}}.

\bibitem[Yang et~al.(2024{\natexlab{d}})Yang, Pang, Feng, Wang, Chen, Zhu, and Liu]{yang2024selfdistillationbridgesdistributiongap}
Yang, Z., Pang, T., Feng, H., Wang, H., Chen, W., Zhu, M., and Liu, Q.
\newblock Self-distillation bridges distribution gap in language model fine-tuning, 2024{\natexlab{d}}.
\newblock URL \url{https://arxiv.org/abs/2402.13669}.

\bibitem[Yin \& Rush(2024)Yin and Rush]{yin2024computeconstraineddataselection}
Yin, J.~O. and Rush, A.~M.
\newblock Compute-constrained data selection, 2024.
\newblock URL \url{https://arxiv.org/abs/2410.16208}.

\bibitem[Yu et~al.(2024)Yu, Jiang, Shi, YU, Liu, Zhang, Kwok, Li, Weller, and Liu]{yu2024metamath}
Yu, L., Jiang, W., Shi, H., YU, J., Liu, Z., Zhang, Y., Kwok, J., Li, Z., Weller, A., and Liu, W.
\newblock Metamath: Bootstrap your own mathematical questions for large language models.
\newblock In \emph{The Twelfth International Conference on Learning Representations}, 2024.
\newblock URL \url{https://openreview.net/forum?id=N8N0hgNDRt}.

\bibitem[Yuan et~al.(2024{\natexlab{a}})Yuan, Chen, Ji, and Gu]{yuan2024spin}
Yuan, H., Chen, Z., Ji, K., and Gu, Q.
\newblock Self-play fine-tuning of diffusion models for text-to-image generation.
\newblock In Globerson, A., Mackey, L., Belgrave, D., Fan, A., Paquet, U., Tomczak, J., and Zhang, C. (eds.), \emph{Advances in Neural Information Processing Systems}, volume~37, pp.\  73366--73398. Curran Associates, Inc., 2024{\natexlab{a}}.
\newblock URL \url{https://proceedings.neurips.cc/paper_files/paper/2024/file/860c1c657deafe09f64c013c2888bd7b-Paper-Conference.pdf}.

\bibitem[Yuan et~al.(2024{\natexlab{b}})Yuan, Cui, Wang, Ding, Wang, Deng, Shan, Chen, Xie, Lin, Liu, Zhou, Peng, Liu, and Sun]{yuan2024eurus}
Yuan, L., Cui, G., Wang, H., Ding, N., Wang, X., Deng, J., Shan, B., Chen, H., Xie, R., Lin, Y., Liu, Z., Zhou, B., Peng, H., Liu, Z., and Sun, M.
\newblock Advancing llm reasoning generalists with preference trees, 2024{\natexlab{b}}.

\bibitem[Yue et~al.(2023)Yue, Qu, Zhang, Fu, Huang, Sun, Su, and Chen]{yue2023mammoth}
Yue, X., Qu, X., Zhang, G., Fu, Y., Huang, W., Sun, H., Su, Y., and Chen, W.
\newblock Mammoth: Building math generalist models through hybrid instruction tuning, 2023.
\newblock URL \url{https://arxiv.org/abs/2309.05653}.

\bibitem[Zeng et~al.(2024)Zeng, Xu, Zhao, Lou, and Chen]{zeng2024automatic}
Zeng, W., Xu, C., Zhao, Y., Lou, J.-G., and Chen, W.
\newblock Automatic instruction evolving for large language models.
\newblock \emph{arXiv preprint arXiv:2406.00770}, 2024.

\bibitem[Zhang et~al.(2024{\natexlab{a}})Zhang, Diao, Zou, and Peng]{zhang2024textbfplumimprovingcodelms}
Zhang, D., Diao, S., Zou, X., and Peng, H.
\newblock $\textbf{PLUM}$: Improving code lms with execution-guided on-policy preference learning driven by synthetic test cases, 2024{\natexlab{a}}.
\newblock URL \url{https://arxiv.org/abs/2406.06887}.

\bibitem[Zhang et~al.(2024{\natexlab{b}})Zhang, Qin, Pi, Zhang, Pan, and Zhang]{TAGCOS}
Zhang, J., Qin, Y., Pi, R., Zhang, W., Pan, R., and Zhang, T.
\newblock Tagcos: Task-agnostic gradient clustered coreset selection for instruction tuning data, 2024{\natexlab{b}}.
\newblock URL \url{https://arxiv.org/abs/2407.15235}.

\bibitem[Zhang et~al.(2024{\natexlab{c}})Zhang, Yu, Sharma, Zhong, Liu, Yang, Wang, Hassan, and Wang]{zhang2024selfexploringlanguagemodelsactive}
Zhang, S., Yu, D., Sharma, H., Zhong, H., Liu, Z., Yang, Z., Wang, S., Hassan, H., and Wang, Z.
\newblock Self-exploring language models: Active preference elicitation for online alignment, 2024{\natexlab{c}}.
\newblock URL \url{https://arxiv.org/abs/2405.19332}.

\bibitem[Zhang et~al.(2024{\natexlab{d}})Zhang, Schwarzschild, Carlini, Kolter, and Ippolito]{zhang2024forcing}
Zhang, Y., Schwarzschild, A., Carlini, N., Kolter, J.~Z., and Ippolito, D.
\newblock Forcing diffuse distributions out of language models.
\newblock In \emph{First Conference on Language Modeling}, 2024{\natexlab{d}}.
\newblock URL \url{https://openreview.net/forum?id=9JY1QLVFPZ}.

\bibitem[Zhao et~al.(2021)Zhao, Mopuri, and Bilen]{zhao2021datasetcondensationgradientmatching}
Zhao, B., Mopuri, K.~R., and Bilen, H.
\newblock Dataset condensation with gradient matching, 2021.
\newblock URL \url{https://arxiv.org/abs/2006.05929}.

\bibitem[Zhao et~al.(2024{\natexlab{a}})Zhao, Ren, Hessel, Cardie, Choi, and Deng]{zhao2024wildchat}
Zhao, W., Ren, X., Hessel, J., Cardie, C., Choi, Y., and Deng, Y.
\newblock Wildchat: 1m chat{GPT} interaction logs in the wild.
\newblock In \emph{The Twelfth International Conference on Learning Representations}, 2024{\natexlab{a}}.
\newblock URL \url{https://openreview.net/forum?id=Bl8u7ZRlbM}.

\bibitem[Zhao et~al.(2024{\natexlab{b}})Zhao, Yu, Hui, Yu, Huang, Li, and Zhang]{zhao2024preliminarystudyintrinsicrelationship}
Zhao, Y., Yu, B., Hui, B., Yu, H., Huang, F., Li, Y., and Zhang, N.~L.
\newblock A preliminary study of the intrinsic relationship between complexity and alignment, 2024{\natexlab{b}}.
\newblock URL \url{https://arxiv.org/abs/2308.05696}.

\bibitem[Zheng et~al.(2024)Zheng, Chiang, Sheng, Li, Zhuang, Wu, Zhuang, Li, Lin, Xing, Gonzalez, Stoica, and Zhang]{zheng2024lmsyschatm}
Zheng, L., Chiang, W.-L., Sheng, Y., Li, T., Zhuang, S., Wu, Z., Zhuang, Y., Li, Z., Lin, Z., Xing, E., Gonzalez, J.~E., Stoica, I., and Zhang, H.
\newblock {LMSYS}-chat-1m: A large-scale real-world {LLM} conversation dataset.
\newblock In \emph{The Twelfth International Conference on Learning Representations}, 2024.
\newblock URL \url{https://openreview.net/forum?id=BOfDKxfwt0}.

\bibitem[Zhong et~al.(2024)Zhong, Liu, Chen, Hu, Zhu, Liu, Jin, and Zhang]{zhong2024distservedisaggregatingprefilldecoding}
Zhong, Y., Liu, S., Chen, J., Hu, J., Zhu, Y., Liu, X., Jin, X., and Zhang, H.
\newblock Distserve: Disaggregating prefill and decoding for goodput-optimized large language model serving, 2024.
\newblock URL \url{https://arxiv.org/abs/2401.09670}.

\bibitem[Zhou et~al.(2023)Zhou, Liu, Xu, Iyer, Sun, Mao, Ma, Efrat, Yu, Yu, Zhang, Ghosh, Lewis, Zettlemoyer, and Levy]{zhou2023lima}
Zhou, C., Liu, P., Xu, P., Iyer, S., Sun, J., Mao, Y., Ma, X., Efrat, A., Yu, P., Yu, L., Zhang, S., Ghosh, G., Lewis, M., Zettlemoyer, L., and Levy, O.
\newblock Lima: Less is more for alignment, 2023.
\newblock URL \url{https://arxiv.org/abs/2305.11206}.

\bibitem[Zhou et~al.(2024{\natexlab{a}})Zhou, Liu, Ma, Yuan, Liu, You, and Yang]{Learnability2}
Zhou, H., Liu, T., Ma, Q., Yuan, J., Liu, P., You, Y., and Yang, H.
\newblock Gauging learnability in supervised fine-tuning data, 2024{\natexlab{a}}.
\newblock URL \url{https://openreview.net/forum?id=KpC3dPumJj}.

\bibitem[Zhou et~al.(2024{\natexlab{b}})Zhou, Liu, Ma, Zhang, Yuan, Liu, You, and Yang]{Learnability3}
Zhou, H., Liu, T., Ma, Q., Zhang, Y., Yuan, J., Liu, P., You, Y., and Yang, H.
\newblock Davir: Data selection via implicit reward for large language models, 2024{\natexlab{b}}.
\newblock URL \url{https://arxiv.org/abs/2310.13008}.

\bibitem[Zhou et~al.(2024{\natexlab{c}})Zhou, Agrawal, Zhang, Indurthi, Zhao, Song, Xu, and Zhu]{zhou2024wpoenhancingrlhfweighted}
Zhou, W., Agrawal, R., Zhang, S., Indurthi, S.~R., Zhao, S., Song, K., Xu, S., and Zhu, C.
\newblock Wpo: Enhancing rlhf with weighted preference optimization, 2024{\natexlab{c}}.
\newblock URL \url{https://arxiv.org/abs/2406.11827}.

\bibitem[Zhou et~al.(2024{\natexlab{d}})Zhou, Xu, Liu, An, Ai, and Huang]{zhou2024explorespuriouscorrelationsconcept}
Zhou, Y., Xu, P., Liu, X., An, B., Ai, W., and Huang, F.
\newblock Explore spurious correlations at the concept level in language models for text classification, 2024{\natexlab{d}}.
\newblock URL \url{https://arxiv.org/abs/2311.08648}.

\bibitem[Zhou et~al.(2024{\natexlab{e}})Zhou, Ning, Hong, Fu, Xu, Li, Lou, Wang, Yuan, Li, et~al.]{zhou2024survey}
Zhou, Z., Ning, X., Hong, K., Fu, T., Xu, J., Li, S., Lou, Y., Wang, L., Yuan, Z., Li, X., et~al.
\newblock A survey on efficient inference for large language models.
\newblock \emph{arXiv preprint arXiv:2404.14294}, 2024{\natexlab{e}}.

\bibitem[Zhuang et~al.(2023)Zhuang, LEI, Liu, Wang, and Guo]{zhuang2023bpo}
Zhuang, Z., LEI, K., Liu, J., Wang, D., and Guo, Y.
\newblock Behavior proximal policy optimization.
\newblock In \emph{The Eleventh International Conference on Learning Representations}, 2023.
\newblock URL \url{https://openreview.net/forum?id=3c13LptpIph}.

\end{thebibliography}
\bibliographystyle{icml2025}
\newpage
\newpage
\section*{NeurIPS Paper Checklist}

\begin{enumerate}

\item {\bf Claims}
    \item[] Question: Do the main claims made in the abstract and introduction accurately reflect the paper's contributions and scope?
    \item[] Answer: \answerYes{} 
    \item[] Justification: See experiments in Section 4 and 5.

\item {\bf Limitations}
    \item[] Question: Does the paper discuss the limitations of the work performed by the authors?
    \item[] Answer: \answerYes{} 
    \item[] Justification: See section 5.6, where we show pursuing in-distribution answers in the wrong way can lead to performance degradations.

\item {\bf Theory assumptions and proofs}
    \item[] Question: For each theoretical result, does the paper provide the full set of assumptions and a complete (and correct) proof?
    \item[] Answer: \answerNA{} 
    \item[] Justification: We do not have theoretical results.

    \item {\bf Experimental result reproducibility}
    \item[] Question: Does the paper fully disclose all the information needed to reproduce the main experimental results of the paper to the extent that it affects the main claims and/or conclusions of the paper (regardless of whether the code and data are provided or not)?
    \item[] Answer: \answerYes{} 
    \item[] Justification: See appendices.

\item {\bf Open access to data and code}
    \item[] Question: Does the paper provide open access to the data and code, with sufficient instructions to faithfully reproduce the main experimental results, as described in supplemental material?
    \item[] Answer: \answerYes{} 
    \item[] Justification: We use publicly available datasets and models. Our method only requires computing the normalized probability of training data, which can be easily done with any open-sourced machine learning codebase.

\item {\bf Experimental setting/details}
    \item[] Question: Does the paper specify all the training and test details (e.g., data splits, hyperparameters, how they were chosen, type of optimizer, etc.) necessary to understand the results?
    \item[] Answer: \answerYes{} 
    \item[] Justification: See Experiments Sections and Appendices.

\item {\bf Experiment statistical significance}
    \item[] Question: Does the paper report error bars suitably and correctly defined or other appropriate information about the statistical significance of the experiments?
    \item[] Answer: \answerYes{} 
    \item[] Justification: We follow standard evaluation paradigms by computing pass@1 accuracy with greedy sampling, which does not explicitly involve randomness.

\item {\bf Experiments compute resources}
    \item[] Question: For each experiment, does the paper provide sufficient information on the computer resources (type of compute workers, memory, time of execution) needed to reproduce the experiments?
    \item[] Answer: \answerYes{} 
    \item[] Justification: See appendicies, where we provide information about GPU resources.
    
\item {\bf Code of ethics}
    \item[] Question: Does the research conducted in the paper conform, in every respect, with the NeurIPS Code of Ethics \url{https://neurips.cc/public/EthicsGuidelines}?
    \item[] Answer: \answerYes{} 
    \item[] Justification: See paper.

\item {\bf Broader impacts}
    \item[] Question: Does the paper discuss both potential positive societal impacts and negative societal impacts of the work performed?
    \item[] Answer: \answerNA{} 
    \item[] Justification: This paper presents work whose goal is to advance the field of Machine Learning. It investigates fundamental aspects of instruction-tuning of language models and should not have direct societal impacts or implications that should be discussed here specifically, to the best of the authors’ knowledge.
    
\item {\bf Safeguards}
    \item[] Question: Does the paper describe safeguards that have been put in place for responsible release of data or models that have a high risk for misuse (e.g., pretrained language models, image generators, or scraped datasets)?
    \item[] Answer: \answerNA{} 
    \item[] Justification: We do not release data or models that have a high risk for misuse.

\item {\bf Licenses for existing assets}
    \item[] Question: Are the creators or original owners of assets (e.g., code, data, models), used in the paper, properly credited and are the license and terms of use explicitly mentioned and properly respected?
    \item[] Answer: \answerYes{} 
    \item[] Justification: Yes, we cite all the datasets and pretrained models used in the paper, which are all open-sourced for research use.

\item {\bf New assets}
    \item[] Question: Are new assets introduced in the paper well documented and is the documentation provided alongside the assets?
    \item[] Answer: \answerNA{} 
    \item[] Justification: We do not release new assets.

\item {\bf Crowdsourcing and research with human subjects}
    \item[] Question: For crowdsourcing experiments and research with human subjects, does the paper include the full text of instructions given to participants and screenshots, if applicable, as well as details about compensation (if any)? 
    \item[] Answer: \answerNA{} 
    \item[] Justification: We do not have crowdsourcing experiments or research with human subjects.

\item {\bf Institutional review board (IRB) approvals or equivalent for research with human subjects}
    \item[] Question: Does the paper describe potential risks incurred by study participants, whether such risks were disclosed to the subjects, and whether Institutional Review Board (IRB) approvals (or an equivalent approval/review based on the requirements of your country or institution) were obtained?
    \item[] Answer: \answerNA{} 
    \item[] Justification: We do not have crowdsourcing experiments or research with human subjects.

\item {\bf Declaration of LLM usage}
    \item[] Question: Does the paper describe the usage of LLMs if it is an important, original, or non-standard component of the core methods in this research? Note that if the LLM is used only for writing, editing, or formatting purposes and does not impact the core methodology, scientific rigorousness, or originality of the research, declaration is not required.
    \item[] Answer: \answerNA{} 
    \item[] Justification: We do not involve LLMs as a core component in implementing or developing the method of this work.

\end{enumerate}

\newpage
\appendix
\onecolumn
\section{Parameter Distance}
We measure the L2 norm of model parameter difference between fine-tuned and  pre-trained checkpoints, as a signal of how much the distribution has drifted during SFT~\cite{information_geometry_applications,cover_thomas_information_theory}. We notice that training over well-matched distribution shifts the parameter less than training over those ill-matched. 

\begin{table}[h!]
    \centering
    \begin{tabular}{lccc}
        \toprule
        & \textbf{Mistral-7B-v0.3} & \textbf{Llama3.1-8B} & \textbf{Qwen2.5} \\
        \midrule
        \textbf{\name}  & 8.006 & 8.196 & 8.426 \\
        \textbf{Worst} & 8.029 & 8.202 & 8.467 \\
        \bottomrule
    \end{tabular}
    \caption{Performance comparison across different models}
    \label{tab:model_comparison}
\end{table}



\section{Further Experiments}
\label{app:additional_exp}
\subsection{Comparing with Reward Based Selection}
We compare \name with purely reward-based selection. where for each instruction, we select the response with the highest scalar reward as determined by a reward model - Skywork-Reward-Llama3.1-8B-v0.2. Once the top-ranked response for each instruction is selected, we proceed with standard supervised fine-tuning on the resulting instruction-response pairs.The RFT setup provides a natural contrast to our proposed GRAPE method by emphasizing reward alignment over base-model alignment, thereby enabling us to disentangle the effects of distribution matching versus reward optimization in SFT data selection. 

As shown in Table~\ref{tab:grape-rft-comparison}, \name outperforms reward-based selection across both models and all benchmarks. These results suggest that aligning supervision with the base model’s own distributional preferences—rather than relying on external reward models—can yield better task performance.

\begin{table}[h]
\small
\centering

\label{tab:grape-rft-comparison}
\begin{tabular}{llccccccc}
\toprule
\textbf{Model} & \textbf{Method} & \textbf{AE WR} & \textbf{AE WR (LC)} & \textbf{LeetCode} & \textbf{MATH} & \textbf{MMLU} & \textbf{BBH} & \textbf{Avg} \\
\midrule
\multirow{2}{*}{LLaMA3.1-8B} 
& GRAPE & 15.2 & 14.8 & 19.4 & 32.1 & 64.5 & 69.6 & \textbf{35.9} \\
& Reward   & 14.0 & 14.5 & 17.2 & 31.3 & 63.3 & 69.0 & 34.9 \\
\midrule
\multirow{2}{*}{Mistral-v0.3-7B} 
& GRAPE & 13.9 & 13.6 & 18.3 & 24.2 & 59.2 & 62.3 & \textbf{31.9} \\
& Reward   & 12.5 & 13.9 & 13.3 & 22.4 & 58.5 & 62.2 & 30.5 \\
\bottomrule
\end{tabular}
\caption{Performance comparison between GRAPE and reward-based selection across benchmarks for LLaMA3.1-8B and Mistral-v0.3-7B. Metrics are benchmark-specific scores (higher is better).}
\end{table}
\subsection{Experiment on OpenHermes}

\begin{wraptable}{r}{0.6\textwidth}  
  \centering
  \small
  \begin{tabular}{cccccc}
    \hline
    \textbf{Item} &
      \textbf{Metric} &
      \textbf{Data} &
      \textbf{\begin{tabular}[c]{@{}c@{}}Llama\\3.1-8B\end{tabular}} &
      \textbf{\begin{tabular}[c]{@{}c@{}}Mistral\\-7B-v0.3\end{tabular}} &
      \textbf{\begin{tabular}[c]{@{}c@{}}Qwen\\2.5-7B\end{tabular}} \\
    \hline
     &  & Subset  &  8.6  &  5.9  &  7.6  \\
     &  & Random  &  8.0  &  6.2  &  9.0  \\
     &
      \multirow{-3}{*}{LC} &
      \cellcolor[HTML]{F3F4FF}\textbf{\name} &
      \cellcolor[HTML]{F3F4FF}\textbf{11.3} &
      \cellcolor[HTML]{F3F4FF}\textbf{8.2} &
      \cellcolor[HTML]{F3F4FF}\textbf{10.8} \\
    \cline{2-6}
     &  & Subset  &  6.2  &  3.9  &  5.2  \\
     &  & Random  &  6.4  &  4.8  &  7.2  \\
    \multirow{-6}{*}{\textbf{\begin{tabular}[c]{@{}c@{}}Alpaca\\-Eval2\end{tabular}}} &
      \multirow{-3}{*}{WR} &
      \cellcolor[HTML]{F3F4FF}\textbf{\name} &
      \cellcolor[HTML]{F3F4FF}\textbf{9.4} &
      \cellcolor[HTML]{F3F4FF}\textbf{7.5} &
      \cellcolor[HTML]{F3F4FF}\textbf{9.6} \\
    \hline
     &  & Subset  & 51.6 & 49.0 & 54.4 \\
     &  & Random  & 51.4 & 49.9 & 55.2 \\
    \multirow{-3}{*}{\textbf{\begin{tabular}[c]{@{}c@{}}Truthful\\-QA\end{tabular}}} &
      \multirow{-3}{*}{MC2} &
      \cellcolor[HTML]{F3F4FF}\textbf{\name} &
      \cellcolor[HTML]{F3F4FF}\textbf{52.7} &
      \cellcolor[HTML]{F3F4FF}\textbf{51.6} &
      \cellcolor[HTML]{F3F4FF}\textbf{56.4} \\
    \hline
  \end{tabular}
  \caption{Results on OpenHermes-2.5. The Subset row refers to training exclusively on the SFT responses over the subset.}
  \label{tab:hermes}
\end{wraptable}

To test the generality of our findings beyond the UltraInteract and Tulu-Olmo settings, we conduct additional experiments on the \textsc{OpenHermes-2.5}~\citep{OpenHermes} dataset—a large-scale, high-quality instruction-tuning corpus with approximately 1 million distinct instructions.

Following the setup from §\ref{sec:tulu_olmo}, we apply \name{} to select from responses aggregated across sources, including~\cite{open_hermes_preferences} and~\citet{huggingface2024openhermes}. For preference-based datasets, we retain only the winning responses to ensure quality, mirroring our earlier selection protocol. This results in 575K unique instructions and 1.34M instruction-response pairs.

As shown in Table~\ref{tab:hermes}, \name{} continues to outperform naive combination strategies. The consistent gains across diverse data sources and model families strengthen our central claim: \name{} is a general-purpose, model-aligned response selection strategy that reliably improves SFT performance in real-world, large-scale instruction tuning.

\subsection{Data Selection For Long Chain-of-Thoughts}

O1-/R1-style long chain-of-thoughts have drawn increasing attention. This paradigm, exemplified by models like OpenAI's O1 and DeepSeek-R1, has shown remarkable success in challenging domains such as mathematics and coding. 
We further experiment with the use of \name in long chain-of-thought distillation. We generate multiple candidate trajectories using R1-Distill-Qwen-1.5B~\cite{deepseekai2025deepseekr1incentivizingreasoningcapability} for a subset of OpenR1-Math~\citep{openr1} dataset --- LUFFY~\citep{luffy}, and verify the correctness of each, retaining the correct ones.

We compare the results on lowest versus highest perplexity instances below in table \ref{tab:long_math}.

\begin{table}[h!]
\centering
\begin{tabular}{ccc}
\hline
\textbf{Model} & \textbf{Perplexity} & \textbf{Acc.} \\
\hline
Qwen2.5-1.5B & Highest & 0.330 \\
Qwen2.5-1.5B & Lowest (\name) & 0.396 \\
\hdashline
Qwen2.5-3B   & Highest & 0.524 \\
Qwen2.5-3B   & Lowest (\name) & 0.544 \\
\hline
\end{tabular}
\caption{Performance metrics on MATH dataset}
\label{tab:long_math}
\end{table}

\subsection{Token-level \name}
To further investigate the alternative uses of our insight, we conduct experiments beyond data selection by incorporating token-level likelihoods directly into the training objective. Specifically, we modified the loss function to weigh each token proportionally to its likelihood.  We trained this variant on the OpenThoughts‑114k \cite{openthoughts} dataset, a curated collection of 114k high-quality reasoning samples spanning domains such as math, science, code etc. Evaluation followed the same benchmark suite as LUFFY \cite{luffy}, including competition-level math datasets (AIME24/25, AMC, MATH‑500, Minerva, OlympiadBench) and general reasoning tests (ARC‑c, GPQA‑Diamond, MMLU‑Pro). Our result in Table \ref{tab:token_level}
show that our token-level likelihood-weighted training yields consistent improvements across several benchmarks.
\begin{table}[ht]
\centering
\begin{tabular}{ccc}
\hline
\textbf{Benchmark} & \textbf{Baseline (\%)} & \textbf{Token-level \name\ (\%)} \\ 
\hline
\textsc{MATH}               & 55.60         & 60.20                   \\
\textsc{OlympiadBench}      & 21.04         & 28.30                   \\
\textsc{Minerva}            & 15.07         & 19.85                   \\
\textsc{AIME-24}            &  3.12         &  4.90                   \\
\textsc{AMC}                & 23.76         & 29.74                   \\
\textsc{AIME-25}            &  4.38         &  3.23                   \\
\textsc{ARC-C}              & 60.84         & 76.62                   \\
\textsc{GPQA-Diamond}       &  7.07         & 19.70                   \\
\textsc{MMLU-Pro}           & 27.02         & 34.35                   \\
\hline
\end{tabular}
\caption{Qwen2.5-3B performance (in \%) on various benchmarks under Baseline vs.\ Token-level \name.}
\label{tab:token_level}
\end{table}

\section{Further Details On Baselines}
\label{app:baseline_details}
This section details the experimental setup for our data selection baselines: \textbf{S2L},\textbf{LESS} and \textbf{NV-Embed}.

\subsection{S2L}

S2L, a state-of-the-art unsupervised data selection baseline, operates through two key steps: training a reference model to capture training dynamics and clustering the resulting trajectories to form a diverse, balanced subset of training data. The reference models used in our setup are specifically selected to enhance S2L’s performance, adhering to the theoretical underpinnings from the original paper that training dynamics remain consistent across models of varying sizes within the same family.

For our experiments, we train small reference models corresponding to the final target models. Specifically, we pair Llama-3.1-8B with Llama-3.2-1B, Qwen-2.5-7B with Qwen-2.5-0.5B, and Mistral-v0.3-7B with itself due to the lack of smaller models in the Mistral family. To minimize computational costs, LoRA is applied when training the Mistral reference model. \textbf{This choice of reference models are better compared to original S2L setup}, which employed a Pythia-70M proxy, thereby improving the fidelity of the selected subset.

Following S2L, the reference models are trained on a random 5\% subset of the dataset over four epochs. This reduced training requirement is justified by prior work, which demonstrates that only partial data is sufficient for the proxy model to learn meaningful training dynamics. During trajectory collection, we record the training loss of all examples at intervals of 500 iterations. The batch size and learning rate schedules are set as batch size of 128 and a learning rate warmup of 3\%, followed by a cosine decay to 2e-5.

We then perform K-means clustering using the Faiss library to efficiently partition the trajectory space into 100 clusters. The number of iterations is set to 20, and we use the Euclidean distance metric to ensure convergence to well-separated clusters. From each cluster, an equal number of examples are sampled to maintain a balanced subset distribution.
\subsection{LESS}
LESS is a state-of-the-art model-based and supervised data selection method that leverages gradient-based influence estimation. Given a small set of validation examples per task, LESS computes the influence of each training example by measuring the weighted cosine similarity between their LoRA gradients across multiple warmup checkpoints. It then aggregates these influence scores by averaging over validation examples within each task, followed by taking the maximum across tasks to obtain a scalar utility score per training example. Training examples are selected greedily based on these scores. In our experiments, we use the same base model for both selection and training, and follow the original LESS setup: 5\% warmup training for 4 epochs and a gradient projection dimension of 8192.

\subsection{NV-Embed}
Embedding-based data selection as detailed in \citep{ivison2025largescaledataselectioninstruction} is a supervised data selection method that ranks training examples by computing cosine similarity between their embeddings and those of validation examples. Unlike model-aware methods like LESS, embedding-based data selection  is model-agnostic: it relies on fixed, pretrained embedding models (in our case, we used NV-embed-v2, the state-of-the-art embedding model) rather than the target model. Instead of aggregating similarity scores into a single utility value per training example, embedding-based data selection  uses a round-robin strategy that iteratively selects the highest-scoring example for each validation instance, ensuring diverse coverage across tasks. We follow the original setup from \citet{ivison2025largescaledataselectioninstruction} in our experiments.

\section{Further Training Details}

We train our models on a 4-GPU Nvidia-GH200 node, with batch size 256 and micro batch size 2. 

\section{Further Ablations on UltraInteract. }
See Tables \ref{tab:ablations} and \ref{tab:self_distill_ultra}. 
\begin{table}[h]
\centering
\small
\begin{tabular}{lr|ccc}
\hline

\multicolumn{2}{c|}{\textbf{Data}} &
  \textbf{Full UI} &
  \textbf{Closest-1} &
  \textbf{Random-1} \\ \hline
\multicolumn{2}{c|}{\textbf{Num. Instances}} &
  \textbf{280K} &
  \textbf{~80K} &
  \textbf{~80K} \\ \hline
\multicolumn{2}{c|}{\textbf{HumanEval}}                        & 46.3 & 42.1 & 41.5 (-) \\
\rowcolor[HTML]{ECF4FF} 
\multicolumn{2}{c|}{\cellcolor[HTML]{ECF4FF}\textbf{LeetCode}} & 15.6 & 13.9 & 11.1 (-) \\
\multicolumn{2}{c|}{\textbf{MBPP}}                             & 50.1 & 52.1 & 49.1 (-) \\ \hline
\rowcolor[HTML]{ECF4FF} 
\multicolumn{1}{l|}{\cellcolor[HTML]{ECF4FF}}            &  \textsc{CoT} & 21.6 & 19.2 & 15.5 (-) \\
\rowcolor[HTML]{ECF4FF} 
\multicolumn{1}{l|}{\multirow{-2}{*}{\cellcolor[HTML]{ECF4FF}\textbf{MATH}}} &
   \textsc{PoT} &
  32.6 &
  24.9 &
  15.1 (-) \\ \hline
\multicolumn{1}{l|}{}                                    &  \textsc{CoT} & 45.9 & 44.1 & 35.3 (-) \\
\multicolumn{1}{l|}{\multirow{-2}{*}{\textbf{GSMPlus}}}  & \textsc{CoT} & 45.3 & 43.2 & 45.2 (-) \\ \hline
\rowcolor[HTML]{ECF4FF} 
\multicolumn{1}{l|}{\cellcolor[HTML]{ECF4FF}}            &  \textsc{CoT} & 16.8 & 15.8 & 15.8     \\
\rowcolor[HTML]{ECF4FF} 
\multicolumn{1}{l|}{\multirow{-2}{*}{\cellcolor[HTML]{ECF4FF}\textbf{TheoremQA}}} &
   \textsc{PoT} &
  20.1 &
  12.9 &
  15.3 \\ \hline
\multicolumn{2}{c|}{\textbf{Avg.}} &
  \textit{\textbf{32.7}} &
  \textit{\textbf{29.8}} &
  \textit{\textbf{27.1(-)}} \\ \hline
\end{tabular}
\caption{Ablations on data selection with \textsc{Mistral-7B-V0.3} by selecting within UltraInteract-SFT (since it contains varying numbers of responses per-instruction). Closest-1 denotes the one closest to the base model's initial distribution. Random-1 is sampled from the entire enlarged dataset formed by both original and generated responses. We use (-) to denote \textbf{Random-1} underperforming \textbf{Closest-1}. }
\label{tab:ablations}
\end{table}
\begin{table}[]
\centering
\small
\begin{tabular}{cccccccccccc}
\hline
\multirow{2}{*}{\textbf{Model}} &
  \multirow{2}{*}{\textbf{Data}} &
  \multirow{2}{*}{\textbf{HE}} &
  \multirow{2}{*}{\textbf{LC}} &
  \multirow{2}{*}{\textbf{MBPP}} &
  \multicolumn{2}{c}{\textbf{MATH}} &
  \multicolumn{2}{c}{\textbf{GSMPlus}} &
  \multicolumn{2}{c}{\textbf{TheoremQA}} &
  \multirow{2}{*}{\textbf{Avg.}} \\ \cline{6-11}
                             &                &      &      &      & CoT  & PoT  & CoT  & PoT  & CoT  & PoT  &      \\ \hline
\multirow{3}{*}{Mistral-7B}  & Self-Distill   & 46.3 & 13.3 & 49.6 & 17.3 & 18.5 & 43.3 & 33.2 & 16.8 & 17.4 & 28.4 \\
                             & Original-UI    & 46.3 & 15.6 & 50.1 & 21.6 & 32.6 & 45.9 & 45.3 & 16.8 & 20.1 & 32.7 \\
                             & Ours           & 52.4 & 15.6 & 53.4 & 28.9 & 34.6 & 50.5 & 52.8 & 17.8 & 20.6 & 36.3 \\ \hline
\multirow{3}{*}{Llama3.1-8B} & Self-Distilled & 47.6 & 6.7  & 51.7 & 22.9 & 12.7 & 47.2 & 35.3 & 18.8 & 21.5 & 29.4 \\
                             & Original-UI    & 54.3 & 11.1 & 58.9 & 29.7 & 31.0 & 53.7 & 51.6 & 20.0 & 20.8 & 36.8 \\
                             & Ours           & 57.3 & 19.4 & 63.8 & 34.8 & 39.2 & 56.6 & 56.1 & 22.5 & 23.9 & 41.5 \\ \hline
\multirow{3}{*}{Llama3.2-3B} & Self-Distilled & 32.3 & 5.6  & 41.9 & 8.8  & 7.0  & 12.1 & 12.1 & 5.9  & 10.5 & 15.1 \\
                             & Original-UI    & 32.9 & 3.9  & 41.6 & 12.8 & 16.1 & 30.8 & 19.5 & 14.6 & 10.5 & 20.3 \\
                             & Ours           & 42.6 & 13.3 & 44.6 & 16.4 & 17.6 & 34.9 & 20.6 & 15.1 & 11.4 & 24.1 \\ \hline
\end{tabular}
\caption{The detailed comparison across benchmarks for self-distillation discussed in Section~\ref{sec:self_distillation}}
\end{table}

\section{Additional Related Works On Model Dependent Data Selection Approaches}

Model-dependent data selection methods leverage internal signals from a target model—such as gradients, embeddings, or log-probabilities—to identify training examples that are most useful for fine-tuning. These approaches have led to strong empirical results across various settings. However, many of them involve substantial computational costs, such as repeated gradient computations or auxiliary model training, which can limit their scalability. We discuss these approaches and the costs they incur in this section. 

\subsection{Notations}

\begin{enumerate}
    \item A training dataset $D = \{ x_i \}_{i=1}^N$ of size $N$; the final language model to be trained on the selected data $\theta$.
    \item We denote \textbf{the average cost of one forward pass} of model $\theta$ on a training example as $F_{\theta}$. As one backward pass is approximately the cost of two forward passes, \textbf{the average cost of one ``gradient pass''} (i.e., one forward + one backward) is thus $3F_{\theta}$.
    \item Another important source of computational cost in data selection comes from the training of additional models. We use $C(\theta, D, T)$ to denote the cost of training model $\theta$ on dataset $D$ for $T$ epochs (i.e., $N \cdot T$ examples are seen in total).
    \item Therefore, we unify the computational cost of most data selection approaches into two parts:
    \begin{enumerate}
        \item \textbf{The training of additional models}. For example, gradient-based influence requires training an additional model on part of the training dataset for $T$ epochs to obtain the checkpoints for gradient computation.
        \item \textbf{The computation of per-sample features}. For example, for each training example, gradient-based influence requires computing its gradient for each saved checkpoint, which means $T$ gradient passes are needed.
    \end{enumerate}
    \item Note that some algorithms may have additional computational costs other than the two parts above, such as clustering or a greedy algorithm for the final data selection. Since the two parts above constitute the majority of computation for almost all the data selection approaches, \textbf{we omit the other cost and only focus on these two.}
\end{enumerate}

\subsection{TLDR: The Final Table}

For GRAPE, we assume that in the training dataset $D$, various responses to the same instruction are already available, thus no additional cost is incurred in the \textit{Response Collection} step of GRAPE. So the computational cost analysis of GRAPE under our framework is:

\begin{itemize}
    \item \textbf{Additional Training}: 0, as GRAPE directly evaluates data using the base model.
    \item \textbf{Per-sample conditional probability}: $NF_{\theta}$, as for a given target model $\theta$, we only need to compute conditional probability for each response (example) once.
\end{itemize}

The table below shows that our method, GRAPE, achieves superior performance with minimal computational cost compared with other model-based data selection approaches.

\begin{table}[h]
    \centering
    \begin{tabular}{@{}lcc@{}}
        \toprule
        & \textbf{Additional Training} & \textbf{Per-Sample Feature Computation} \\
        \midrule
        \textbf{GRAPE (ours)} & $0$ & $NF_{\theta}$ \\
        Gradient-based influence (LESS) & $C(\theta_{\text{lora}}, D_{\text{warmup}}, T)$ & $3T \cdot NF_{\theta}$ \\
        In-run gradient-based influence & $C(\theta, D, 1)$ & $0$ \\
        Gradient matching & $C(\theta_{\text{lora}}, D_{\text{warmup}}, T)$ & $3T \cdot NF_{\theta}$ \\
        Gradient norm & $m \cdot C(\theta, D, 1)$ & $3m \cdot NF_{\theta}$ \\
        Embedding-based & $0$ & $NF_{\theta}$ \\
        Simple uncertainty indicators & $0$ & $NF_{\theta}$ \\
        Perplexity & $C(\theta_{\text{ref}}, D_{\text{ref}}, 1)$ & $NF_{\theta_{\text{ref}}}$ \\
        Learnability & $C(\theta, D, 1)$ & $2 \cdot NF_{\theta}$ \\
        Loss trajectory (S2L) & $C(\theta_{\text{ref}}, D, T)$ & $T \cdot NF_{\theta_{\text{ref}}}$ \\
        \bottomrule
    \end{tabular}
    \caption{Computational cost comparison of data selection methods.}
\end{table}

\subsection{Gradient-based Methods}

Gradients have long been an important source of information for training data selection, as they directly affect the whole optimization process of language models. Three kinds of model-based gradient-based data selection approaches have been proposed:

\begin{enumerate}
    \item Gradient-based influence
    \item Gradient matching
    \item Gradient norm
\end{enumerate}

\subsubsection{Gradient-based Influence}

Gradient-based influence computes the pairwise influence scores between each pair of training and validation examples. Training data with the highest influence are selected, as training on them leads to the theoretically largest decrease in model loss on validation data. LESS \cite{xia2024less} formulates the pairwise influence scores as the cosine similarity between the gradients of training and validation data, and computes these gradient features using the following two steps:

\begin{enumerate}
    \item LoRA-train the final model on part of the whole training dataset, denoted as $D_{\text{warmup}}$, for $T$ epochs, and save the $T$ model checkpoints.
    \item For each data point, compute its LoRA gradient with each of the $T$ checkpoints, and later aggregate these $T$ gradients together in the cosine similarity expression.
\end{enumerate}

Therefore, the computational cost of gradient-based influence is:

\begin{itemize}
    \item \textbf{Additional training}: $C(\theta_{\text{lora}}, D_{\text{warmup}}, T)$.
    \item \textbf{Per-sample gradient for each checkpoint}: $NT \cdot 3F_{\theta} = 3 T \cdot NF_{\theta}$.
\end{itemize}

In order to reduce the cost incurred by per-sample gradient computation, recent work has developed \textit{in-run gradient-based influence} that directly computes the dot product between gradients without the need for separate gradient computations. However, this approach incorporates the dot product computations into the standard training process, which means in order to obtain pairwise influence scores for \textbf{the whole training set}, \textbf{a full training run} has to be done on all the training data. This incurs inefficiency when we do not actually need full dataset training. Moreover, the pairwise scores here only show the model’s \textbf{``dynamic preference''}: scores computed at the $t$-th iteration only reflect the model’s preference at this specific iteration. It is not theoretically guaranteed that these scores reflect the model’s preference from the beginning of training. Thus, the cost of in-run gradient-based influence is:

\begin{itemize}
    \item \textbf{Additional Training}: $C(\theta, D, 1)$.
    \item \textbf{Per-sample gradient}: 0.
\end{itemize}

\subsubsection{Gradient Matching}

Gradient matching also requires per-sample gradients, but utilizes their information in a different way. It performs clustering based on these gradient features to group similar data, and then applies an iterative greedy selection algorithm. \textbf{In order to scale to LLM-level gradient computation and clustering, TAGCOS \cite{TAGCOS} completely follows the warmup training and gradient computation pipeline of LESS \cite{xia2024less}.} As the computational bottleneck here is still the gradient computation instead of clustering or iterative selection, \cite{TAGCOS} also shares the same computational cost as \cite{xia2024less}:

\begin{itemize}
    \item \textbf{Additional training}: $C(\theta_{\text{lora}}, D_{\text{warmup}}, T)$.
    \item \textbf{Per-sample gradient for each checkpoint}: $NT \cdot 3F_{\theta} = 3 T \cdot NF_{\theta}$.
\end{itemize}

\subsubsection{Gradient Norm}

The $L_2$-norms of gradient vectors can also serve as effective indicators for data selection. \cite{GraNd} proposes GraNd, which obtains a utility score for each training point based on its gradient norm early in the training. More specifically, it starts from $m$ different model weight initializations, trains each model on the whole dataset to obtain per-sample gradient norms, and finally averages the $m$ gradient norms for each training point to obtain the final GraNd score. Therefore, the computational cost of GraNd is shown below:

\begin{itemize}
    \item \textbf{Additional training}: $m \cdot C(\theta, D, 1)$.
    \item \textbf{Per-sample gradient for each weight initialization}: $Nm \cdot 3F_{\theta} = 3m \cdot NF_{\theta}$.
\end{itemize}

\subsection{Embedding-based Methods}

Embedding-based methods project the whole training set into an embedding space to quantify the information of each data point and their interactions. For model-based embedding-based selection methods, the embeddings are usually computed by the final model $\theta$ to align with its preference.

Under a supervised data selection setup where validation data representing target task distributions are available, \textbf{Representation-based Data Selection} (RDS; \cite{RDS1, RDS2}) computes the embedding similarity between training and validation data, and selects training points that are most similar to the target distribution in the embedding space.

For an unsupervised setup where only the embeddings of training data are accessible, \textbf{geometry-based coreset sampling} methods are widely used \cite{Coreset}. Grounded on the intuition that close samples in the embedding space often share similar properties, a \textbf{diverse} subset can be obtained by controlling the minimum distance between any two selected data points. Among them, using K-center greedy sampling to select embedding-based \textbf{facility locations} has been proven especially effective for instruction fine-tuning of LLMs \cite{FacilityLocations}.

These embedding-based approaches share similar computational costs: they do not need any additional model training and can directly extract useful per-sample embeddings using the last-layer hidden states of the pretrained final model $\theta$. Thus, their computational cost is shown below:

\begin{itemize}
    \item \textbf{Additional training}: 0.
    \item \textbf{Per-sample embedding computation}: $NF_{\theta}$.
\end{itemize}

\subsection{LogProb-based Methods}

LogProb-based methods also directly utilize the target LLM to evaluate the utility of each training data point.

\subsubsection*{4.1 Simple Uncertainty-based Indicators}

Some simple model-based indicators inspired by the notion of uncertainty have been shown effective for a long time and recently extended to data selection for LLM instruction tuning \citep{Uncertainty1, FacilityLocations}. \cite{FacilityLocations} demonstrates the effectiveness of various indicators including \textbf{mean entropy, least confidence, mean margin}, etc. These simple indicators do not require additional training and can also be directly obtained with the pretrained final model $\theta$. Their computational cost is shown below:

\begin{itemize}
    \item \textbf{Additional training}: 0.
    \item \textbf{Per-sample per-token logits computation}: $NF_{\theta}$.
\end{itemize}

\subsubsection{Perplexity (PPL)}

PPL is also a long-standing data selector and has been shown effective for LLM-scale data selection. Typically, a split of the training dataset, $D_{\text{ref}}$, is needed to train $\theta_{\text{ref}}$, a reference model that will be used to compute PPL for the whole training set.

A common approach is to use the final model $\theta$ as the reference model $\theta_{\text{ref}}$ to ensure the alignment in PPL patterns \cite{PPL1}, but prior work \cite{ankner2024perplexed} also shows that a reference model much smaller than the final model can also be an effective PPL-based data selector. The computational cost for PPL-based selection is shown below:

\begin{itemize}
    \item \textbf{Additional training}: $C(\theta_{\text{ref}}, D_{\text{ref}}, 1)$.
    \item \textbf{Per-sample PPL computation}: $NF_{\theta_{\text{ref}}}$.
\end{itemize}

\subsubsection{Learnability}

In addition, \textbf{learnability \citep{Learnability1, Learnability2, Learnability3}} is a more effective metric than pure uncertainty or PPL, as it excludes uncertain but unlearnable points (e.g., noisy or less task-relevant) by considering the decrease in per-sample loss before and after the model is fully trained. More specifically, it trains the final model $\theta$ on the full training dataset to obtain a strong reference model $\theta_{\text{ref}}$, and then computes the difference of loss on each training example between $\theta$ and $\theta_{\text{ref}}$. In this way, it requires two forward passes for per-sample computation:

\begin{itemize}
    \item \textbf{Additional training}: $C(\theta, D, 1)$.
    \item \textbf{Per-sample learnability computation}: $2 \cdot NF_{\theta}$.
\end{itemize}

\subsubsection{Loss Trajectory}

Moreover, logprob-based methods can also obtain finer-grained information from the \textbf{training dynamics} of LLMs. S2L \cite{yang2024s2l} obtains a feature vector for \textbf{each} training point by collecting their \textbf{training loss trajectories} over $T$-epoch training on a small reference model $\theta_{\text{ref}}$, and then applies K-means clustering to equally sample data points from each trajectory cluster. Prior work shows its superiority over other logprob-based indicators, but it also comes with significant computational cost:

\begin{itemize}
    \item \textbf{Additional training}: $C(\theta_{\text{ref}}, D, T)$. Here the choice of $\theta_{\text{ref}}$ is especially important, as prior work \cite{S2LRef} shows that reference models that come from the same model family as the final model tend to have similar loss trajectories of training data, so they can preserve more fidelity in their loss trajectory patterns.
    \item \textbf{Per-sample loss trajectory computation}: $T \cdot NF_{\theta_{\text{ref}}}$. Note that $T$ here is typically much larger than that in gradient-based influence computation, so the computational cost of this gradient-free approach can be even higher than gradient-based methods.
\end{itemize}

\section{Multi-Round \name}

To evaluate whether \name continues to provide benefits after an initial round of fine-tuning, we conducted a second-round experiment using \textsc{LLaMA3.1-8B}. In the first round, we selected the best responses per instruction using \name and fine-tuned the model accordingly. Then, we re-applied \name on the newly fine-tuned model to select a fresh set of responses and conducted another round of fine-tuning. This second iteration led to further performance improvements across multiple benchmarks, including AlpacaEval, WizardEval (both original and LeetCode), and real-world tasks such as MATH, MMLU, and BBH. Notably, \name Round 2 improved average benchmark scores from 35.9\% to 37.3\%, demonstrating that the method remains effective and even compounding when iteratively applied.

\begin{table}[h]
\centering

\label{tab:grape_round2}
\begin{tabular}{lcccccc}
\toprule
\textbf{Method} & AE & WR & WR (LC) & MATH & MMLU & BBH \\
\midrule
GRAPE Round 1 & 15.2 & 14.8 & 19.4 & 32.1 & 64.5 & 69.6 \\
GRAPE Round 2 & \textbf{17.6} & \textbf{19.6} & 18.9 & \textbf{33.2} & 64.5 & \textbf{70.0} \\
\bottomrule
\end{tabular}
\caption{Performance of \textsc{LLaMA3.1-8B} after two rounds of \name fine-tuning.}
\end{table}

\section{Details Of Correlation Analysis}
\label{app:heatmap}

To analyze how well different models align with the training distribution, we conducted a perplexity-based study on the Tulu-v3 training set. Specifically, we randomly sampled 1{,}000 instances from the training data and used a wide array of generator models to produce responses for each instruction. For each response, we computed its perplexity under each base model and ranked the responses by perplexity, with lower perplexity indicating better matching of distribution. We then identified the top-1 response per instance for each base model and visualized the overall rankings using a heatmap (Figure~\ref{fig:breakdown}).

\end{document}